\documentclass{article}

\usepackage[final]{neurips_2022}

\usepackage[utf8]{inputenc} 
\usepackage[T1]{fontenc}    
\usepackage{hyperref}       
\usepackage{url}            
\usepackage{booktabs}       
\usepackage{amsfonts}       
\usepackage{nicefrac}       
\usepackage{microtype}      
\usepackage{xcolor}         
\usepackage{subfiles}
\usepackage{comment}
\usepackage{enumitem}
\usepackage{graphicx} 
\usepackage{bmpsize}

\usepackage[export]{adjustbox}
\usepackage{wrapfig}
\usepackage{tcolorbox}
\usepackage{tablefootnote}
\usepackage{float} 
\usepackage{colortbl}
\usepackage[labelformat=empty, position=top]{subcaption}

\definecolor{bluecite}{HTML}{0875b7}
\definecolor{Gray}{gray}{0.95}

\newtcolorbox{mybox}[1]{colback=bluecite!5!white,colframe=bluecite!75!black,fonttitle=\bfseries,title=#1}

\hypersetup{
  colorlinks=true,
  citecolor=bluecite,
  linkcolor=magenta
}

\title{Towards a Standardised Performance Evaluation Protocol for Cooperative MARL}

\author{%
Rihab Gorsane$^{1}$\thanks{Equal contribution. Corresponding author: r.gorsane@instadeep.com} \\
\And
Omayma Mahjoub$^{12*}$\thanks{Work done during an internship at InstaDeep.}\\
\And
Ruan de Kock$^{1*}$\\
\And
Roland Dubb$^{13\dagger}$ \\
\And
Siddarth Singh$^{1}$ \\
\And
Arnu Pretorius$^{1}$ \\
\AND
\vspace{-5mm}\\
$^{1}$InstaDeep\\
$^{2}$National School of Computer Science, Tunisia\\
$^{3}$University of Cape Town, South Africa\\
}

\begin{document}

\maketitle

\begin{abstract}
Multi-agent reinforcement learning (MARL) has emerged as a useful approach to solving decentralised decision-making problems at scale. Research in the field has been growing steadily with many breakthrough algorithms proposed in recent years. In this work, we take a closer look at this rapid development with a focus on evaluation methodologies employed across a large body of research in cooperative MARL. By conducting a detailed meta-analysis of prior work, spanning 75 papers accepted for publication from 2016 to 2022, we bring to light worrying trends that put into question the true rate of progress. We further consider these trends in a wider context and take inspiration from single-agent RL literature on similar issues with recommendations that remain applicable to MARL. Combining these recommendations, with novel insights from our analysis, we propose a standardised performance evaluation protocol for cooperative MARL. We argue that such a standard protocol, if widely adopted, would greatly improve the validity and credibility of future research, make replication and reproducibility easier, as well as improve the ability of the field to accurately gauge the rate of progress over time by being able to make sound comparisons across different works. Finally, we release our meta-analysis data publicly on our project website for future research on evaluation: \href{https://sites.google.com/view/marl-standard-protocol/}{https://sites.google.com/view/marl-standard-protocol}
\end{abstract}


\section{Introduction} 

Empirical evaluation methods in single-agent reinforcement learning (RL)\footnote{In this paper, we use the term "RL" to exclusively refer to \emph{single-agent} RL, as opposed to RL as a field of study, of which MARL is a subfield.} have been closely scrutinised in recent years \citep{IslamReproducibility2017, Machado2017, MSc-2017-Henderson, Zhang2018, henderson2019deep,colas2018random, colas2019hitchhikers, chan-2020-measuringReliability, jordan2020evaluating, engstrom2020implementation, agarwal2022deep}. In this context, the impact of a lack of rigour and methodological standards has already been observed. Fortunately, just as these issues have arisen and been identified, they have also been accompanied by suggested solutions and recommendations from these works. 

Multi-agent reinforcement learning (MARL) extends RL with capabilities to solve large-scale decentralised decision-making tasks where many agents are expected to coordinate to achieve a shared objective \citep{MARL, ReviewCoopMARL-2019-Oroojlooyjadid,GameTheoryMARLOverviewYang}. In this cooperative setting, where there is a common goal and rewards are shared between agents, sensible evaluation methodology from the single-agent case often directly translates to the multi-agent case. However, despite MARL being far less developed and entrenched than RL, arguably making adoption of principled evaluation methods easier, the field remains affected by many of the same issues as found in RL. Implementation variance, inconsistent baselines, and insufficient statistical rigour still affect the quality of reported results. Although work specifically addressing these issues in MARL have been rare, there have been recent publications that implicitly observe some of the aforementioned issues and attempt to address their symptoms but arguably not their root cause \citep{SurpisingPPO-ChaoYu-2021, papoudakis2021benchmarking, hu2022rethinking}. These works usually attempt to perform new summaries of performance or look into the code-level optimisations used in the literature to provide insight into the current state of research. 

In this paper, we argue that to facilitate long-term progress in future research, MARL could benefit from the literature found in RL on evaluation. The highlighted issues and recommendations from RL may serve as a guide towards diagnosing similar issues in MARL, as well as providing scaffolding around which a standardised protocol for evaluation could be developed. In this spirit, we survey key works in RL over the previous decade, each relating to a particular aspect of evaluation, and use these insights to inform potential standards for evaluation in MARL. 

\begin{wrapfigure}{r}{0.5\textwidth}
    \centering
    \includegraphics[width=0.5\textwidth]{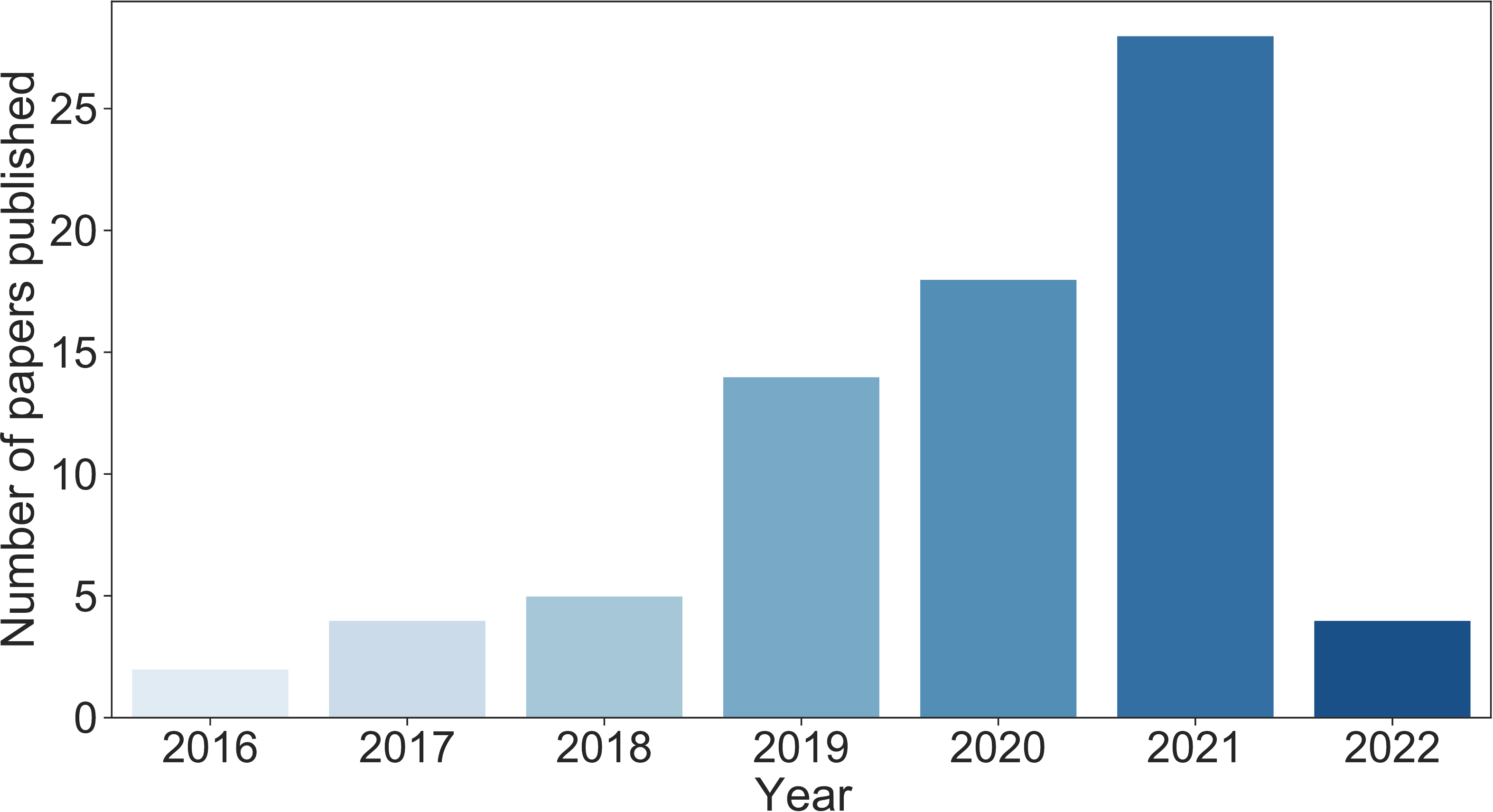}
    \caption{Recorded papers by year in the meta-analysis on evaluation methodologies in cooperative MARL.}
    \label{fig: number_of_papers}
\end{wrapfigure}

For each aspect of evaluation considered, we provide a detailed assessment of the corresponding situation in MARL through a meta-analysis of prior work. In more detail, our meta-analysis involved manually annotating MARL evaluation methodologies found in research papers published between 2016 to 2022 from various conferences including NeurIPS, ICML, AAMAS and ICLR, with a focus on \textit{deep} cooperative MARL (see Figure \ref{fig: number_of_papers}). In total, we collected data from 75 cooperative MARL papers accepted for publication. Although we do not claim our dataset to comprise the entire field of modern deep MARL, to the best of our knowledge, our data includes all popular and recent deep MARL algorithms and methodologies from seminal papers. We believe this dataset is the first of its kind and we have made it publicly available for further analysis.

By mining the data on MARL evaluation from prior work, we highlight how certain trends, worrying inconsistencies, poor reporting, a lack of uncertainty estimation, and a general absence of proper standards for evaluation, is plaguing the current state of MARL research, making it difficult to draw sound conclusions from comparative studies. We combine these findings with the earlier issues and recommendations highlighted in the literature on RL evaluation, to propose a standardised performance evaluation protocol for cooperative MARL research. 

In addition to a standardised protocol for evaluation, we expose trends in the use of benchmark environments and suggest useful standards for environment designers that could further improve the state of evaluation in MARL. We touch upon aspects of environment bias, manipulation, level/map cherry picking as well as scalability and computational considerations. Our recommendations pertain to designer specified standards concerning the use of a particular environment, its agreed upon settings, scenarios and version, as well as improved reporting by authors and preferring the use of environments designed to test generalisation. 

There are clear parallels between RL and MARL evaluation, especially in the cooperative setting. Therefore, we want to emphasise that our contribution in this work is less focused on innovations in protocol design (of which much can be ported from RL) and more focused on data-driven insights on the current state of MARL research and a proposal of standards for MARL evaluation.  


\section{From RL to MARL evaluation: lessons, trends and recommendations} 

In this section, we provide a list of key lessons from RL that are applicable to MARL evaluation. We highlight important issues identified from the literature on RL evaluation, and for each issue, provide an assessment of the corresponding situation and trends in MARL. Finally, we conclude each lesson with recommendations stemming from our analysis and the literature. 


\subsection{Lesson 1: Know the true source of improvement and report everything}

 \textbf{Issue in RL} -- \textit{Confounding code-level optimisations and poor reporting}: It has been shown empirically that across some RL papers there is considerable variance in the reported results for the same algorithms evaluated on the same environments \citep{henderson2019deep, jordan2020evaluating}. This variance impedes the development of novel algorithmic developments by creating misleading performance comparisons and making direct comparisons of results across papers difficult. Implementation and code-level optimisation differences can have a significant impact on algorithm performance and may act as confounders during performance evaluation \citep{engstrom2020implementation,andrychowicz2020matters}. It is rare that these implementation and code-level details are reported, or that appropriate ablations are conducted to pinpoint the true source of performance improvement.

\begin{figure}[t]
    \centering
    \begin{subfigure}[t]{0.02\textwidth}
        \scriptsize
        \textbf{(a)}
    \end{subfigure}
    \begin{subfigure}[t]{0.28\textwidth}
        \includegraphics[width=\linewidth, valign=t]{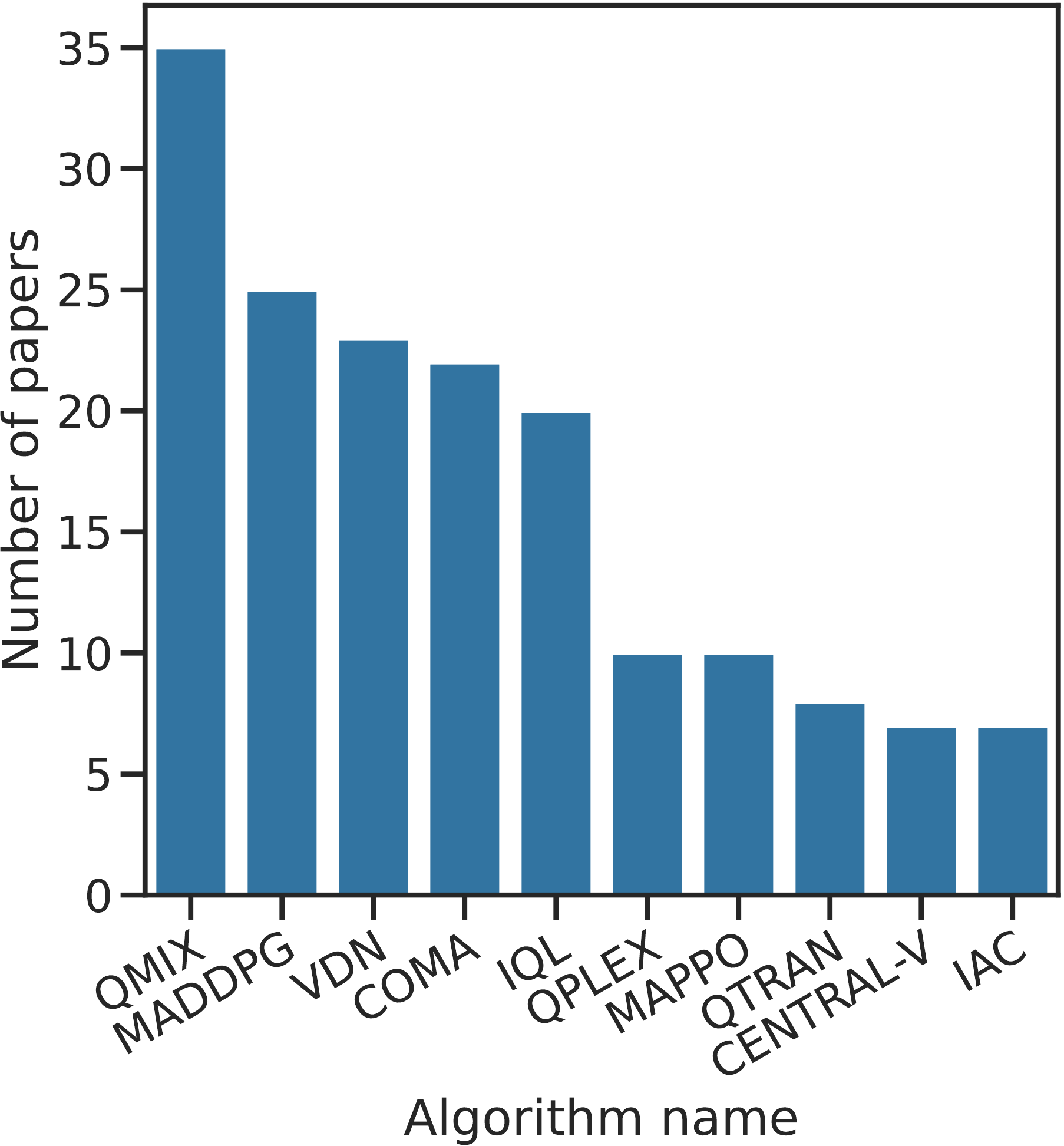}
    \end{subfigure}
    \hspace{1mm}
    \begin{subfigure}[t]{0.02\textwidth}
        \scriptsize
        \textbf{(b)}
    \end{subfigure}
    \begin{subfigure}[t]{0.58\textwidth}
        \includegraphics[width=\linewidth, valign=t]{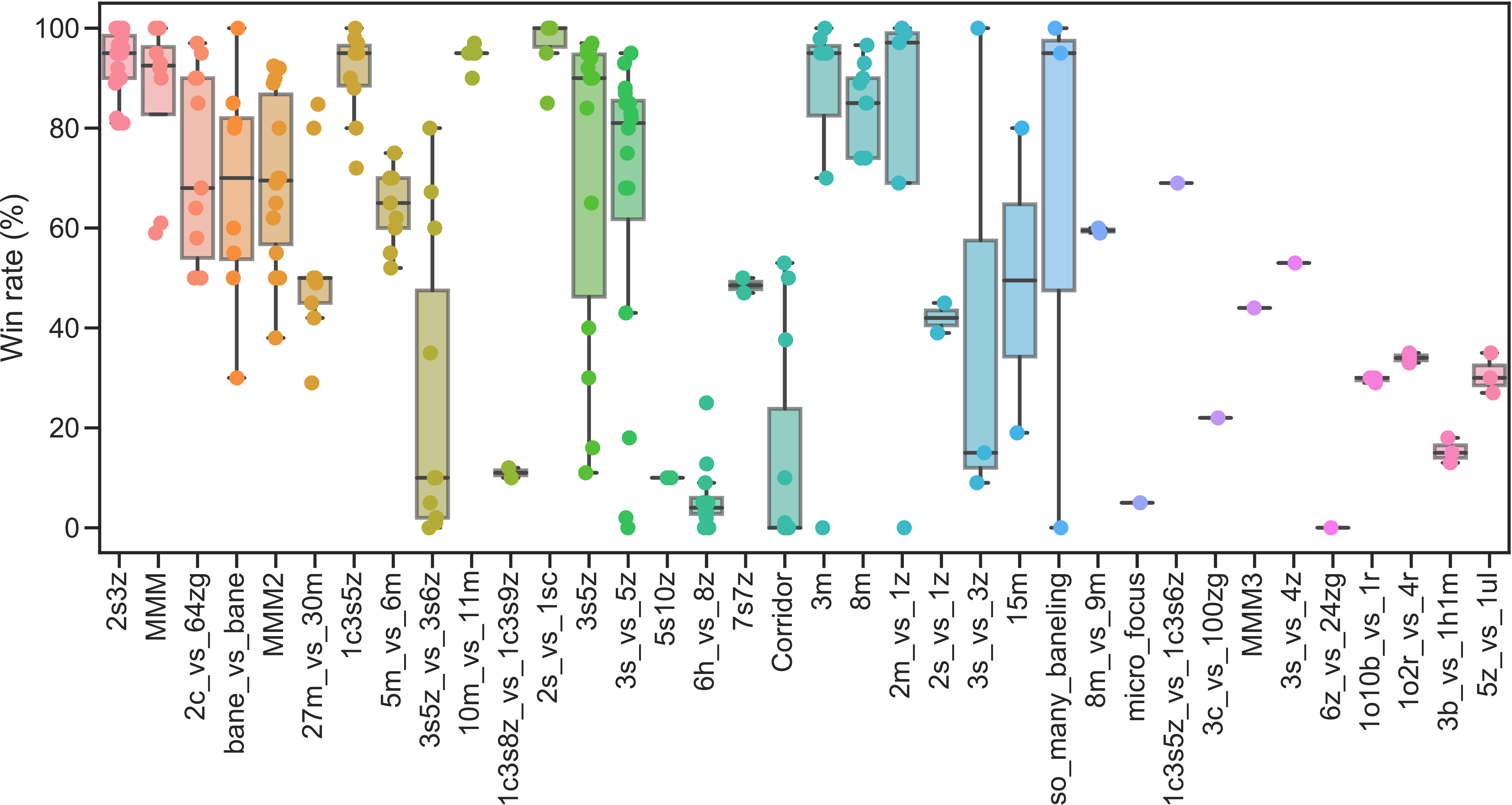}
    \end{subfigure}

    \vspace{0.4cm}

    \begin{subfigure}[t]{0.02\textwidth}
        \scriptsize
        \textbf{(c)}
    \end{subfigure}
    \begin{subfigure}[t]{0.45\textwidth}
        \includegraphics[width=\linewidth, valign=t]{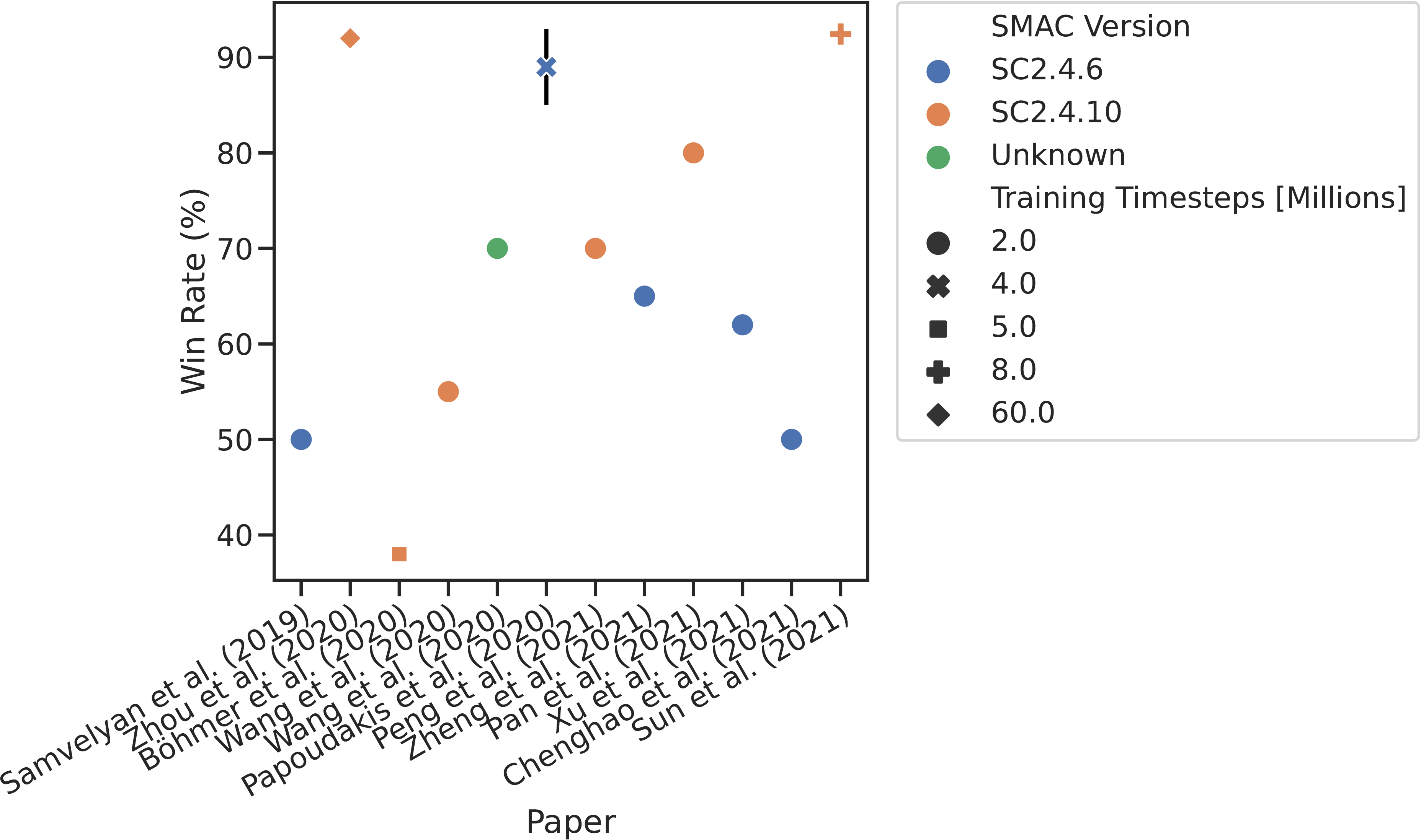}    
    \end{subfigure}
    \hspace{1mm}
    \begin{subfigure}[t]{0.02\textwidth}
        \scriptsize
        \textbf{(d)}
    \end{subfigure}
    \begin{subfigure}[t]{0.4\textwidth}
        \includegraphics[width=\linewidth, valign=t]{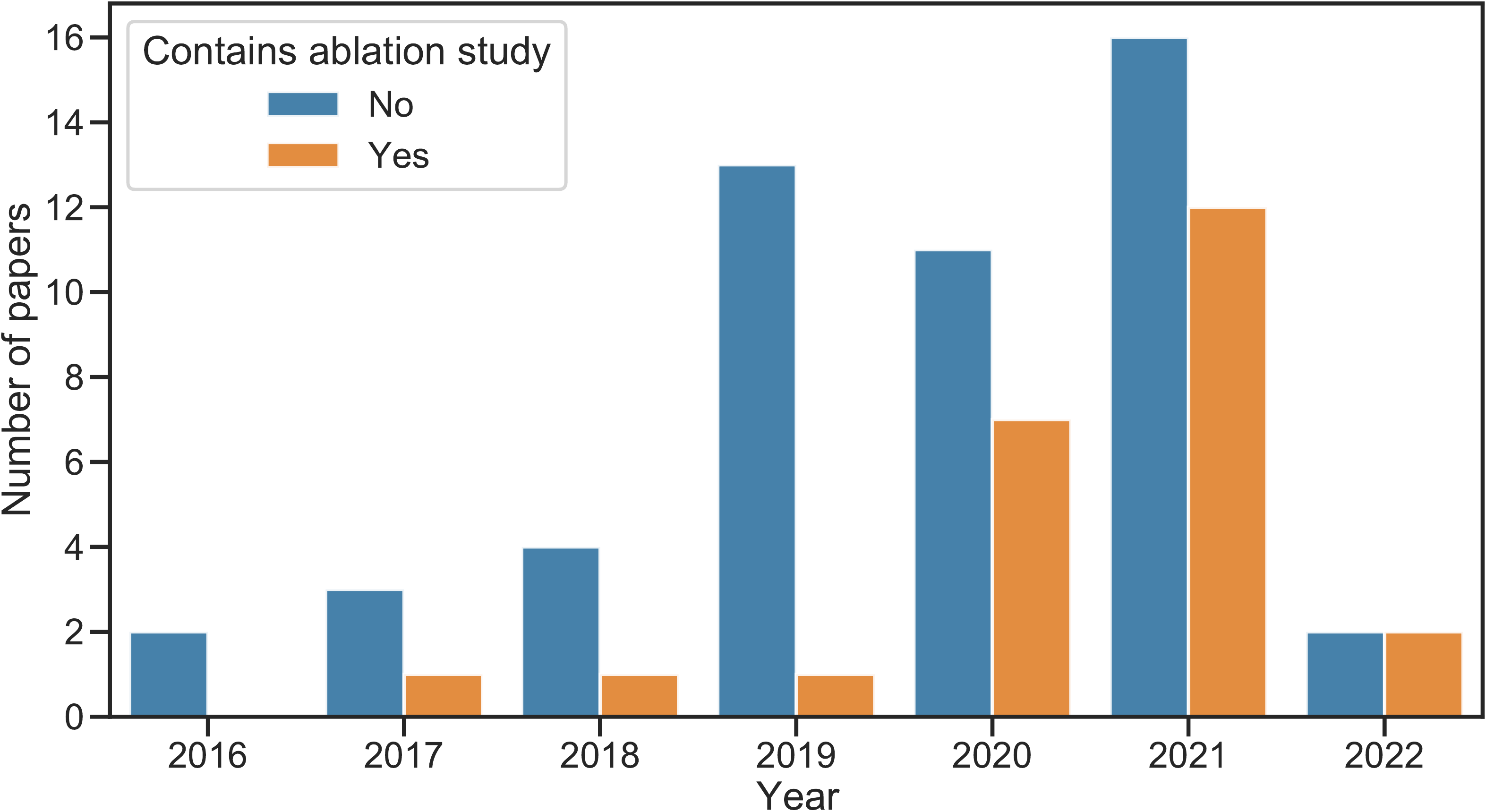}    
    \end{subfigure}
    \caption{\textit{Inconsistencies in performance reports and a lack of ablation studies}. \textbf{(a)} MARL algorithms ranked by popularity. \textbf{(b)} Historical performance of QMIX on different SMAC maps across papers. \textbf{(c)} Performance of QMIX on the MMM2 SMAC map as reported in different papers. \textbf{(d)} A count of papers over the years containing any type of ablation study as part of their evaluation of a newly proposed algorithm.}
    \label{fig: smac mmm2 qmix}
\end{figure}

 \textbf{The situation in MARL} -- \textit{Similar inconsistencies and poor reporting but a promising rise in ablation studies}: There already exist some work in MARL showing the effects of specific code-level optimisations on evaluation performance \citep{hu2021rethinking}. Employing different optimisers, exploration schedules, or simply setting the number of rollout processes to be different, can have a significant effect on algorithm performance. To better understand the variance in performance reporting across works in MARL, we focused on QMIX \citep{rashid2018qmix}, the most popular algorithm in our dataset (as shown in Figure \ref{fig: smac mmm2 qmix} (a)). In Figure \ref{fig: smac mmm2 qmix} (b), we plot the performance of QMIX tested on different maps from the StarCraft multi-agent challenge (SMAC) environment \citep{samvelyan19smac}, a popular benchmark in MARL. On several maps, we find wildly different reported performances with large discrepancies across papers. Although it is difficult to pin down the exact source of these differences in reports, we zoom in with our analysis to only consider a single environment, in this case \textit{MMM2}. We find that some of the variance is explained by differences in the environment version as well as the length of training time, as shown in Figure \ref{fig: smac mmm2 qmix} (c). However, even when both of these aspects are controlled for, as well as any implementation or evaluation details mentioned in each paper, differences in performance are still observed (as seen by comparing orange and blue circles, respectively). This provides evidence that unreported implementation details, or differences in evaluation methodology account for some of the observed variance and act as confounders when comparing performance across papers (similar inconsistencies in other maps are shown in the Appendix). We finally consider studying the explicit attempts in published works at understanding the sources of algorithmic improvement through the use of ablation studies. We find that very few of these studies were performed in the earlier years of MARL (see Figure \ref{fig: smac mmm2 qmix} (d)). However, even though roughly 40\% of papers in 2021 still lacked any form of ablation study, we find a promising trend showing that ablation studies have become significantly more prevalent in recent~years. 

\textbf{Recommendations} -- \textit{Report all experimental details, release code and include ablation studies}: \cite{henderson2019deep} emphasise that for results to be reproducible, it is important that papers report \textit{all} experimental details. This includes hyperparameters, code-level optimisations, tuning procedures, as well as a precise description of how the evaluation was performed on both the baseline and novel work. It is also important that code be made available to easily replicate findings and stress test claims of general performance improvements. Furthermore, \cite{engstrom2020implementation} propose that algorithm designers be more rigorous in their analysis of the effects of individual components and how these impact performance through the use of detailed ablation studies. It is important that researchers practice diligence in attributing how the overall performance of algorithms and their underlying algorithmic behavior are affected by different proposed innovations, implementation details and code-level optimisations. In the light of our above analysis, we argue that the situation is no different in MARL, and therefore suggest the field adopt more rigorous reporting and conduct detailed ablation studies of proposed innovations.

\subsection{Lesson 2: Use standardised statistical tooling for estimating and reporting uncertainty} 

\textbf{Issue in RL} -- \textit{Results in papers do not take into account uncertainty}: We have discussed how different implementations of the same algorithm with the same set of hyperparameters can lead to drastically different results. However, even under such a high degree of variability, typical methodologies often ignore the uncertainty in their reporting \citep{colas2018random, colas2019hitchhikers, jordan2020evaluating, agarwal2022deep}. Furthermore, most published results in RL make use of point estimates like the mean or median performance and do not take into account the statistical uncertainty arising from only using a finite number of testing runs. For instance, \cite{agarwal2022deep} found that the current norm of using only a few runs to evaluate the performance of an RL algorithm is insufficient and does not account for the variability of the point estimate used. Furthermore, \cite{agarwal2022deep} also revealed that the manner in which point estimates are chosen varies between authors. This inconsistency invalidates direct comparison between results across papers. 

\begin{figure}
    \centering
    \begin{subfigure}[t]{0.02\textwidth}
        \scriptsize
        \textbf{(a)}
    \end{subfigure}
    \begin{subfigure}[t]{0.25\textwidth}
        \includegraphics[width=\linewidth, valign=t]{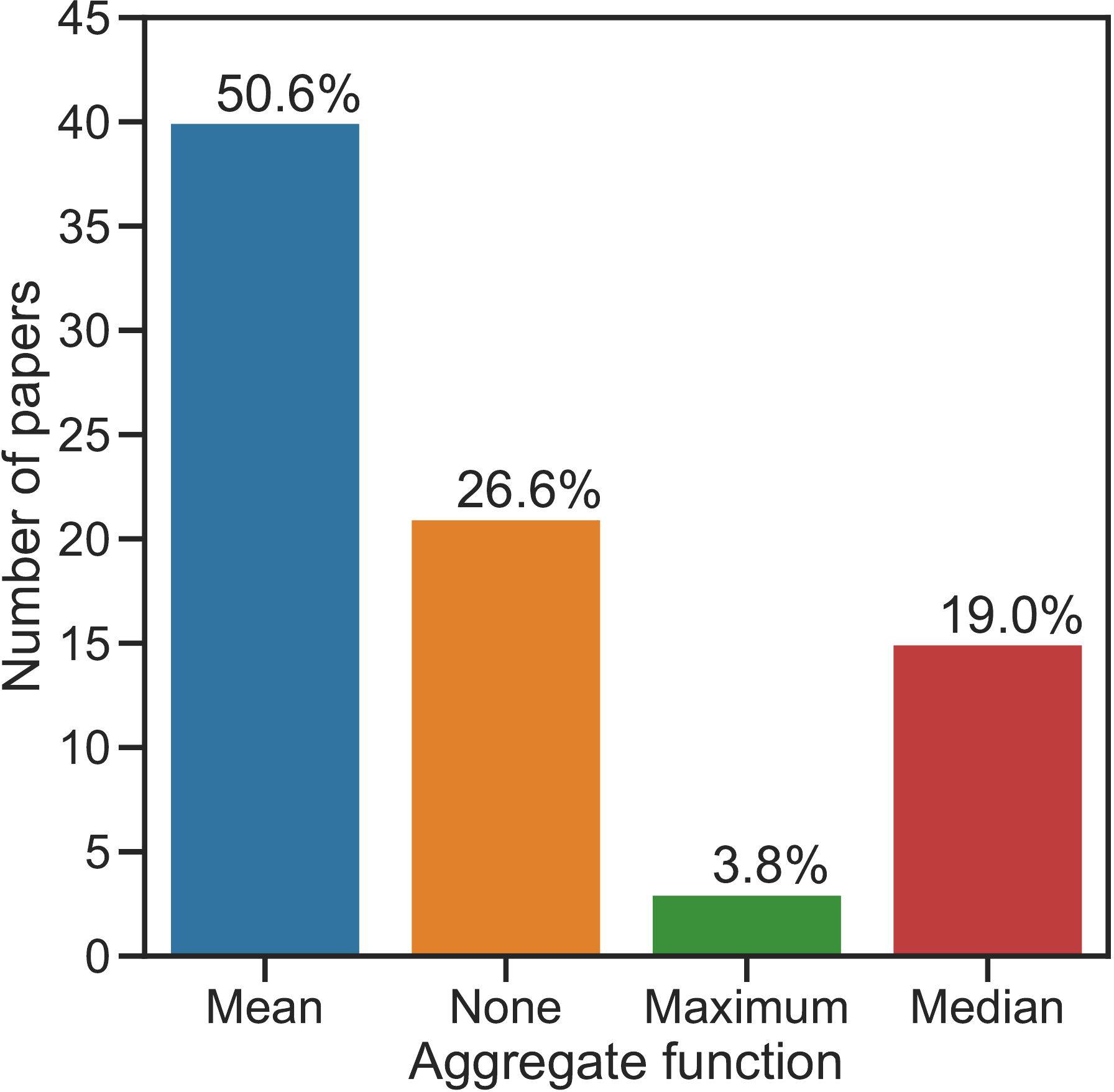}
    \end{subfigure}
    \hspace{1mm}
    \begin{subfigure}[t]{0.02\textwidth}
        \scriptsize
        \textbf{(b)}
    \end{subfigure}
    \begin{subfigure}[t]{0.23\textwidth}
        \includegraphics[width=\linewidth, valign=t]{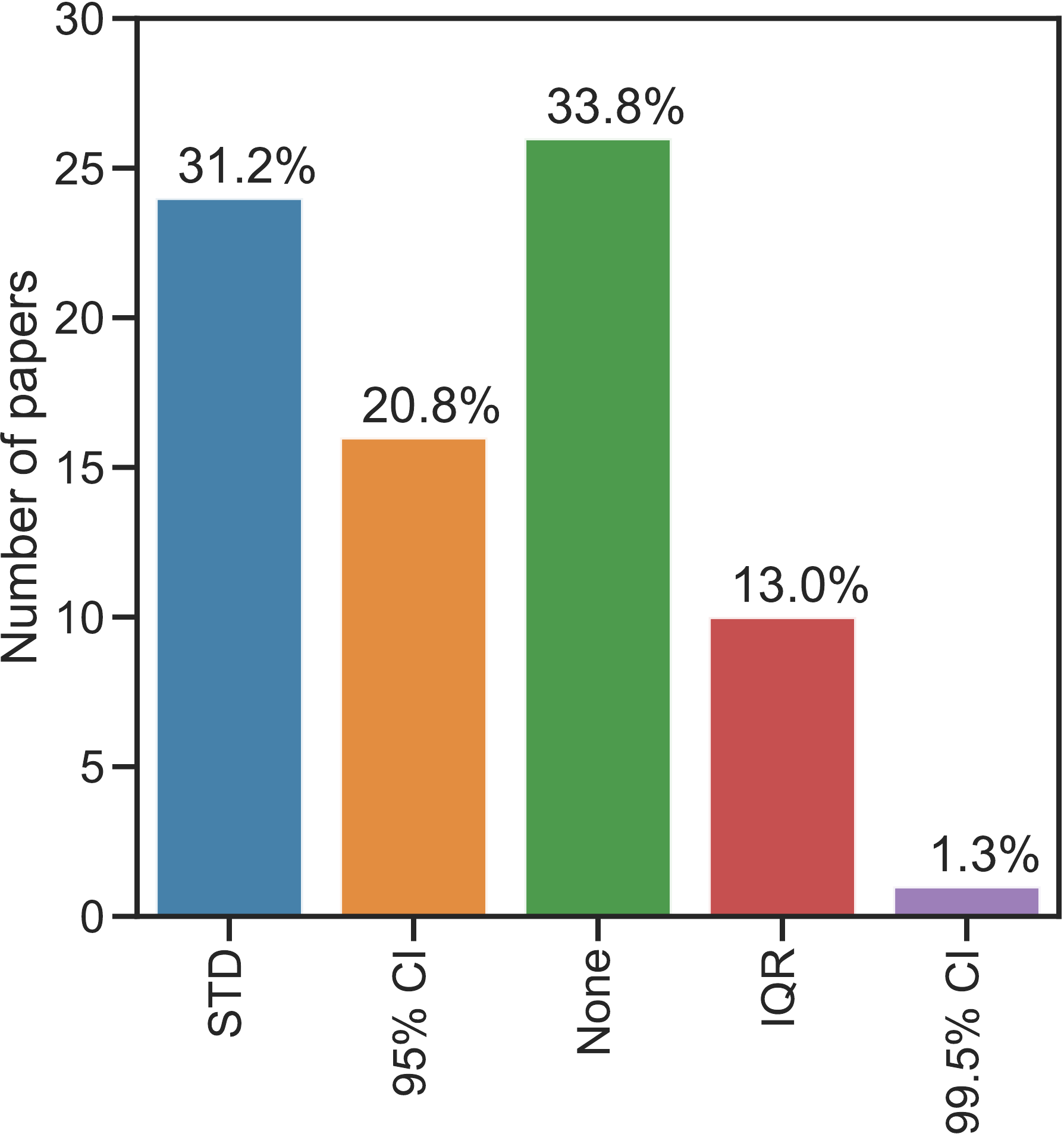}
    \end{subfigure}
    \hspace{1mm}
    \begin{subfigure}[t]{0.02\textwidth}
        \scriptsize
        \textbf{(c)}
    \end{subfigure}
    \begin{subfigure}[t]{0.4\textwidth}
        \includegraphics[width=\linewidth, valign=t]{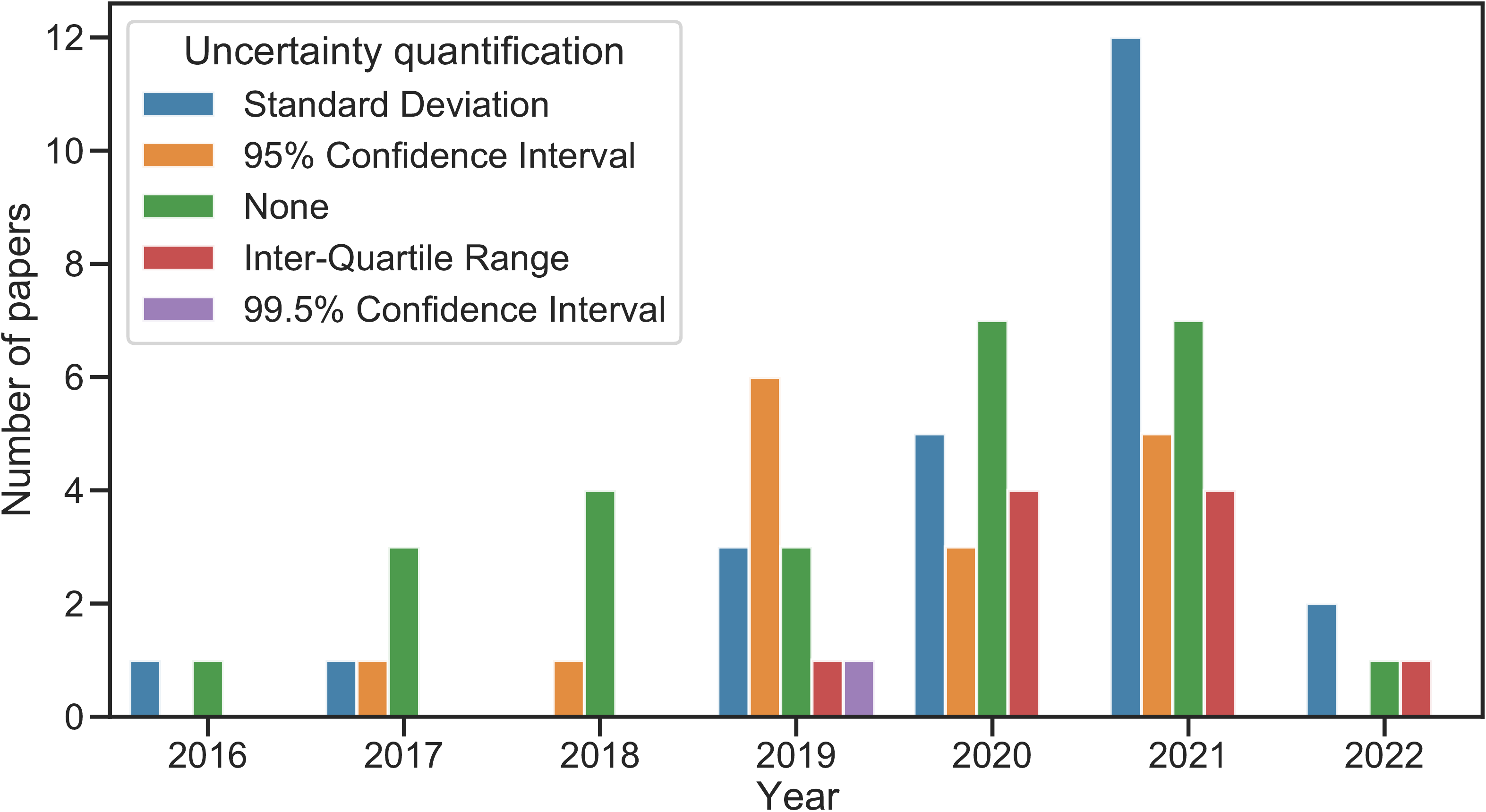}
    \end{subfigure}

    \vspace{0.4cm}

    \begin{subfigure}[t]{0.02\textwidth}
        \scriptsize
        \textbf{(d)}
    \end{subfigure}
    \begin{subfigure}[t]{0.29\textwidth}
        \includegraphics[width=\linewidth, valign=t]{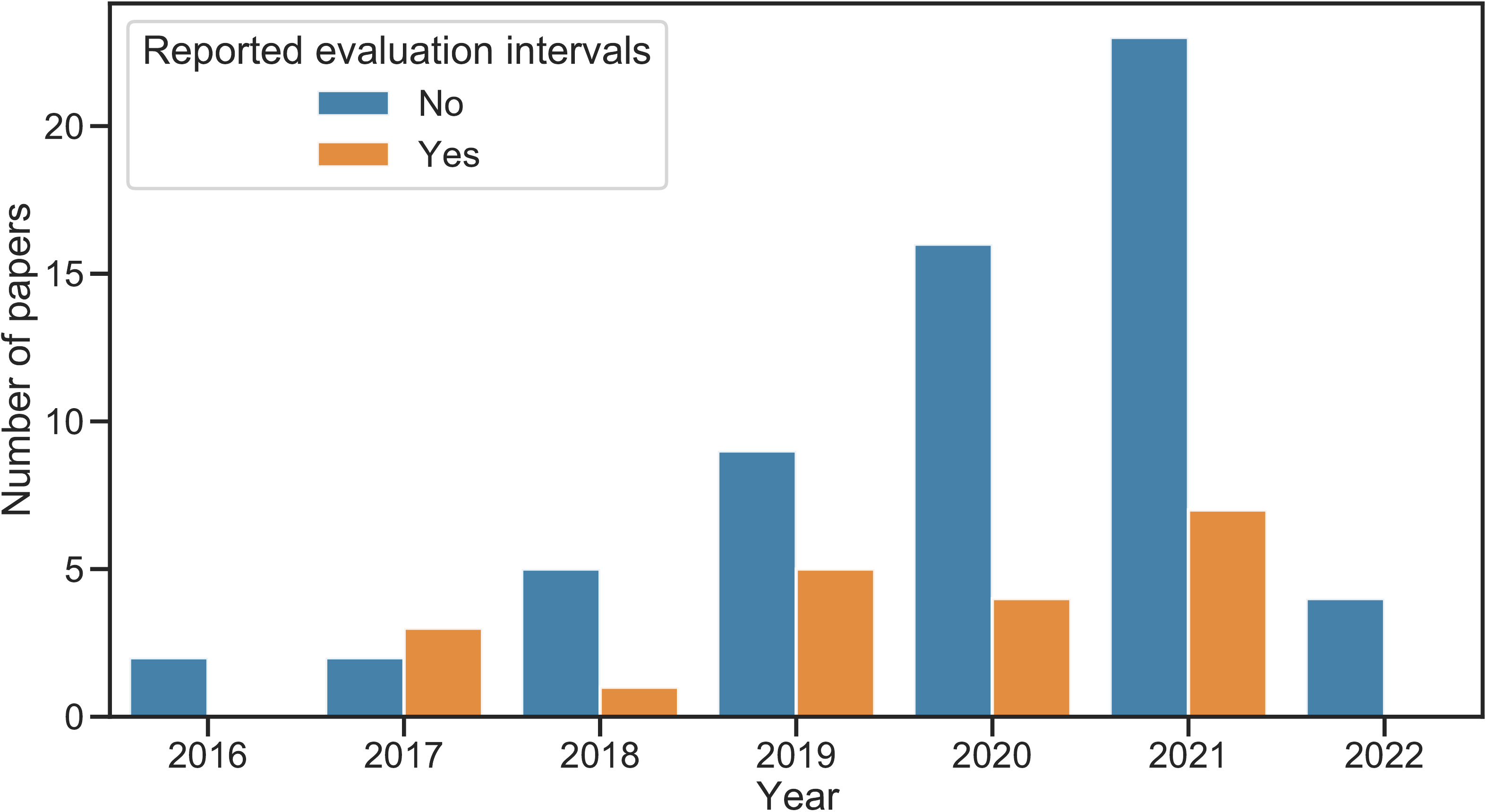}
    \end{subfigure}
    \hspace{1mm}
    \begin{subfigure}[t]{0.02\textwidth}
        \scriptsize
        \textbf{(e)}
    \end{subfigure}
    \begin{subfigure}[t]{0.29\textwidth}
        \includegraphics[width=\linewidth, valign=t]{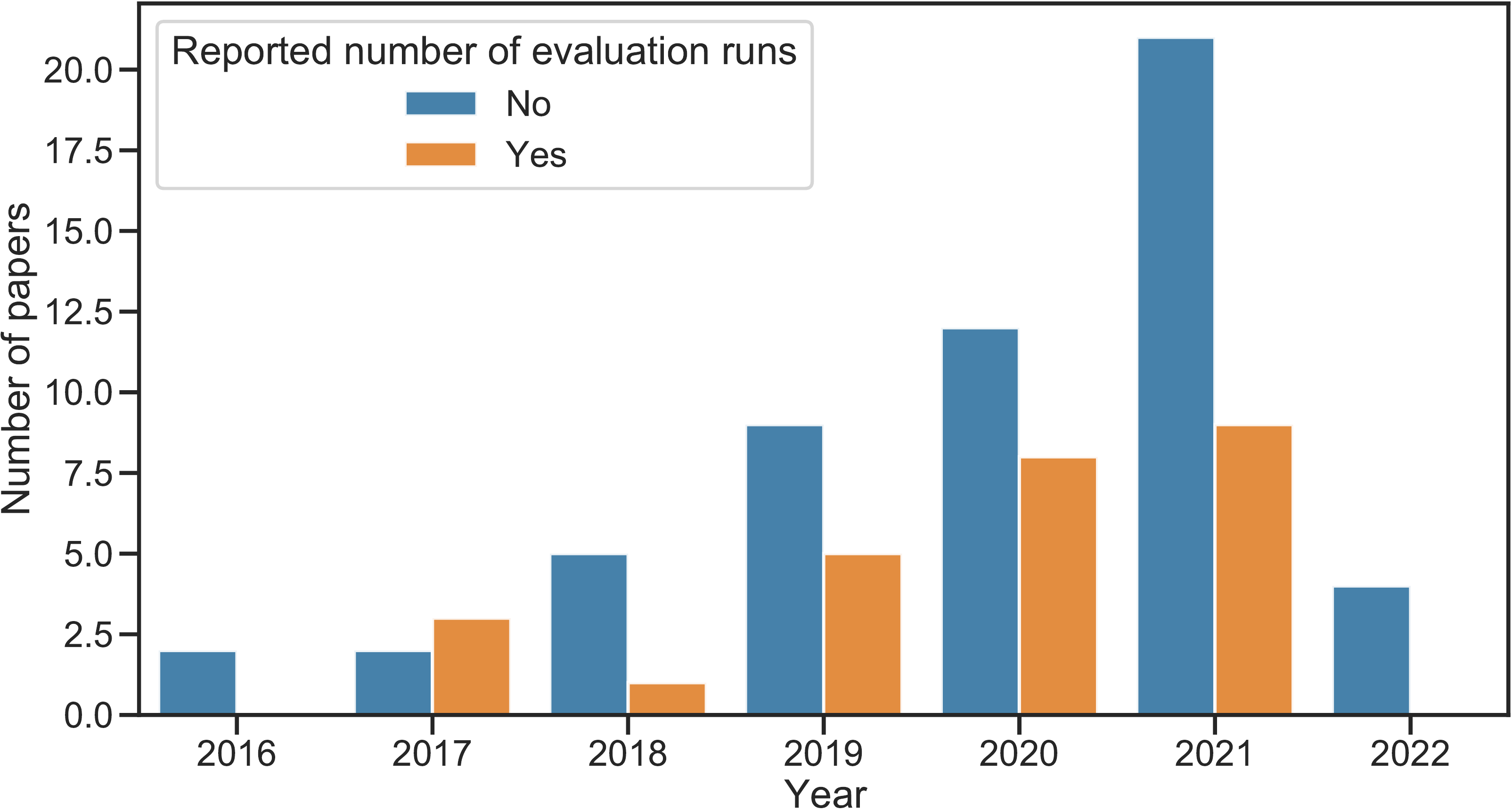}
    \end{subfigure}
    \hspace{1mm}
    \begin{subfigure}[t]{0.02\textwidth}
        \scriptsize
        \textbf{(f)}
    \end{subfigure}
    \begin{subfigure}[t]{0.29\textwidth}
        \includegraphics[width=\linewidth, valign=t]{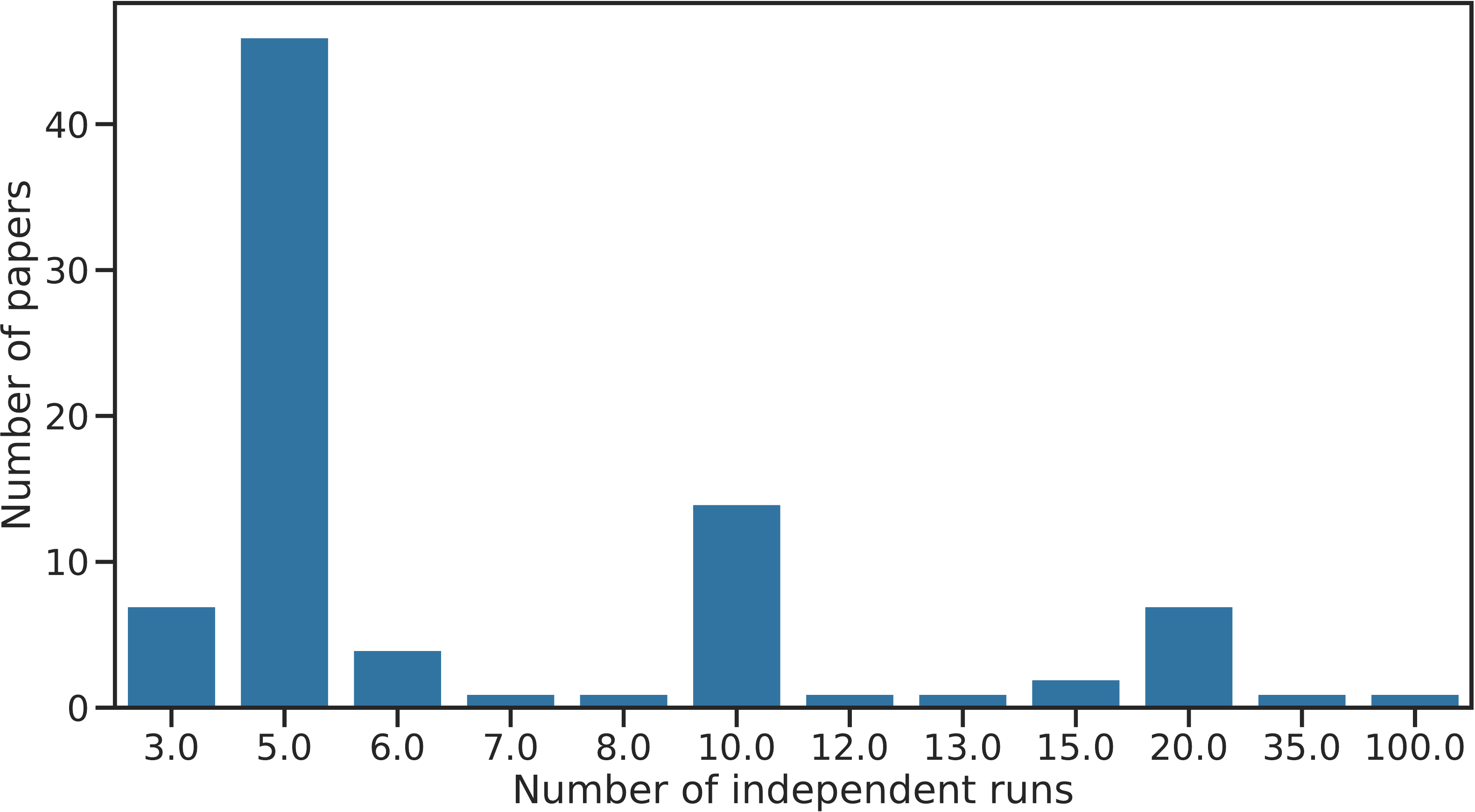}
    \end{subfigure}
    \caption{\textit{Trends in performance aggregation, uncertainty quantification and omissions in reporting on key aspects of evaluation methodology}. \textbf{(a)} Distribution of performance aggregation metrics. None means that the aggregation metric was not specified in the paper. \textbf{(b)} Distribution of uncertainty methods: $\alpha\%$ confidence interval (CI), standard deviation (STD) and interquartile range (IQR). \textbf{(c)} Trends in uncertainty quantification over time. \textbf{(d)} Reporting of the evaluation interval used. \textbf{(e)} Reporting on the number of evaluation runs used. \textbf{(f)} Number of independent runs used in published work.}
    \label{fig: uncertainty estimation}
\end{figure}

\textbf{The situation in MARL} -- \textit{A lack of shared standards for uncertainty estimation and concerning omissions in reporting}: In Figure \ref{fig: uncertainty estimation} (a)-(c), we investigate the use of statistical aggregation and uncertainty quantification methods in MARL. We find considerable variability in the methods used, with little indication of standardisation. Perhaps more concerning is a complete absence of proper uncertainty quantification from one third of published papers. On a more positive note, we observe an upward trend in the use of standard deviation as an uncertainty measure in recent years, particularly in 2021. Furthermore, it has become fairly standard in MARL to evaluate algorithms at regular intervals during the training phase by conducting a certain number of independent evaluation runs in the environment using the current set of learned parameters. This procedure is then followed for each independent training run and results are aggregated to assess final performance. In our analysis, we find that key aspects of this procedure are regularly omitted during reporting, as shown in Figure \ref{fig: uncertainty estimation} (d)-(e). Specifically, in (d), we find many papers omit details on the exact evaluation interval used, and in (e), a similar trend in omission regarding the exact number of independent evaluation runs used. Finally, in Figure \ref{fig: uncertainty estimation} (f), we plot the number of independent runs used during training, showing no clear standard. Given the often high computational requirements of MARL research, it is not surprising that most works opt for a low number of independent training runs, however this remains of concern when making statistical claims. 

\textbf{Recommendations} -- \textit{Standardised statistical tooling and uncertainty estimation including detailed reporting}: As mentioned before, the computational requirements in MARL research often make it prohibitively difficult to run many independent experiments to properly quantify uncertainty. One approach to make sound statistical analysis more tractable, is to pool results across different tasks using the bootstrap \citep{Efron1992}. In particular, for RL, \cite{agarwal2022deep} recommend computing stratified bootstrap confidence intervals, where instead of only using the original set of data to calculate confidence intervals, the data is resampled with replacement from $M$ tasks, each having $N$ runs. This process is repeated as many times as needed to approximate the sampling distribution of the statistic being calculated. Furthermore, when making a summary of overall performance across tasks it has been shown that the mean and median are insufficient, the former being dominated by outliers and the latter having higher variance. Instead, \cite{agarwal2022deep} propose the use of the interquartile mean (IQM) which is more robust to outlier scores than the mean and more statistically efficient than the median. Finally, \cite{agarwal2022deep} propose the use of probability of improvement scores, sample efficiency curves and performance profiles, which are commonly used to compare the performance of optimization algorithms \citep{performance_profiles}. These performance profiles are inherently robust to outliers on both ends of the distribution tails and allow for the comparison of relative performance at a glance. In the shared reward setting, where it is only required to track a single return value, we argue that these tools fit the exact needs of cooperative MARL, as they do in RL. Furthermore, in light of our analysis, we strongly recommend a universal standard in the use, and reporting of, evaluation parameters such as the number of independent runs, evaluation frequency, performance metrics and statistical tooling, to make comparisons across different works easier and more fair. 


\begin{figure}
    \centering
    \begin{subfigure}[t]{0.02\textwidth}
        \scriptsize
        \textbf{(a)}
    \end{subfigure}
    \begin{subfigure}[t]{0.6\textwidth}
        \includegraphics[width=\linewidth, valign=t]{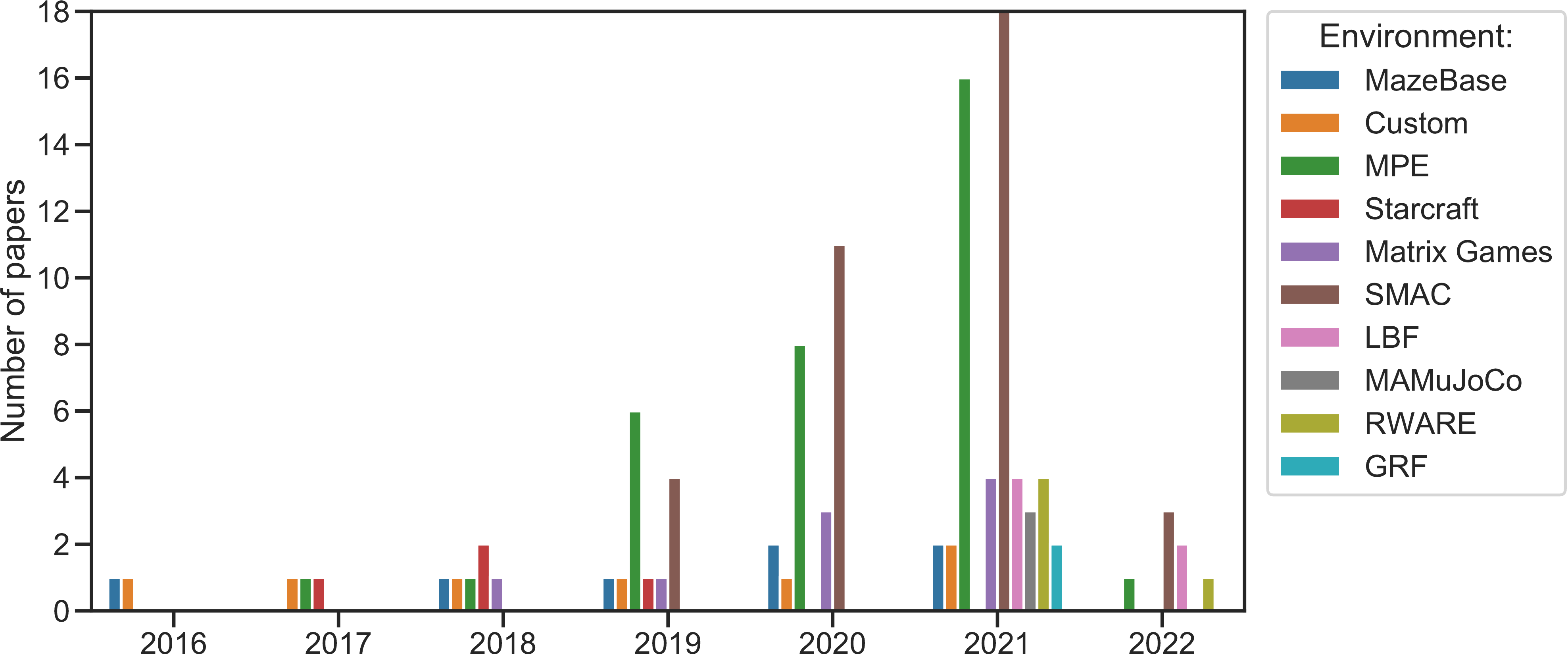}
    \end{subfigure}
    \hspace{1mm}
    \begin{subfigure}[t]{0.02\textwidth}
        \scriptsize
        \textbf{(b)}
    \end{subfigure}
    \begin{subfigure}[t]{0.28\textwidth}
        \includegraphics[width=\linewidth, valign=t]{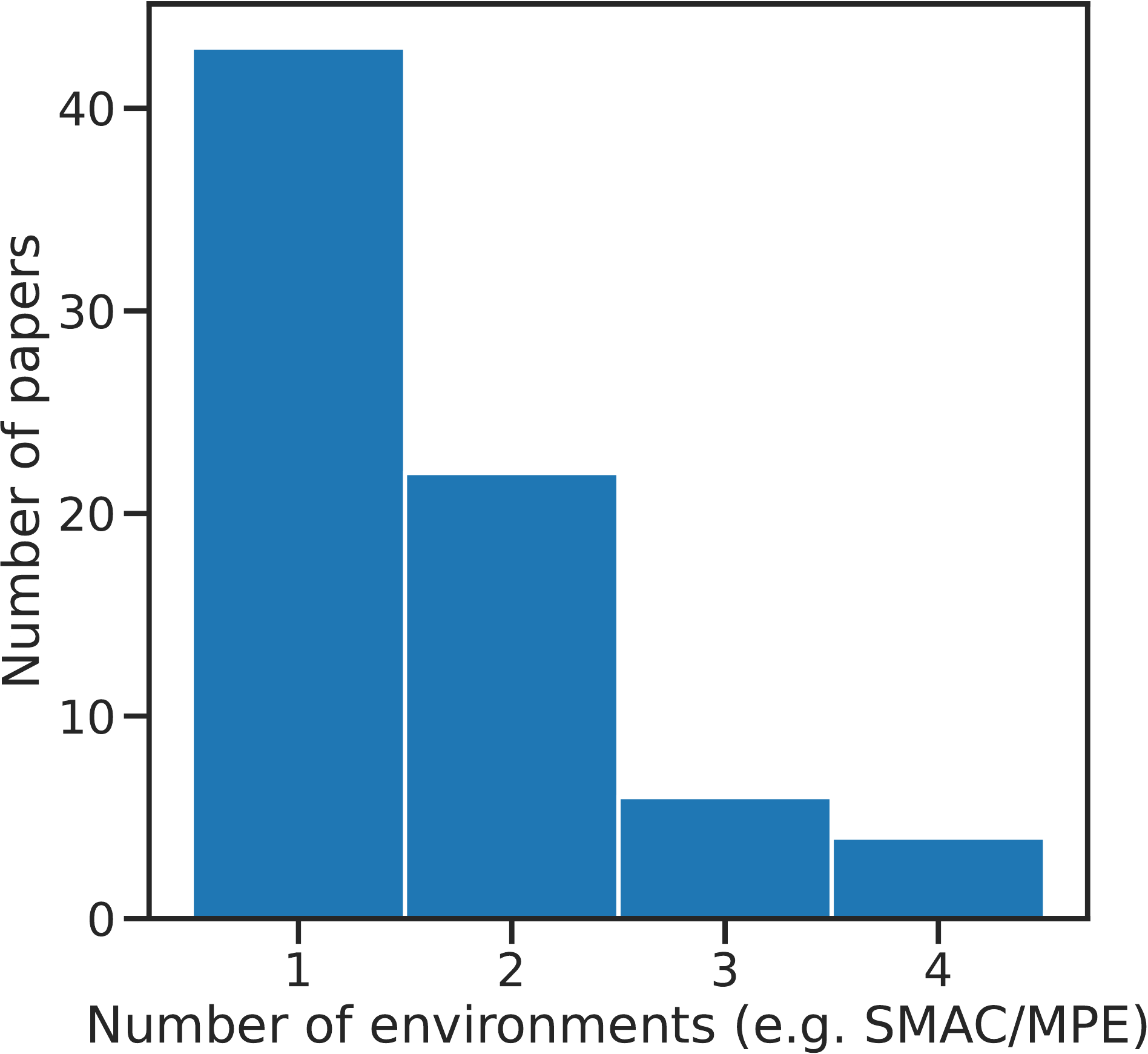}
    \end{subfigure}

    \vspace{0.4cm}

    \begin{subfigure}[t]{0.02\textwidth}
        \scriptsize
        \textbf{(c)}
    \end{subfigure}
    \begin{subfigure}[t]{0.22\textwidth}
        \includegraphics[width=\linewidth, valign=t]{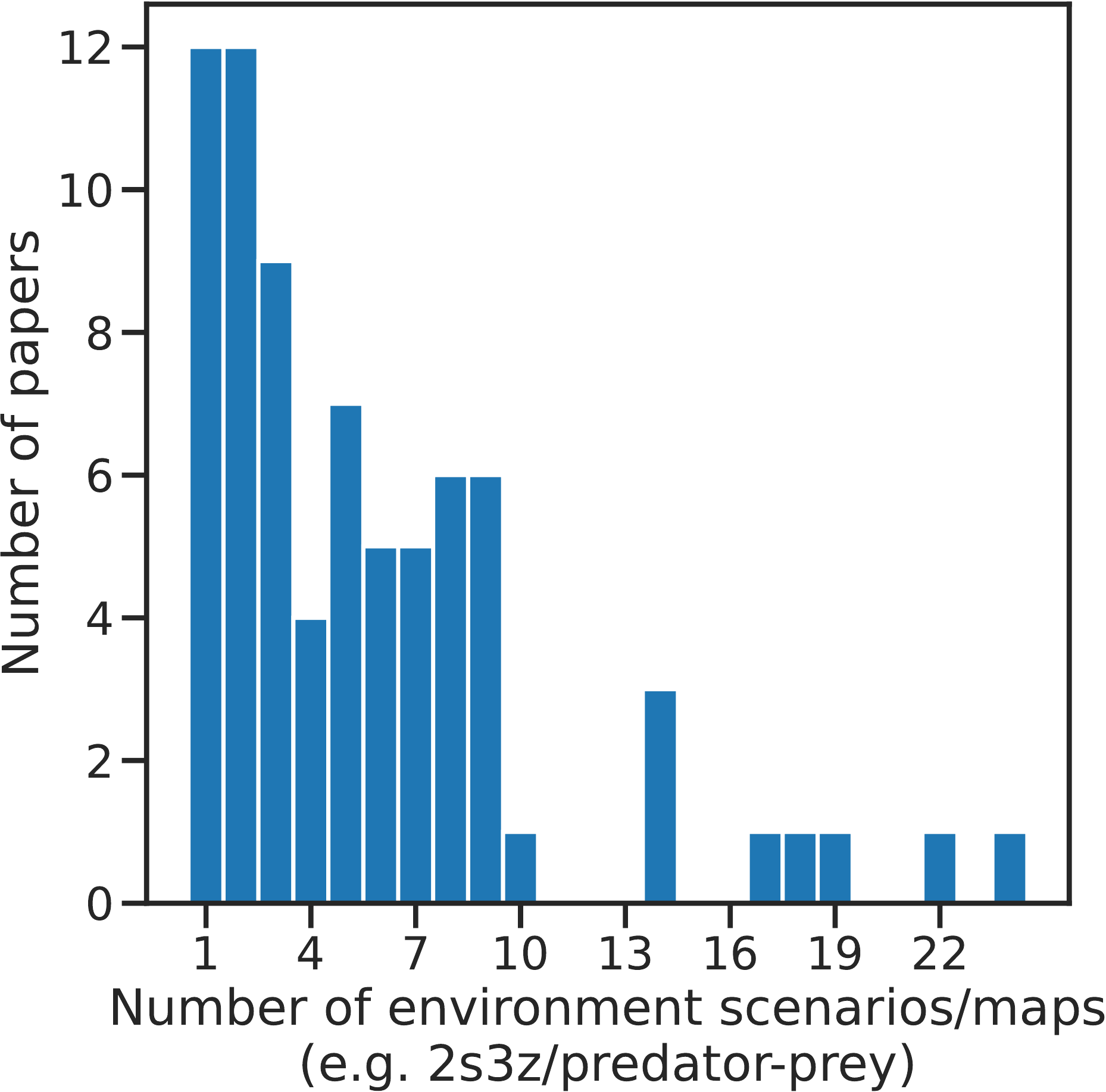}
    \end{subfigure}
    \hspace{1mm}
    \begin{subfigure}[t]{0.02\textwidth}
        \scriptsize
        \textbf{(d)}
    \end{subfigure}
    \begin{subfigure}[t]{0.24\textwidth}
        \includegraphics[width=\linewidth, valign=t]{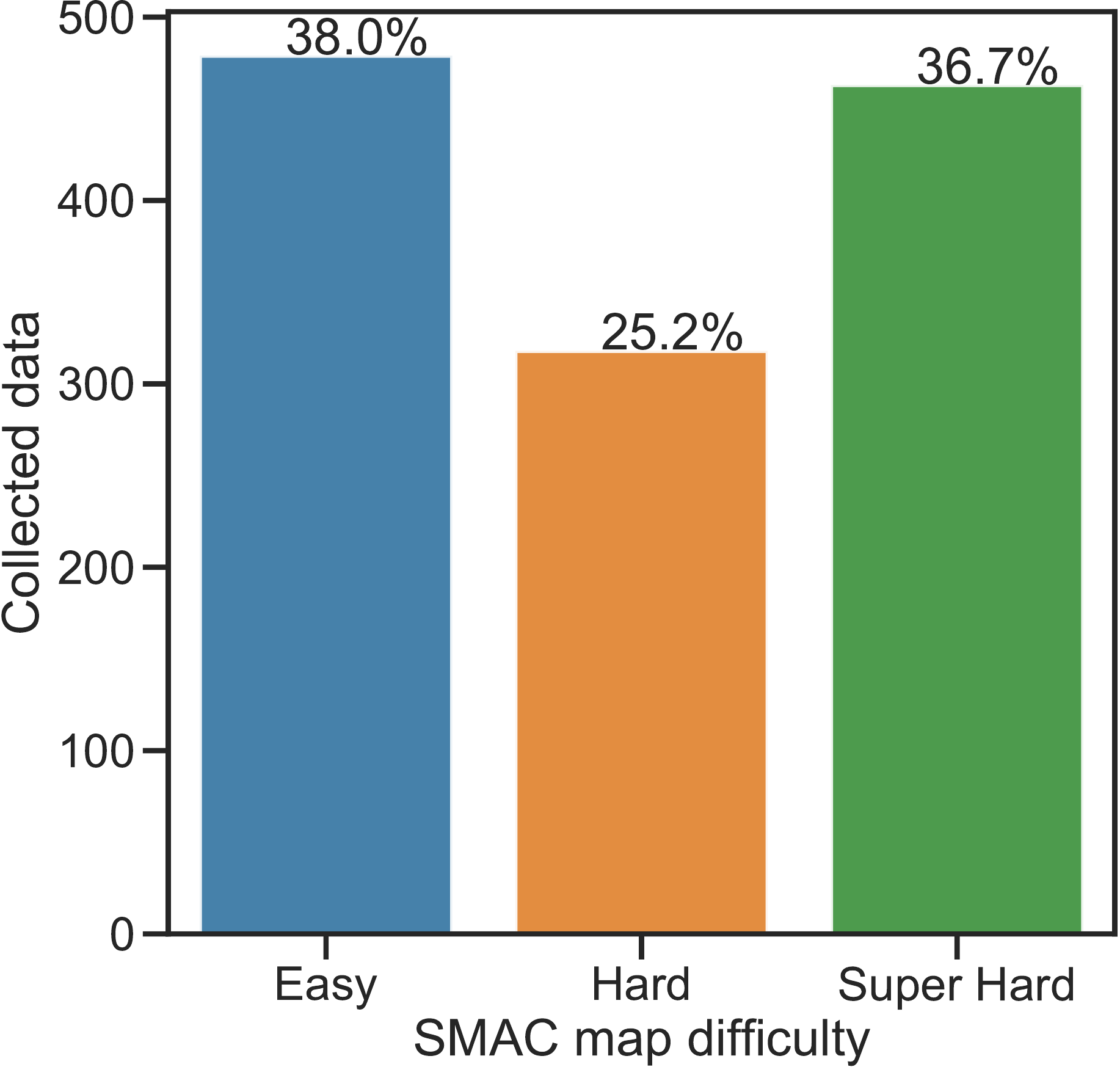}
    \end{subfigure}
    \hspace{1mm}
    \begin{subfigure}[t]{0.02\textwidth}
        \scriptsize
        \textbf{(e)}
    \end{subfigure}
    \begin{subfigure}[t]{0.4\textwidth}
        \includegraphics[width=\linewidth, valign=t]{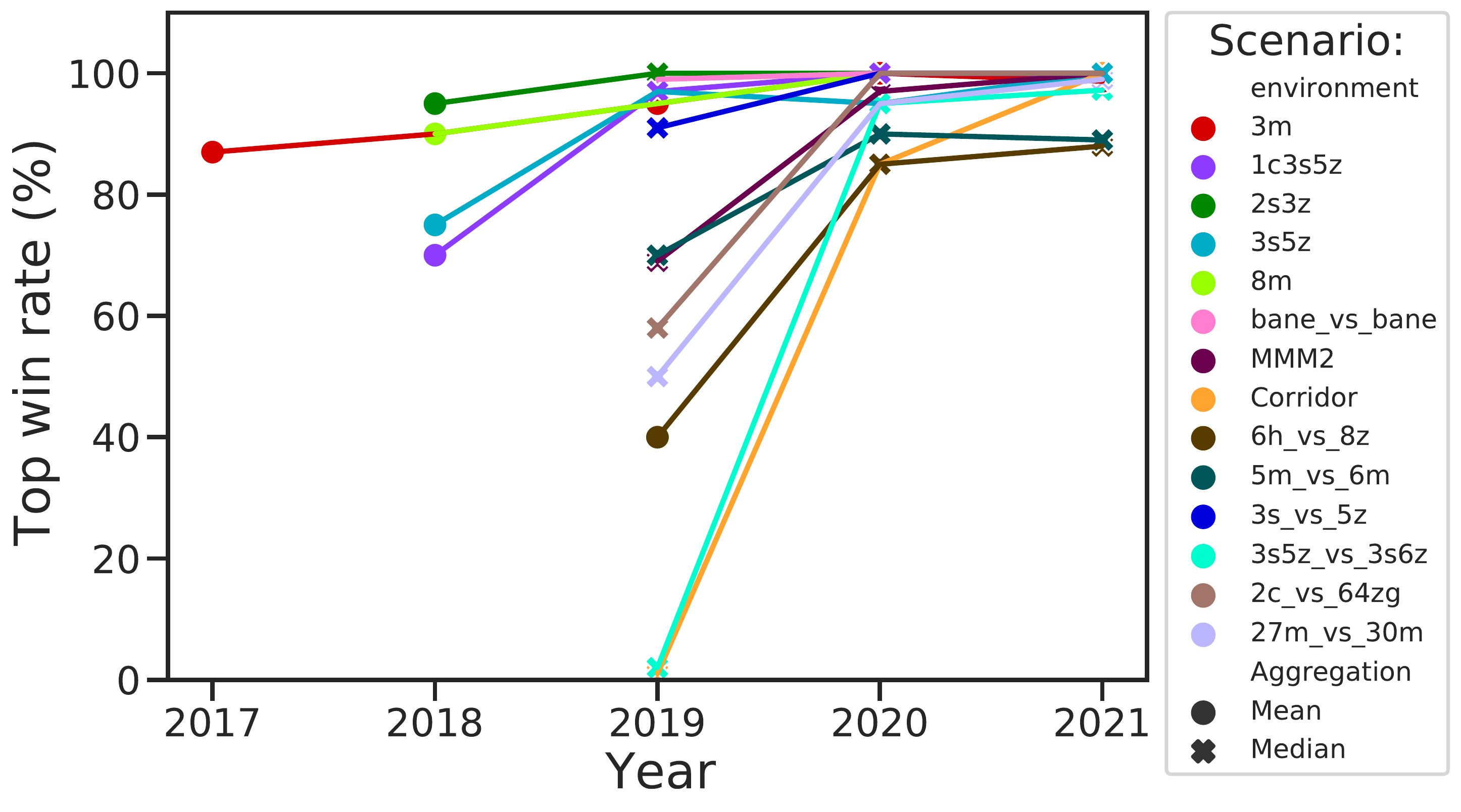}
    \end{subfigure}
    \caption{\textit{Environment popularity, usage trends in papers and potential evidence of overfitting on SMAC.} \textbf{(a)} Environment adoption over time. \textbf{(b)} Number of environments used in papers. \textbf{(c)} Number of scenarios/tasks/maps used in papers. \textbf{(d)} Distribution of task difficulty of SMAC maps used in papers. \textbf{(e)} Performance trends on popular SMAC maps: Aggregation is the aggregate function used for the different reported values.}
    \label{fig: smac env}
\end{figure}

\subsection{Lesson 3: Guard against environment misuse and overfitting}

\textbf{Issue in RL} -- \textit{Over-tuning algorithms to perform well on a specific environment}: As early as 2011 issues with evaluation in RL came to the foreground in the form of \textit{environment overfitting}. \cite{whiteson2011protecting} raised the concern that in the context of RL, researchers might over-tune algorithms to perform well on a specific benchmark at the cost of all other potential use cases. More specifically, \cite{whiteson2011protecting} define environment overfitting in terms of a desired target distribution. When an algorithm performs well in a specific environment but lacks performance over a target \textit{distribution} of environments, it can be deemed to have overfit that particular environment.

\textbf{The situation in MARL} -- \textit{One environment to rule them all -- on the use and misuse of SMAC}: The StarCraft multi-agent challenge (SMAC) \citep{samvelyan19smac} has quickly risen to prominence as a key benchmark for MARL research since it's release in 2019, rivaled only in use by the multi-agent particle environment (MPE) introduced by \cite{lowe2017multi} (see Figure \ref{fig: smac env} (a)). SMAC and its accompanied MARL framework PyMARL (introduced in the same paper), fulfills several desirable properties for benchmarking: offering multiple maps of various difficulty that test key aspects of MARL algorithms and providing a unified API for running baselines as well as state-of-the-art algorithms on SMAC. Unfortunately, the wide-spread adoption of SMAC has also caused several issues, relating to environment overfitting and cherry picking of results, putting into question the credibility of claims made while using it as a benchmark. 

To illustrate the above point, we start by highlighting that many MARL papers use only a single environment (e.g. SMAC or MPE) for evaluation, as shown in Figure \ref{fig: smac env} (b). This is often deemed acceptable since both SMAC and MPE provide many different tasks, or maps. For instance, in SMAC, there are 23 different maps providing a wide variety in terms of the number of agents, agent types and game dynamics. However, there is no standard, or agreed upon set of maps to use for benchmarking novel work, which makes it easy for authors to selectively subsample maps post-experiment based on the outcomes of their proposed algorithm. As shown in Figure \ref{fig: smac env} (c), although environments like SMAC and MPE offer many different testing scenarios, it is typical for papers to only use a small number of these in their reported experiments.

\begin{figure}
    \centering
    \includegraphics[width=0.34\textwidth]{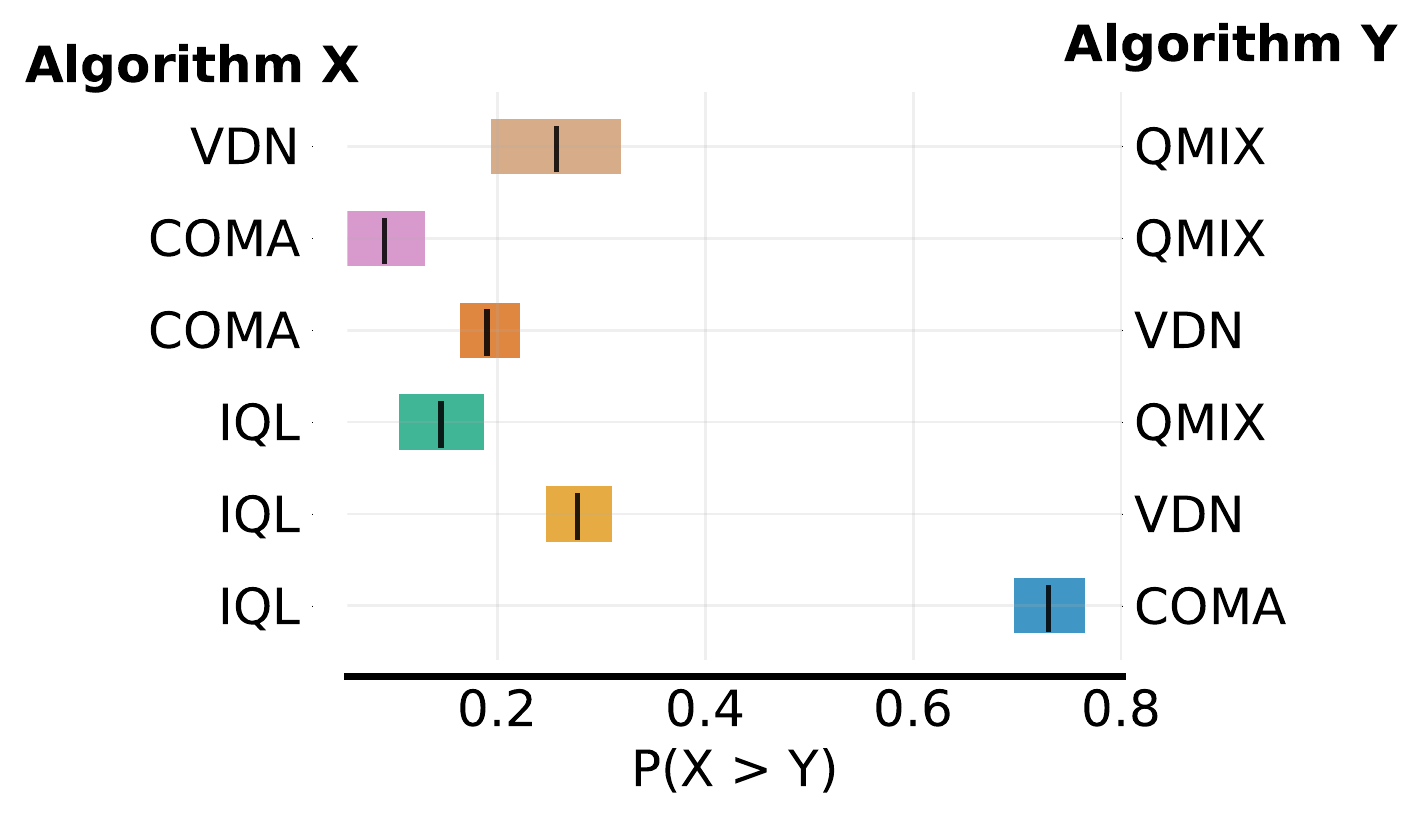}
    \hspace{1mm}
    \includegraphics[width=0.33\textwidth]{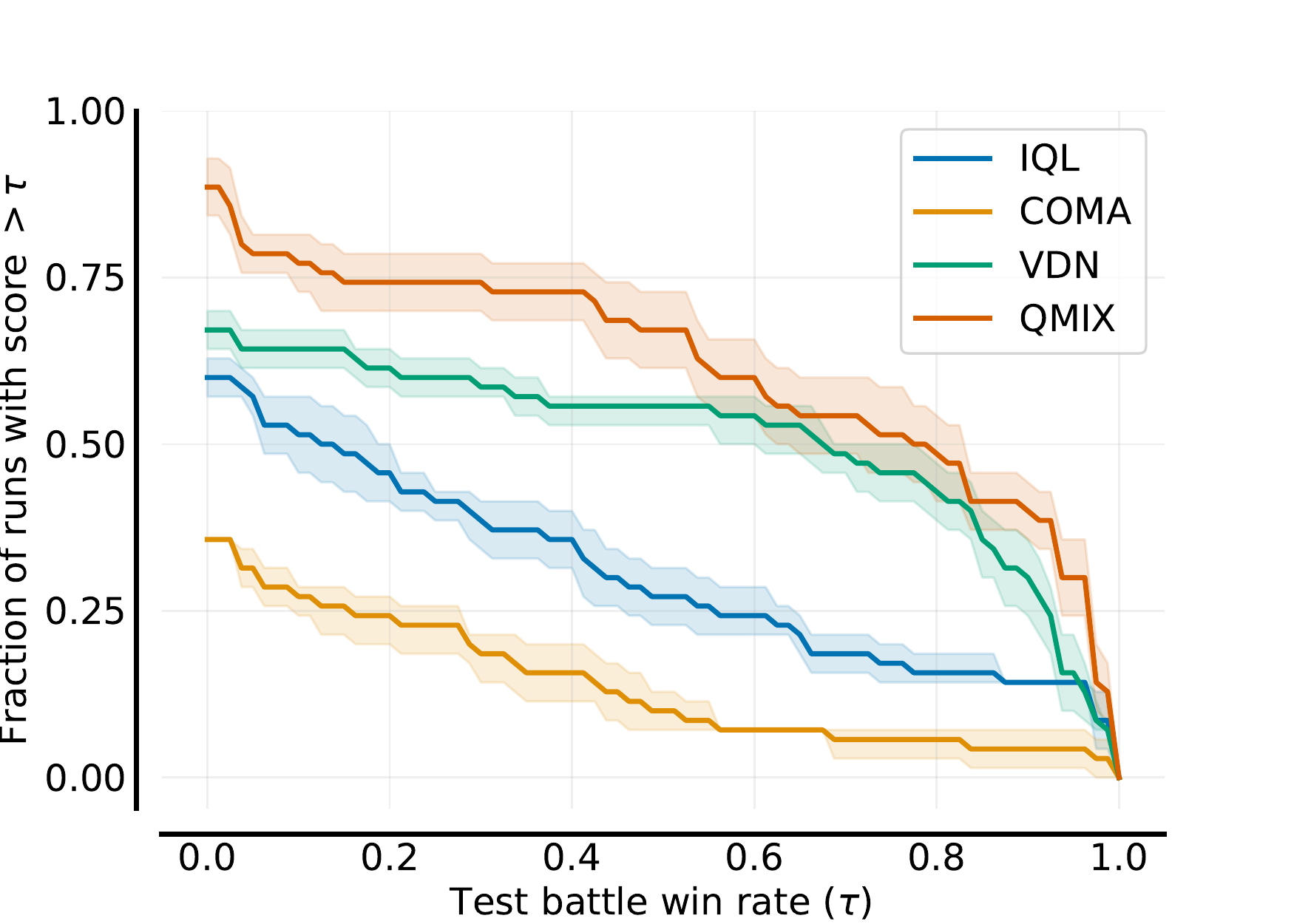}
    \includegraphics[width=0.29\textwidth]{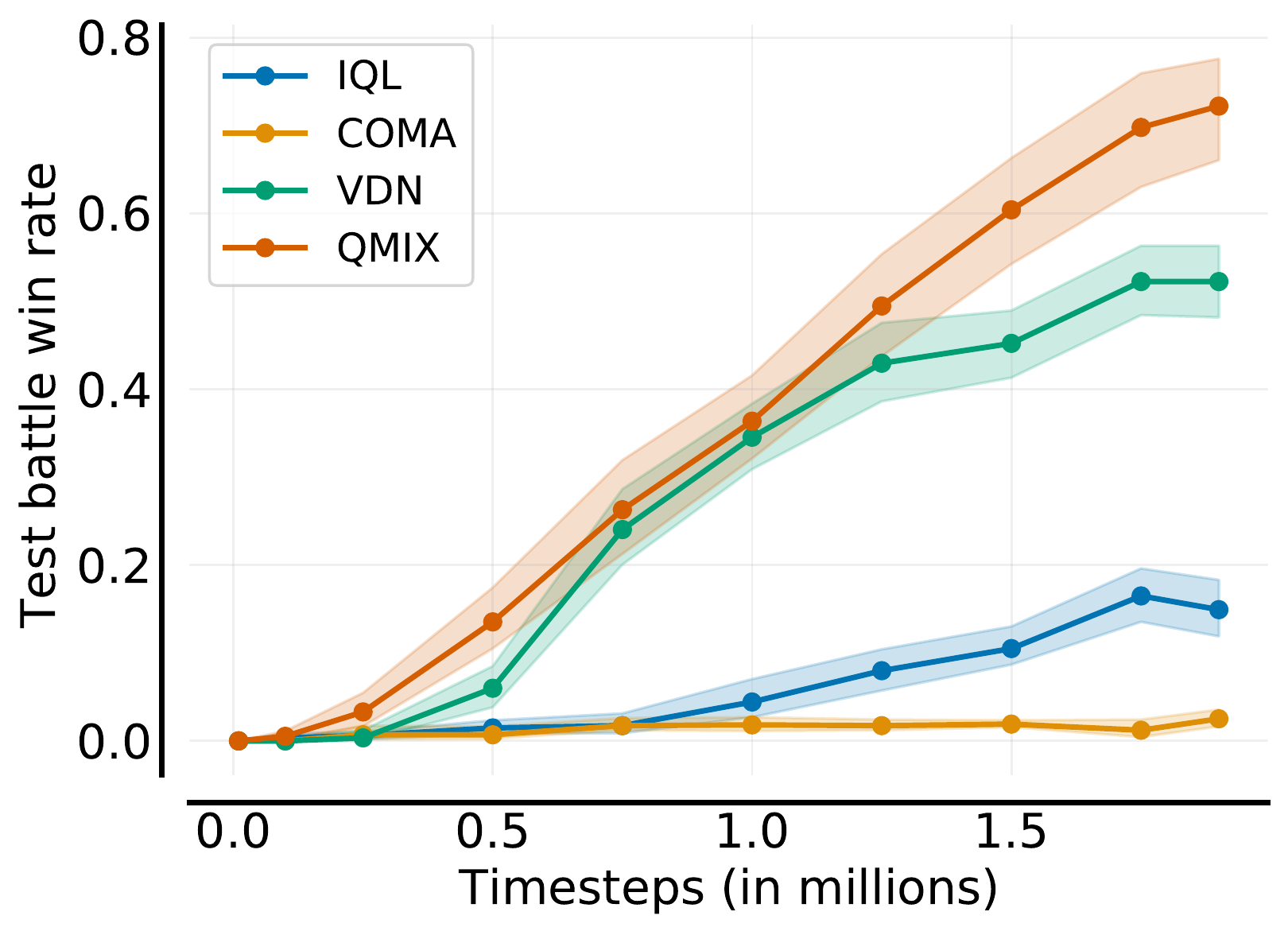}
    \includegraphics[width=0.34\textwidth]{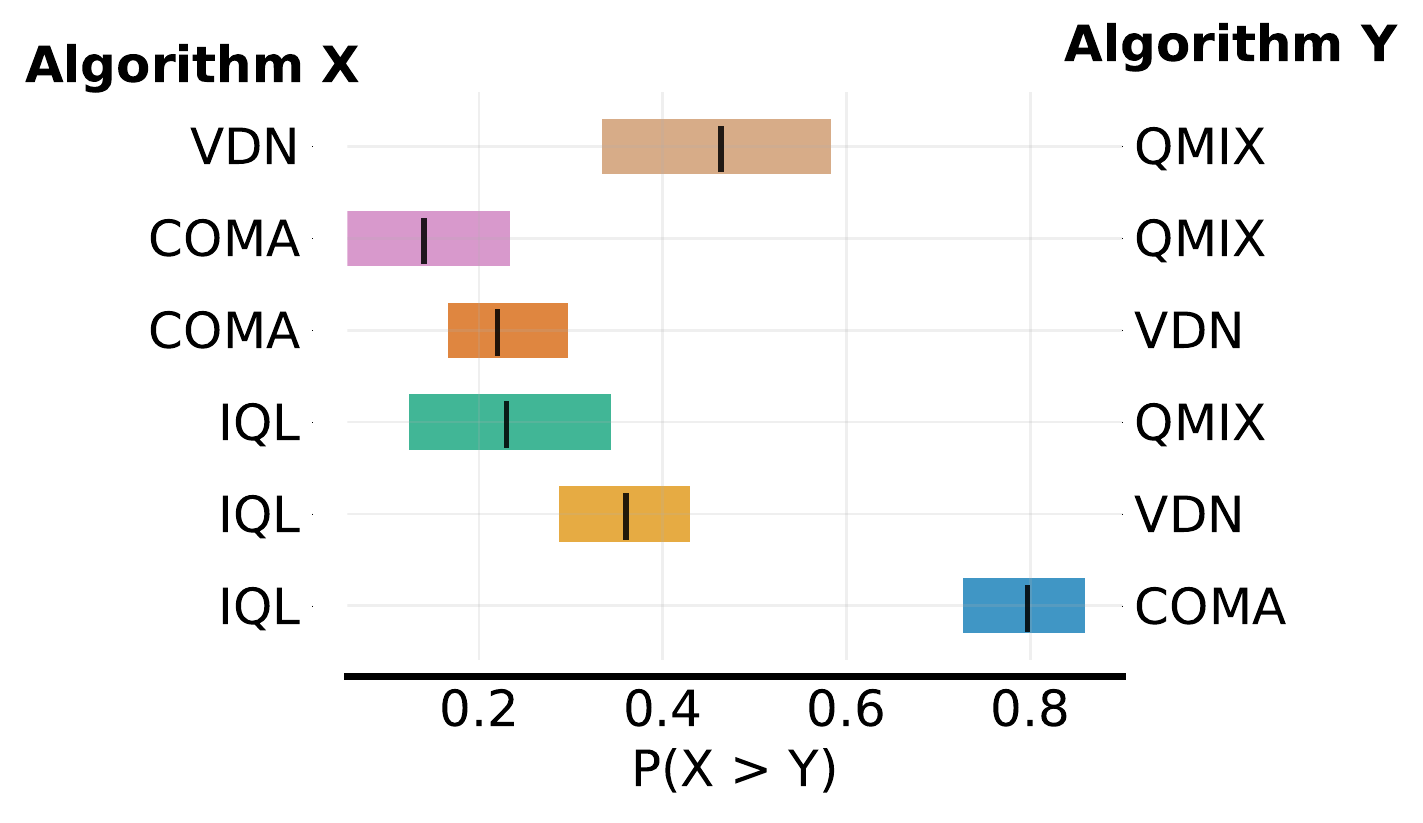}
    \hspace{1mm}
    \includegraphics[width=0.33\textwidth]{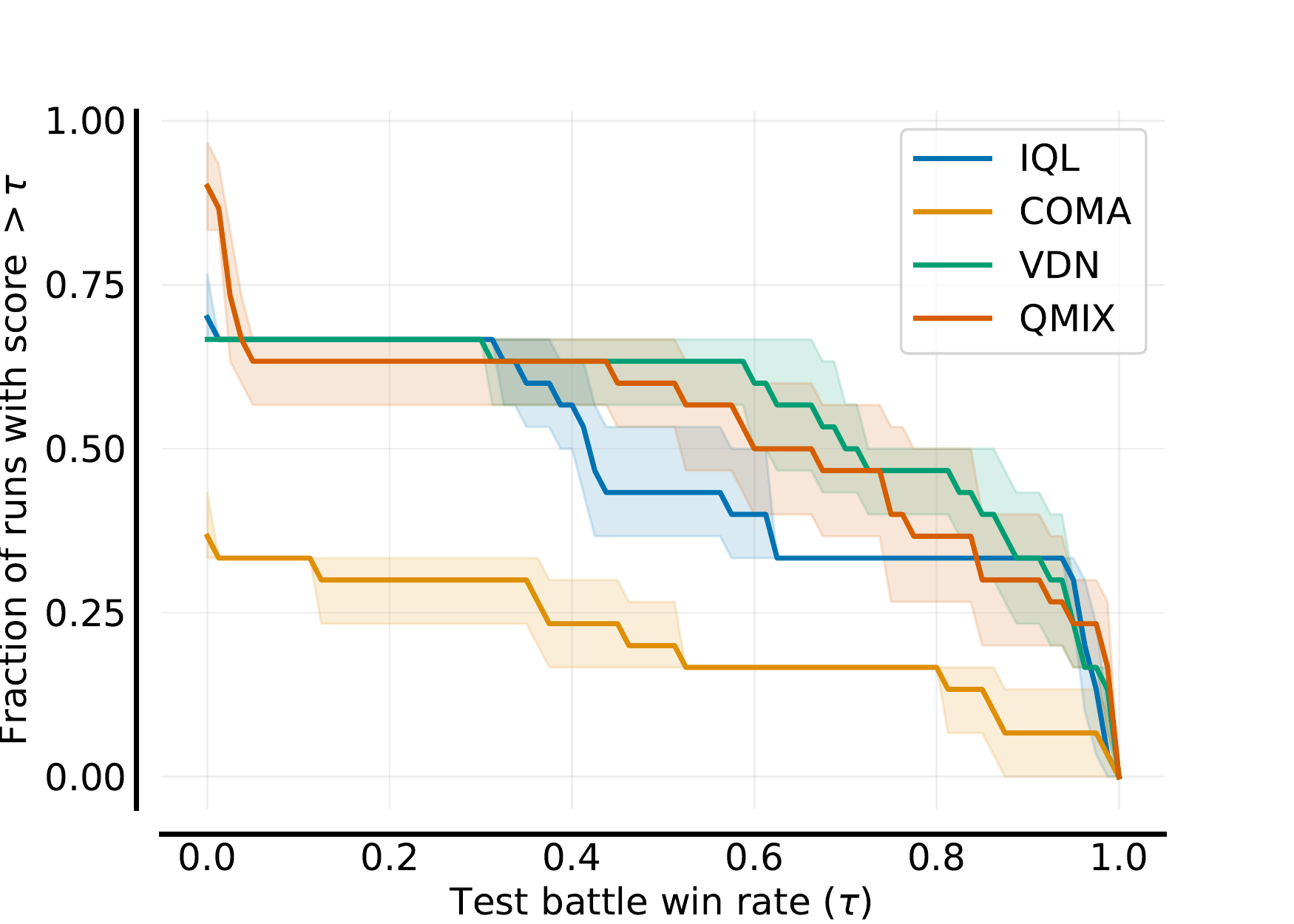}
    \includegraphics[width=0.29\textwidth]{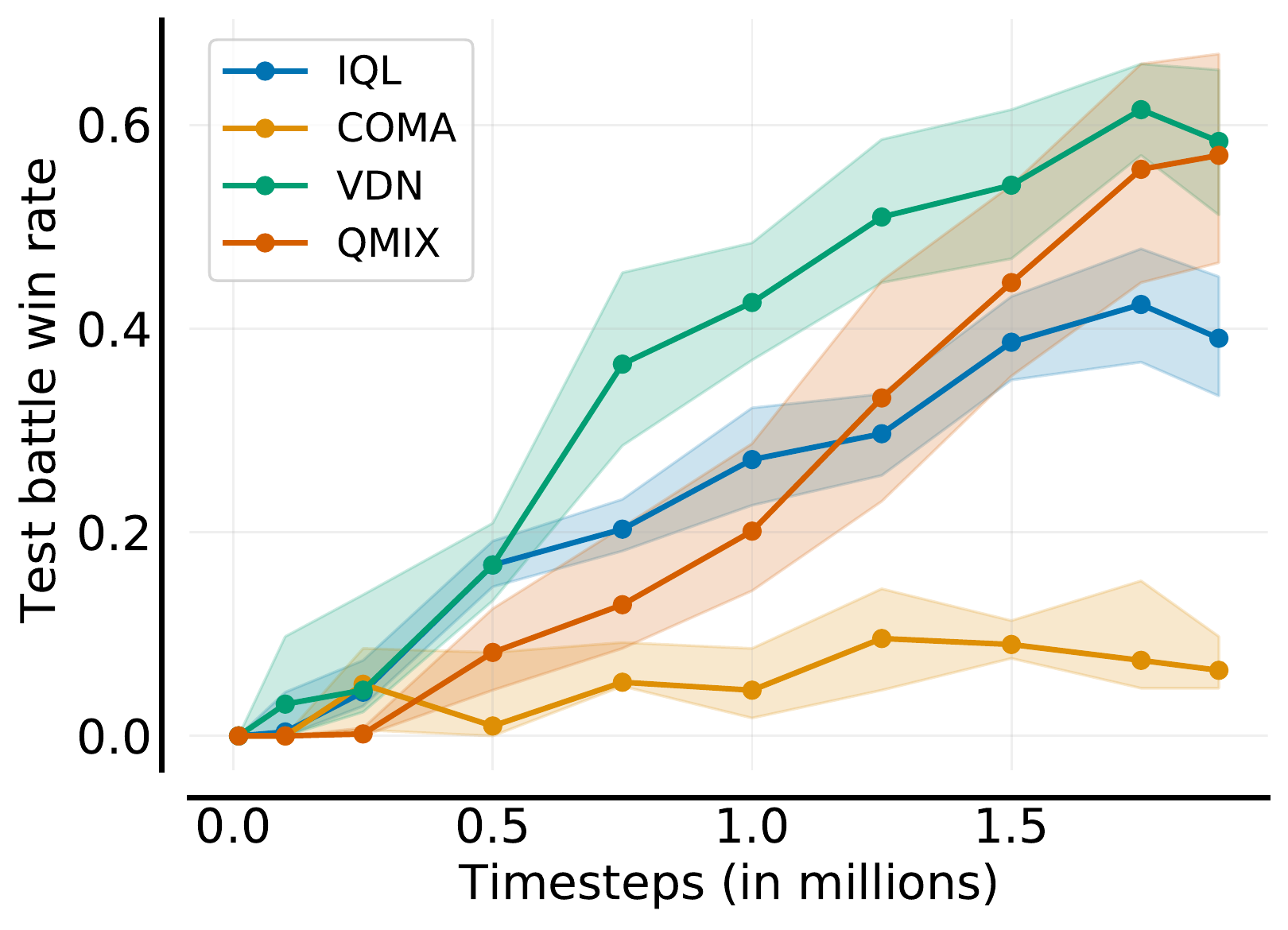}
    \caption{\textit{Reanalysis of original SMAC experiments conducted by \cite{samvelyan19smac} including probability of improvement, performance profiles and sample efficiency curves}. \textbf{Top row:} All 14 SMAC maps used in the original analysis. \textbf{Bottom row:} Subset of 6 maps, including 2 easy, 2 medium and 2 hard: 2s\_vs\_1sc, 3s\_vs\_5z, bane\_vs\_bane, 5m\_vs\_6m, 6h\_vs\_8z and corridor}
    \label{fig: smac re-analysis}
\end{figure}

To concretely expose the potential danger in this author map selection bias, we redo the original analysis performed by \cite{samvelyan19smac}, using the authors' exact experimental data that was made publicly available\footnote{We applaud the authors for making their raw evaluation data publicly available. This data can be found here: \href{https://github.com/oxwhirl/smac}{https://github.com/oxwhirl/smac}.}, containing five independent runs for IQL \citep{tampuu2015multiagent}, COMA \citep{foerster2018counterfactual}, VDN \citep{sunehag2017value} and QMIX \citep{rashid2018qmix} on 14 SMAC maps.\footnote{The maps chosen were: 1c3s5z, 2c\_vs\_64zg, bane\_vs\_bane, MMM2, 10m\_vs\_11m, 27m\_vs\_30m, 5m\_vs\_6m, 2s3z, 2s\_vs\_1sc, s5z, 3s\_vs\_5z, 6h\_vs\_8z, 3s5z\_vs\_3s6z and corridor} The top row of Figure \ref{fig: smac re-analysis} shows the results of this analysis performed using the statistical tools recommended by \cite{agarwal2022deep}, including the probability of improvement between algorithms, performance profiles and sample efficiency curves. The results support the original claims made by \cite{samvelyan19smac}, namely that QMIX is a superior algorithm to that of VDN, COMA and IQL, both in terms of performance and sample efficiency. However, by simply sampling a smaller set of two easy, medium and hard maps (a common spread in the literature, see Figure \ref{fig: smac env} (d)), from the original 14, giving 6 maps in total (a reasonable number according to prior work, see Figure \ref{fig: smac env} (c)), we are able to change the outcome of the analysis in support of no difference in performance between VDN and QMIX, as well as finding VDN to be more sample efficient. This is shown in the bottom row of Figure \ref{fig: smac re-analysis} and highlights the danger of a lack of standards regarding which \textit{fixed} set of maps should be used for benchmarking.

We end our investigation into SMAC (and refer the reader to the Appendix for additional discrepancies uncovered), by looking at historical performance trends. In Figure \ref{fig: smac env} (e), we show the top win rate achieved by an algorithm in a specific year for 14 of the most popular maps used in prior work. We find that by 2021, most of these maps have converged to a win rate close or equal to 100\%, while only a few maps are still situated around 80-90\%. Given that many of these maps repeatedly feature across papers, and will likely be used in future work, it begs the question to what extent the MARL community has already overfit to SMAC as an evaluation benchmark. 

\textbf{Recommendations} -- \textit{Standardised environment sets and testing for generalisation}: To solve the issue of environment overfitting, \cite{whiteson2011protecting} propose the use of a generalised evaluation methodology. In this approach, environments (for tuning algorithms) are freely sampled from some generalised environment set. Separately, algorithm evaluation is performed on a second set of sampled environments from the same generalised environment set, acting as a test set analogous to that used in supervised learning. Recent work in this direction include benchmarks such as Procgen \citep{cobbe2020leveraging}, which use procedural generation to implicitly construct a distribution from which to sample test tasks. In SMAC and other environments, it is common practice to only evaluate on the exact map the algorithm was trained on, under that exact same conditions, and to not specifically test for generalisation across unseen tasks. However, many MARL algorithms are still highly sensitive to small changes in the environment and often fail to generalise to new unseen tasks \citep{carion2019structured, zhang2020multi, mahajan2022generalization}. This calls for more work on MARL generalisation and we recommend a stronger focus on benchmarks designed specifically to test generaliation. However, it has been noted that procedurally generated benchmarks may reduce the precision of research \citep{kirk2021survey}, making it more difficult to track progress. Furthermore, MARL exibits several unique and challenging difficulties when it comes to building algorithms able to generalise, likely requiring many years of future work to surmount \citep{mahajan2022generalization}. Therefore, in certain cases, it might make more sense for researchers to take smaller and more precise steps towards key innovations in algorithm design by still relying on traditional environment sets. In this setting, we strongly advocate using fixed environment sets, where ideally these are selected by the designers of each environment and are accompanied by exact instruction for their configuration, so as to be consistent across papers.
\newpage
\section{Towards a standardised evaluation protocol for MARL} \label{sec: eval protocol}

In this section, we pool together the observations and recommendations from the previous section to provide a standardised performance evaluation protocol for cooperative MARL. We are realistic in our efforts, knowing that a single protocol is unlikely to be applicable to all MARL research. However, echoing recent work on evaluation \citep{ulmer2022experimental}, we stress that many of the issues highlighted previously stem from a lack of \emph{standardisation}. Therefore, we believe a default "off-the-shelf" protocol that is able capture most settings, could provide great value to the community. If widely adopted, such a standardised protocol would make it easier and more accurate to compare across different works and remove some of the noise in the signal regarding the true rate of progress in MARL research. A summarised version of our protocol is given in the blue box at the end of this section and a concrete demonstration of its usage can be found in the Appendix.

\textbf{Benchmarks and Baselines. } Before giving details on the proposed protocol, we first briefly comment on benchmarks and baselines used in experiments. These choices often depend on the research question of interest, the novel work being proposed and key algorithmic capabilities to be tested. However, as alluded to in our analysis, we recommend that environment designers take full ownership regarding how their environments are to be used for evaluation. For example, if an environment has several available static tasks, the designers should specify a fixed compulsory minimum set for experiments to avoid biased subsampling by authors. It could also be helpful if designers keep track of state-of-the-art (SOTA) performances on tasks from published works and allow authors to submit these for vetting. We also strongly recommend using more than a single environment (e.g. SMAC) and preferring environments that test generalisation. Regarding baselines, we recommend that at minimum the published SOTA contender to novel work should be included. For example, if the novel proposal is a value-based off-policy algorithm for discrete environments, at minimum, it must be compared to the current SOTA value-based off-policy algorithm for discrete environments. Finally, all baselines must be tuned fairly with the same compute budget.

\begin{wrapfigure}{r}{0.5\textwidth}
    \centering
    \includegraphics[width=0.45\textwidth]{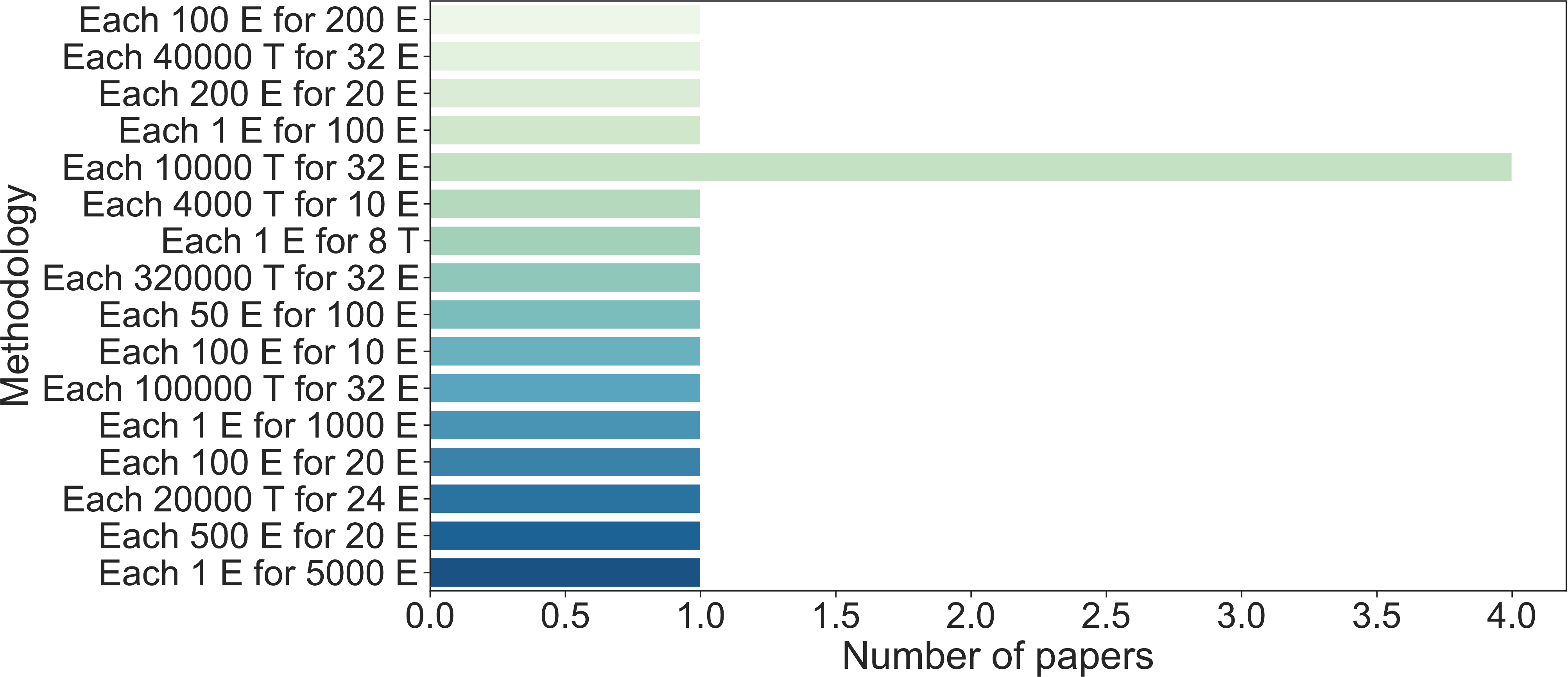}

    \vspace{0.4cm}

    \includegraphics[width=0.22\textwidth]{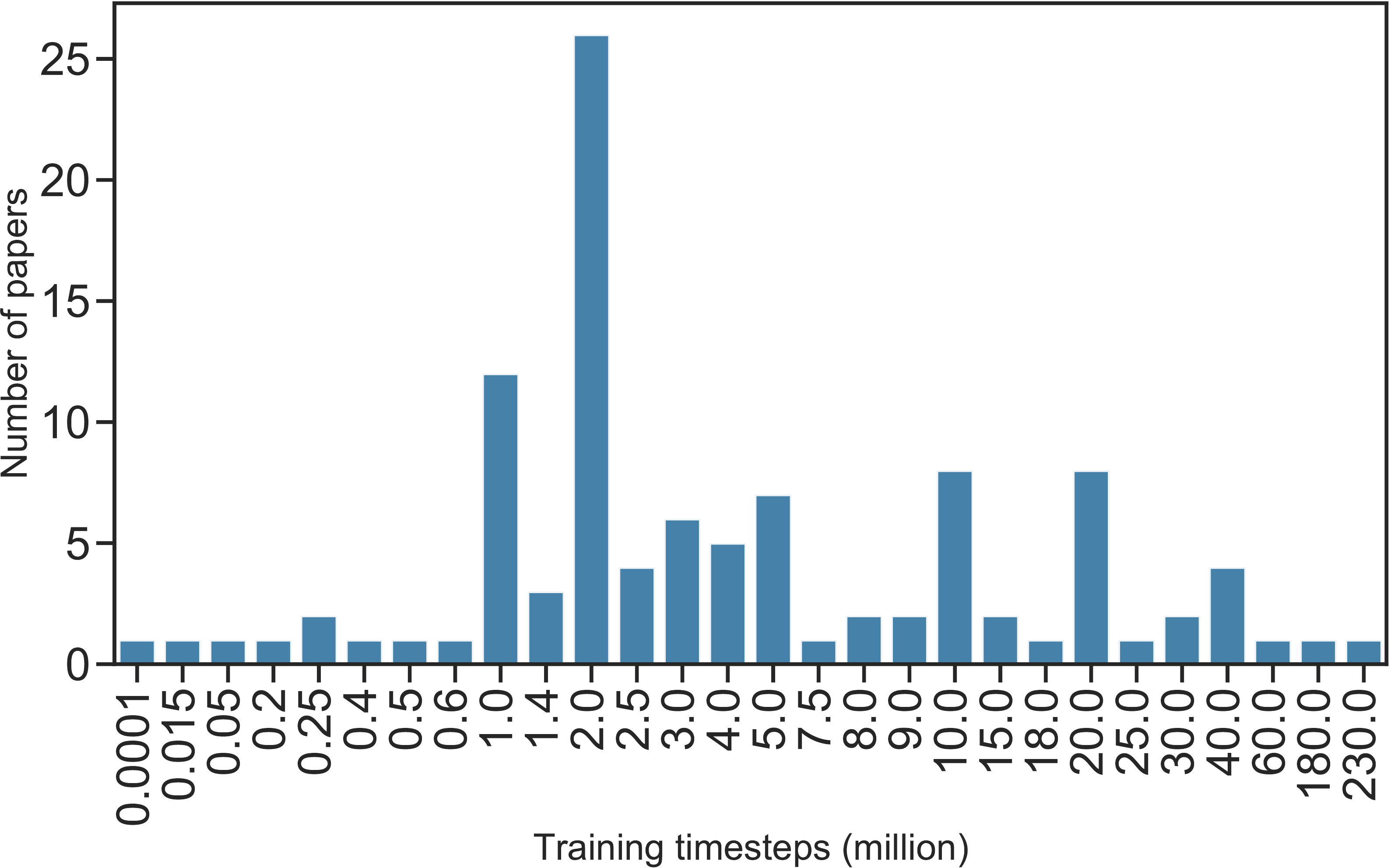}
    \includegraphics[width=0.24\textwidth]{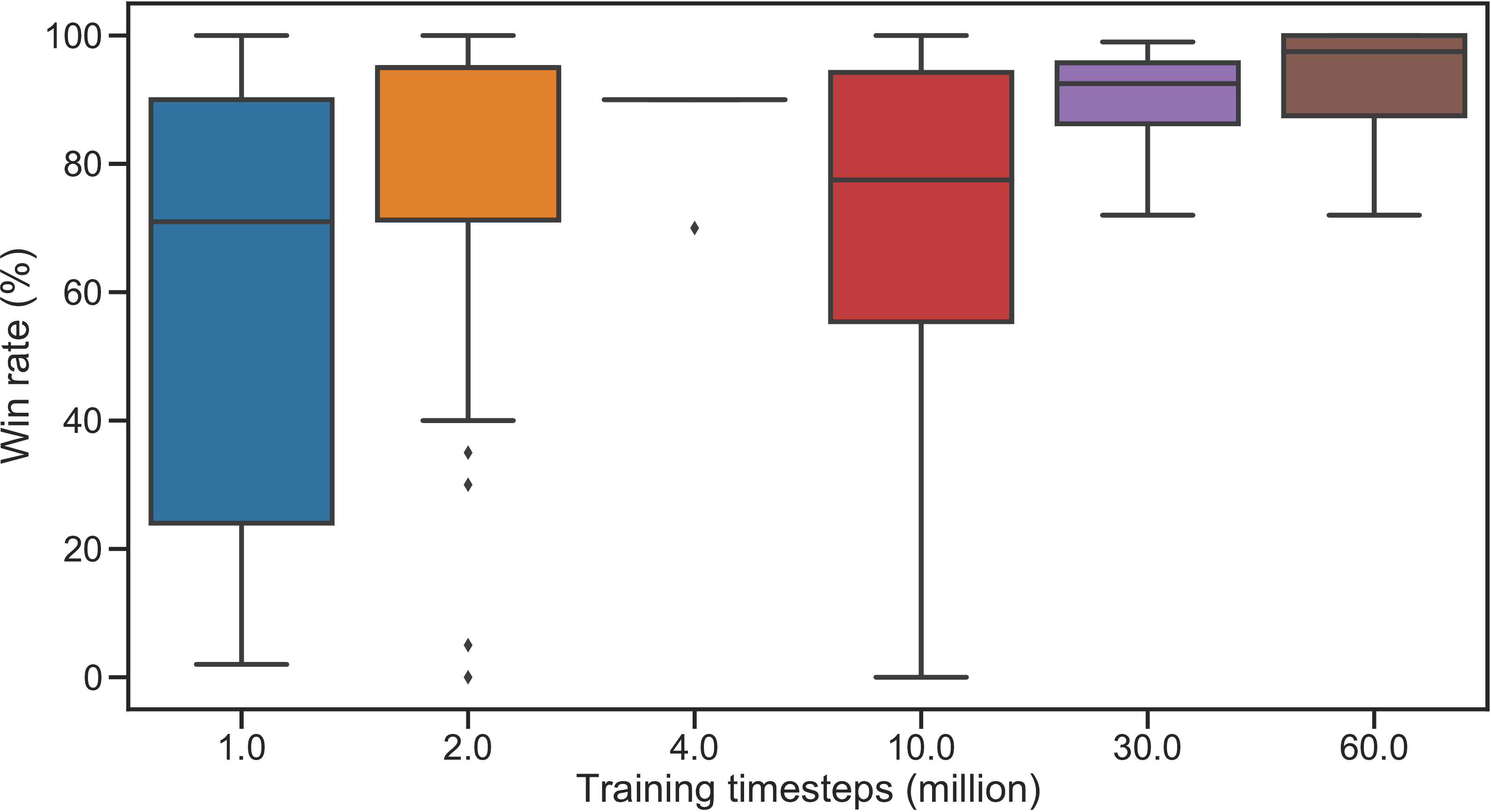}
    \caption{\textbf{Top:} Evaluation methodologies in papers: how frequent and for how many runs to do evaluation. ``E'' refers to episode and ``T'' to timestep.  \textbf{Bottom:} Number of total training timesteps used in papers (left) and win rates per training timesteps on 2s3z (right).}
    \label{fig: exp parameters}
\end{wrapfigure}

\textbf{Evaluation parameters. }  In Figure \ref{fig: exp parameters}, we show the evaluation parameters for the number of evaluation runs, the evaluation interval (top) and the training time (bottom left) used in papers. We find it is most common to use 32 evaluation episodes at every 10000 timesteps (defined as the steps taken by agents when acting in the environment) and to train for 2 million timesteps in total. We note that these numbers are skewed towards earlier years of SMAC evaluation and that recent works have since explored far longer training times. However, we argue that these longer training times are not always justified (e.g. see the bottom right of Figure \ref{fig: exp parameters}). Furthermore, SMAC is one of the most expensive MARL environments (Appendix A.7 in \citet{papoudakis2021benchmarking}) and for the future accessibility of research in terms of scale and for fair comparisons across different works, we recommend the above commonly used values as reasonable starting defaults and support the view put forward by \cite{dodge2019show} that results should be interpreted as a function of the compute budget used. We of course recognise that these evaluation parameters can be very specific to the environment, or task, and again we urge environment designers to play a role in helping the community develop and adopt sensible standards. Finally, as done by \cite{papoudakis2021benchmarking}, we recommend treating off-policy and on-policy algorithms differently and train on-policy algorithms for a factor of 10 more timesteps than off-policy algorithms due to differences in sample efficiency. As argued by \cite{papoudakis2021benchmarking}, with modern simulators, the wall-clock time of on-policy training done in this way is typically not much slower than off-policy training.

\textbf{Performance and uncertainty quantification. } To aggregate performance across evaluation intervals, we recommend using the \textit{absolute} performance metric proposed by \cite{colas2018gep}, computed using the best average (over training runs) joint policy found during training. Typically, practitioners checkpoint the best performing policy parameters to use as the final model, therefore it makes sense to do evaluation in a similar way. However, to account for not averaging across different evaluation intervals, \cite{colas2018gep} recommend increasing the number of independent evaluation runs using the best policy by a factor of 10 compared to what was used at other intervals. To quantify uncertainty, we recommend using the mean with 95\% confidence intervals (CIs) at each evaluation interval (computed over independent evaluation runs), and when aggregating across tasks within an environment, we recommend using the tools proposed by \cite{agarwal2022deep}, in particular, the inter-quartile mean (IQM) with 95\% stratified Bootstrap CIs.

\textbf{Reporting. } We strongly recommend reporting \textit{all} relevant experimental details including: hyperparameters, code-level optimisations, computational requirements and frameworks used. Taking inspiration from \textit{model cards} \citep{mitchell2019model}, we provide templates for reporting in the Appendix. Furthermore, we recommend providing experimental results in multiple formats, including plots \textit{and} tables per task and environment as well as making all raw experimental data and code publicly available for future analysis and easy comparison. Finally, we encourage authors to include detailed ablation studies in their work, so as to be able to accurately attribute sources of improvement.

\begin{mybox}{A Standardised Performance Evaluation Protocol for Cooperative MARL}
    \footnotesize
    \textbf{Input: } Environments with tasks $t$ from a set $\mathcal{T}$. Algorithms $a \in \mathcal{A}$, including baselines and novel work.\\

    \textbf{1. Evaluation parameters -- defaults}
    \begin{itemize}
        \item Number of training \textit{timesteps}, $T = 2$ million (off-policy), $T= 20$ million (on-policy).
        \item Number of independent training \textit{runs}, $R = 10$ (from \cite{agarwal2022deep})
        \item Number of independent evaluation \textit{episodes} per interval, $E = 32$.
        \item Evaluation \textit{intervals}, $i \in \mathcal{I}$, at every 10000 timesteps.
    \end{itemize}
    \textbf{2. Performance and uncertainty quantification}
    \begin{enumerate}
        \item Performance metric: Always use returns $G$ (applicable to all environments), \emph{and} the environment specific metric (e.g. Win rate).
        \item Per task evaluation: Compute the mean $G^a_t$ over $E$ episodes at each evaluation interval $i$, where $G^a_t$ is the return of algorithm $a$ on task $t$, with 95\% CI, for all $a$.
        \item Per environment evaluation: 
        \begin{itemize}
            \item Compute the normalised \emph{absolute} return \citep{colas2018gep} as the mean return of $10\times E = 320$ evaluation episodes using the best joint policy found during training and normalising the return to be in the range $[0,1]$ using $(G^a_t - \min(G_t))/(\max(G_t)-\min(G_t))$, where $G_t$ is the return for all algorithms on task $t$. 
            \item For each algorithm $a$, form an evaluation matrix with shape $(R, |\mathcal{T}|)$ where each entry is the normalised absolute return for a specific training run on a specific task. 
            \item Compute the IQM and optimality gap with 95\% stratified Bootstrap CIs, probability of improvement scores and performance profiles, to compare the algorithms, using the tools proposed by \cite{agarwal2022deep}.\footnote{\scriptsize These can be found in the \texttt{rliable} library: \href{https://github.com/google-research/rliable}{https://github.com/google-research/rliable}} Sample efficiency curves can be computed by using normalised returns at each evaluation interval.
        \end{itemize}
    \end{enumerate}
    \textbf{3. Reporting}
    \begin{itemize}
        \item Experiments: All hyperparameters, code-level optimisations, computational requirements and framework details.
        \item Plots: All task and environment evaluations as well as ablation study results.
        \item Tables: Normalised absolute performance per task with 95\% CI for all tasks, IQM with 95\% stratified Bootstrap CIs per environment for all environments.
        \item Public repository: Raw evaluation data and code implementations.
    \end{itemize}
\end{mybox}

\section{Conclusion and future work}

In this work, we argue for the power of standardisation. In a fast-growing field such as MARL, it becomes ever more important to be able to dispel illusions of rapid progress, potentially misleading the field and resulting in wasted time and effort. We hope to break the spell by proposing a sensible standardised performance evaluation protocol, motivated in part by the literature on evaluation in RL, as well as by a meta-analysis of prior work in MARL. If widely adopted, such a protocol could make comparisons across different works much faster, easier and more accurate. However, certain aspects of evaluation are better left standardised outside of the control or influence of authors, such as protocols pertaining to the use of benchmark environments. We believe this is an overlooked issue and an important area for future work by the community, and specifically environment designers, to jointly establish better standards and protocols for evaluation and environment use. Finally, we encourage the community to move beyond the use of only one or two environments with static task sets (e.g. MPE and SMAC) and focus more on building algorithms, environments and tools for improving generalisation in MARL.

A clear limitation of our work is our focus on the cooperative setting. Interesting works have developed protocols and environments for evaluation in both the competitive and mixed settings \citep{omidshafiei2019alpha,rowland2019,leibo2021scalable}. We find this encouraging and argue for similar efforts in the adoption of proposed standards for evaluation.

\newpage

\begin{ack}
The authors would like to kindly thank the following people for useful discussions and feedback on this work: Jonathan Shock, Matthew Morris, Claude Formanek, Asad Jeewa, Kale-ab Tessera, Sinda Ben Salem, Khalil Gorsan Mestiri, Chaima Wichka and Sasha Abramowitz.
\end{ack}

\bibliographystyle{IEEEtranN}
\bibliography{bib_file}
\newpage
\appendix
\section{Data collection and annotation methodology}
This section outlines the search methodology and data recording practices used to collect the dataset of algorithm performance and evaluation methodologies for the field of cooperative MARL.\footnote{Meta-analysis dataset on MARL evaluation \url{https://bit.ly/3LpxAMb}} The dataset used in the main body of this paper reflects the algorithm evaluation practices of published cooperative MARL papers only. We note that the original data collection was not restricted to accepted publications and cooperative MARL, as it instead attempts to incorporate all prominent and contemporary deep MARL algorithms and approaches from all available studies. This is reflected in this appendix, where we refer to data collected from all recorded papers (published, rejected, unknown, and non-cooperative) as \textit{all papers}. Similarly, we refer to the data collected from cooperative published papers (which were used in the main body of this work) as \textit{the main papers}. The non-published papers and non-cooperative published papers are referred to as \textit{the other papers}.

\subsection{Paper search strategy}
\begin{figure}[hbt!]
\centering
\includegraphics[width=0.8\textwidth]{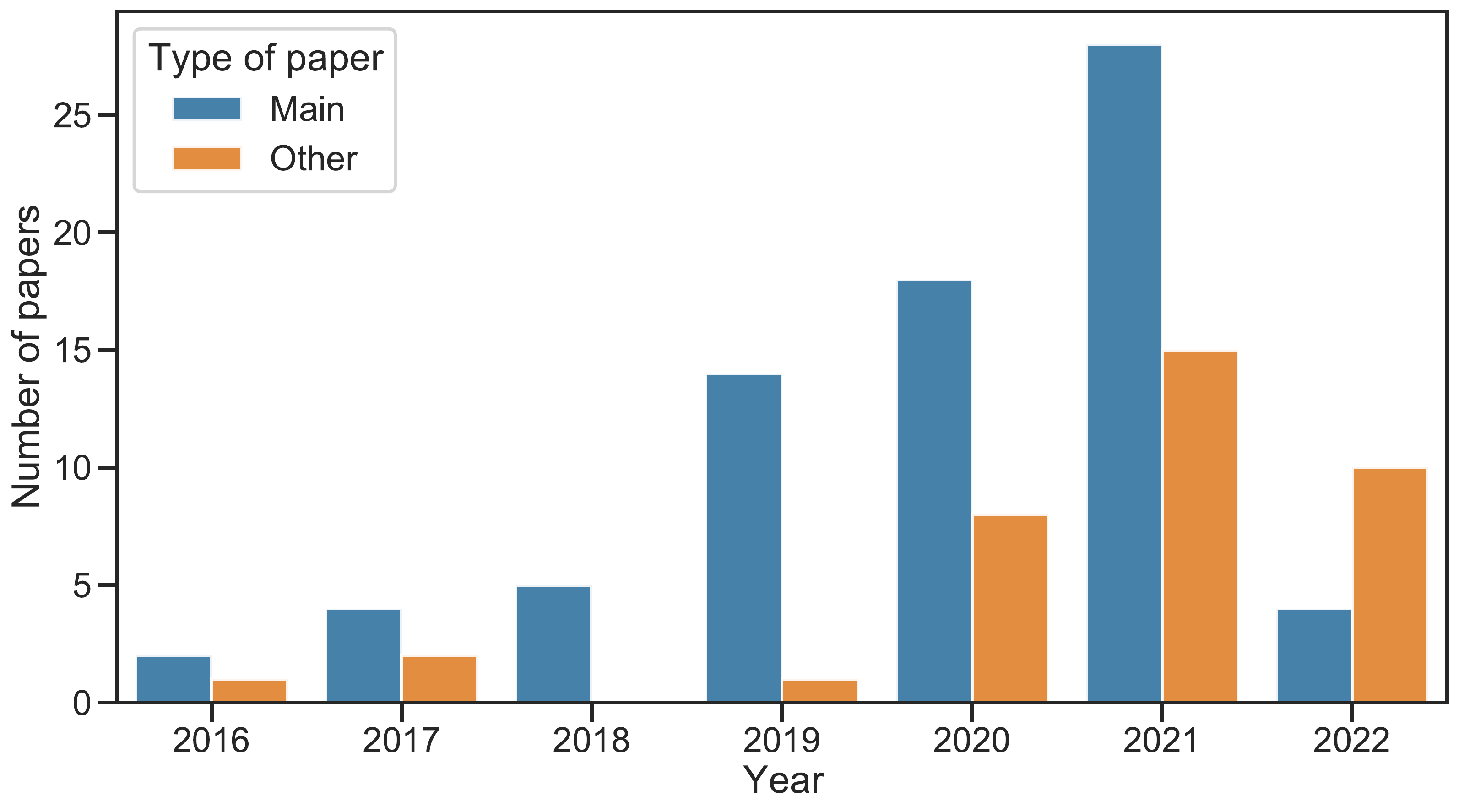}
\caption{Recorded papers by year}

\end{figure}
In order to gather data on MARL algorithm performance evaluation, we gathered relevant MARL research papers which were published between the years 2016 and 2022. To identify relevant studies, we searched for relevant research key terms, such as “Multi-agent RL”, “MARL evaluation” and “Benchmarking MARL”. We searched the arXiv website for these terms in different combinations of the title, abstract, and keywords. Additionally, several papers were included from the reference list of other papers. Although we do not claim to have a dataset comprised of all modern deep MARL algorithms, we strive to collect data on at least all of the most widely used deep MARL algorithms. To our knowledge, all major deep MARL algorithms are represented in our dataset and this dataset is the first of its kind. The search queries were finalized on the 8th of April 2022. The published research papers that we recorded can be found in Table \ref{tab:papers}, where these were published at various conferences including ICML, NeurIPS, ICLR, and others. 

\subsection{Filtering data to find relevant studies}

Following the initial data collection, the dataset was refined to ensure relevance using the following criteria:
\begin{itemize}
    \item The papers must be either peer reviewed conference or journal papers, and published in the English language.
    \item Papers were restricted to only those which focus exclusively on the cooperative MARL case.
\end{itemize}

\begin{table}[H]
\caption{Published cooperative MARL research papers collected and manually annotated for data analysis of algorithm performance evaluation methods.}
\label{tab:papers}
\resizebox{0.7\textwidth}{!}{\begin{minipage}{\textwidth}
\begin{tabular}{@{}llll@{}}
\toprule
Title  & Authors  & Conference \\ 
\midrule
Learning Multiagent Communication with Backpropagation     & \cite{Sukhbaatar-2016-LearningCommunicationBackprop}  &  NeurIPS \\
Learning to Communication in Deep Multi-Agent Reinforcement Learning     & \cite{Foerster-2016-LearningCommunication}  &  NeurIPS \\
Deep Decentralized Multi-task Multi-Agent Reinforcement Learning under Partial Observability  & \cite{Shayegan-2017-decentralizedMARLPartialObs}  &  ICML \\
Multi-Agent Actor-Critic for Mixed Cooperative-Competitive Environments   & \cite{MAACMixed-2017-Lowe}  &  NeurIPS \\
Stabilising Experience Replay for Deep Multi-Agent Reinforcement Learning   & \cite{stabilising-2017-Foerster}  &  ICML \\
MultiAgent Soft-Q Learning   & \cite{MASoftQ-2018-Wei}  &  AAAI \\
Counterfactual Multi-Agent Policy Gradients  & \cite{foerster-2018-counterfactual} & AAAI \\
Value-Decomposition Networks For Cooperative Multi-Agent Learning Based On Team Reward & \cite{sunehag-2018-valueDecomp} & AAMAS\\ 
QMIX: Monotonic Value Function Factorisation for Deep Multi-Agent Reinforcement Learning  & \cite{qmix-2018-Rashid}  &  ICML \\
Learning when to Communicate at Scale in Multiagent Cooperative and Competitive Tasks  & \cite{singh2018LearningCommsScale}  &  ICLR \\
Actor-Attention-Critic for Multi-Agent Reinforcement Learning  & \cite{actorAttention-2019-Chaudhuri}  &  ICML \\
Efficient Communication in Multi-Agent Reinforcement Learning via Variance Based Control & \cite{Zhang-2019-efficientComms}  &  NeurIPS \\
MAGNet: Multi-agent Graph Network for Deep Multi-agent Reinforcement Learning& \cite{Malysheva-2019-Magnet}  & IEEE \\
Modelling the Dynamic Joint Policy of Teammates with Attention Multi-agent DDPG  & \cite{mao-2019-modelling} & AAMAS \\
The StarCraft Multi-Agent Challenge & \cite{SMAC-2019-Samvelyan}  & AAMAS \\
Social Influence as Intrinsic Motivation for Multi-Agent Deep Reinforcement Learning & \cite{Jaques-2019-SocialInfluence}  & ICML \\
LIIR: Learning Individual Intrinsic Reward in Multi-Agent Reinforcement Learning & \cite{LIIR-2019-Du}  & NeurIPS \\
MAVEN: Multi-Agent Variational Exploration & \cite{MAVEN-2019-Mahajan}  & NeurIPS \\
Multi-Agent Common Knowledge Reinforcement Learning & \cite{Schroeder-2019-MAcommonKnowledge}  & NeurIPS \\
A Structured Prediction Approach for Generalization in Cooperative Multi-Agent Reinforcement Learning & \cite{Carion-2019-Structured}  & NeurIPS \\
TarMAC: Targeted Multi-Agent Communication & \cite{TarMAc-2019-Das}  & ICML \\
QTRAN: Learning to Factorize with Transformation for Cooperative Multi-Agent Reinforcement Learning & \cite{qtran-2019-Son}  & ICML \\
Influence-Based Multi-Agent Exploration & \cite{Wang-2020Influence-Based}  &  ICLR \\
Multi-Agent Game Abstraction via Graph Attention Neural Network & \cite{Liu_Wang_Hu_Hao_Chen_Gao_2020}  &  AAAI \\
Feudal Multi-Agent Hierarchies for Cooperative Reinforcement Learning & \cite{Ahilan-2019-Feudal}  &  AAMAS \\
PIC: Permutation Invariant Critic for Multi-Agent Deep Reinforcement Learning & \cite{pic-2020-Liu}  & CoRL \\
Action Semantics Network: Considering the Effects of Actions in Multiagent Systems & \cite{Wang-2020-ActionSemantics}  & ICLR \\
Succinct and Robust Multi-Agent Communication With Temporal Message Control & \cite{zhang-2020-succinctMAComms}  & NeurIPS \\
Learning Multi-Agent Coordination for Enhancing Target Coverage in Directional Sensor Networks & \cite{xu-2020-coordinationTargetCoverage}  & NeurIPS \\
Learning Nearly Decomposable Value Functions Via Communication Minimization & \cite{wang2020learning}  & ICLR \\
Promoting Coordination through Policy Regularization in Multi-Agent Deep Reinforcement Learning & \cite{Promoting-2020-Roy}  & NeurIPS \\
Shapley Q-value: A Local Reward Approach to Solve Global Reward Games& \cite{wang-2020-shapley}  & AAAI \\
Deep Coordination Graphs & \cite{Boehmer-2020-deepCoordinationGraph}  & ICML \\
Evolutionary Population Curriculum for Scaling Multi-Agent Reinforcement Learning  & \cite{Long-2020-evolutionary}  & ICLR \\
Shared Experience Actor-Critic for Multi-Agent Reinforcement Learning & \cite{christianos-2020-shared}  & NeurIPS \\
SMIX($\lambda$): Enhancing Centralized Value Functions for Cooperative Multi-Agent Reinforcement Learning  & \cite{smix-2020-wen}  & AAAI \\
Learning Transferable Cooperative Behavior in Multi-Agent Teams  & \cite{agarwal-2020-learningTransferable} & AAMAS \\
Comparative Evaluation of Cooperative Multi-Agent Deep Reinforcement Learning Algorithms & \cite{papoudakis2020comparative}  &  AAMAS\\
Learning Individually Inferred Communication for Multi-Agent Cooperation  & \cite{Ding-2020-Learning}  & NeurIPS \\
Simplified Action Decoder for Deep Multi-Agent Reinforcement Learning & \cite{Hu2020Simplified}  & ICLR \\
Learning Implicit Credit Assignment for Cooperative Multi-Agent Reinforcement Learning & \cite{zhou-2020-learning}  & NeurIPS \\
Variational Automatic Curriculum Learning for Sparse-Reward Cooperative Multi-Agent Problems & \cite{chen-2021-variational}  & NeurIPS \\
Pessimism Meets Invariance: Provably Efficient Offline Mean-Field Multi-Agent RL & \cite{chen-2021-pessimism}  & NeurIPS \\
Deep Implicit Coordination Graphs for Multi-agent Reinforcement Learning & \cite{Li-2021-DeepImplicitCoord}  & AAMAS \\
DFAC Framework: Factorizing the Value Function via Quantile Mixture for Multi-Agent Distributional Q-\\Learning & \cite{dfac-2021-sun}  & ICML \\
Scaling Multi-Agent Reinforcement Learning with Selective Parameter Sharing & \cite{scaling-2021-christianos}  & ICML \\
Towards Understanding Cooperative Multi-Agent Q-Learning with Value Factorization & \cite{wang-2021-towards}  & NeurIPS \\
Investigation of Independent Reinforcement Learning Algorithms in Multi-Agent Environments & \cite{lee2021investigation}  & NeurIPS \\
Celebrating Diversity in Shared Multi-Agent Reinforcement Learning & \cite{celebrating-2021-chenghao}  & NeurIPS \\
RODE: Learning Roles to Decompose Multi-Agent Tasks & \cite{RODE-2021-Wang}  & ICLR \\
Local Advantage Actor-Critic for Robust Multi-Agent Deep Reinforcement Learning & \cite{xiao-2021-local} & IEEE MRS \\
MMD-MIX: Value Function Factorisation with Maximum Mean Discrepancy for Cooperative Multi-Agent\\ Reinforcement Learning & \cite{zniwei-2021-mmdmix}  &  IJCNN\\
The Emergence of Individuality & \cite{emergence-2021-jiang}  &  ICML\\
QVMix and QVMix-Max: Extending the Deep Quality-Value Family of Algorithms to Cooperative Multi-\\Agent Reinforcement Learning & \cite{qvmixqvmax-2021-Leroy}  & AAAI \\
Weighted QMIX: Expanding Monotonic Value Function Factorisation for Deep Multi-Agent Reinforcement\\Learning & \cite{rashid-2021-weightedQMix}  & NeurIPS \\
Value-Decomposition Multi-Agent Actor-Critics & \cite{adams-2021-valueDecomp}  & AAAI \\
Regularized Softmax Deep Multi-Agent Q-Learning & \cite{pan-2021-regularized}  &  NeurIPS\\
Cooperative Exploration for Multi-Agent Deep Reinforcement Learning & \cite{coop-2021-meila}  &  ICML\\
Domain-Aware Multiagent Reinforcement Learning in Navigation & \cite{domain-2021-saeed}  &  IJCNN\\
Evaluating Generalization and Transfer Capacity of Multi-Agent Reinforcement Learning Across Variable\\Number of Agents & \cite{guresti-2021-evaluating}  &  AAAI\\
Episodic Multi-agent Reinforcement Learning with Curiosity-driven Exploration & \cite{Zheng-2021-episodic}  &  NeurIPS\\
Benchmarking Multi-Agent Deep Reinforcement Learning Algorithms in Cooperative Tasks & \cite{papoudakis2021benchmarking}  &  NeurIPS\\
Centralizing State-Values in Dueling Networks for Multi-Robot Reinforcement Learning Mapless Navigation & \cite{centralizing-2021-marchesini}  &  IROS\\
QPLEX: Duplex Dueling Multi-Agent Q-Learning  & \cite{wang2021qplex}  &  ICLR\\
Settling the Variance of Multi-Agent Policy Gradients  & \cite{kuba-2021-settling}  &  NeurIPS\\
FACMAC: Factored Multi-Agent Centralised Policy Gradients & \cite{peng-2021-facmac}  &  NeurIPS\\
Multi-Agent Incentive Communication via Decentralized Teammate Modeling & \cite{LeiYuan2022MultiAgentIC}  &  AAAI\\
LIGS: Learnable Intrinsic-Reward Generation Selection for Multi-Agent Learning  & \cite{mguni2022ligs}  &  ICLR\\
ToM2C: Target-oriented Multi-agent Communication and Cooperation with Theory of Mind & \cite{wang2022tomc}  & ICLR \\
Trust Region Policy Optimisation in Multi-Agent Reinforcement Learning & \cite{kuba2022trust}  &  ICLR\\
Reinforcement Learning for Location-Aware Warehouse Scheduling  & \cite{stavroulakis2022reinforcement} & ICLR\\ 
Multi-agent Transfer Learning in Reinforcement Learning-based Ride-sharing Systems & \cite{Castagna-2022-MAtransfer} & ICAART\\ 
Off-Policy Correction For Multi-Agent Reinforcement Learning  & \cite{zawalski-2022-offpolicy} & AAMAS \\
Local Advantage Networks for Cooperative Multi-Agent Reinforcement Learning  & \cite{avalos-2022-local} & AAMAS \\
A Deeper Understanding of State-Based Critics in Multi-Agent Reinforcement Learning & \cite{Lyu-2022-deeper} & AAAI \\
\bottomrule
\end{tabular}
\end{minipage}}
\end{table}

\subsection{Annotations}

The collected dataset was manually annotated to record methods of algorithm performance evaluation. The dataset records the algorithms, environments, and tasks used as well as all aspects relating to the algorithm performance evaluation procedure that was available from the papers. The following provides further details on the data annotation procedure:

\begin{enumerate}[label=A{{\arabic*}}.]
\item The names and dates of papers are recorded along with the conferences they are published into and research labs associated with the authors.

\item The algorithms being evaluated are recorded. In some cases the paper-specific names of algorithms have been appropriately adapted. This is in cases where uniquely named algorithms have only minor differences from their baselines. Further details of this standardisation appear in subsection \ref{sec:algAnotation}. The algorithm libraries used are recorded when applicable (e.g. EPyMARL \cite{papoudakis2021benchmarking}).

\item We recorded the MARL environments, their sub-tasks/maps/scenarios and the choice of version used for evaluation. Environment sub-tasks with different names, but which refer to the identical sub-tasks were given standardised names (e.g. cooperative communication is the second name for Speaker-Listener task in MPE).

\item With regard to performance measurement, we recorded the aggregation functions across runs or episodes (e.g. means) and, the metrics used (e.g. SMAC win rates or max rewards) along with their measure of spread such as reported confidence interval values or standard deviations. Additionally, to compare between cases of when win rates or rewards are recorded, we report the \textit{general metric} used.

\item On occasion, data is only provided from performance plots and not from tables. Hence our dataset records whether data is presented using plots or in tabular form. When data is only provided by plots, we record the final value for a given metric as shown on a plot. For the purpose of our records being as accurate as possible, we ensure these values are within 5\% of their true plotted value. Since we cannot exactly determine the confidence bounds from plots alone we omit recording such values in these cases. However, we do still record the type of uncertainty measure used, as presented by the author (and where available elsewhere, the uncertainty values).

\item The evaluation intervals (evaluation frequency) and independent evaluations per interval (evaluation duration) were also recorded along with their units (e.g. episodes or timesteps). This includes the number of training runs and number of random seeds used. Here, evaluation intervals that refer to the same measurement across papers were standardized (e.g. rounds are changed to episodes).

\item We record whether reported results are from previous works, i.e. when reported results are from other cited papers and are not reproduced in the particular paper being recorded.
\end{enumerate}

\subsubsection{Environment annotations}
\begin{itemize}
    \item All SMAC win rates are reported as percentages (out of 100) and not probabilities (out of 1).
    \item We record an environment as paper-specific if it is created by the authors of a particular paper and is not utilized in any other article.
\end{itemize}

\subsubsection{Algorithm annotations}
\label{sec:algAnotation}

In the process of collecting the data for this paper it came to our attention that several algorithms go by slightly different names across multiple papers. For the purposes of our analysis we have standardised these naming choices, based on algorithm descriptions made by authors in their respective papers, to more standardised naming conventions. \textit{IAC-V} is first mentioned in the paper that presents COMA \cite{foerster-2018-counterfactual}. Due to the paper emerging very early into the growth of cooperative MARL naming had not yet been normalised however, \textit{IAC-V} is described as a standard advantage actor-critic (AAC) algorithm using parameter sharing and can instead just be referred to as IAC. \textit{PSMADDPG} \cite{mao-2019-modelling} is a variant of MADDPG that makes use of parameter sharing which is the norm in many other publications. Interestingly the original MADDPG paper \cite{MAACMixed-2017-Lowe} does not make use of this. \textit{PSMADDPG} can be considered to be MADDPG with a different implementation choice and is grouped with MADDPG as the underlying algorithm is not altered. Both \textit{A3C} and \textit{A2C} are named in the publications used in this analysis \cite{Wang-2020-ActionSemantics, Jaques-2019-SocialInfluence}. A2C and A3C refer to the method by which the AAC algorithm is implemented to run using multiple parallel workers with A2C being the synchronous and A3C being the asynchronous variant \cite{mnih2016}. Very early MARL papers referred to independent Q learning simply as Deep Q Network \cite{Tampuu2015}. As MARL developed further it became more important to distinguish between independent and centralised learners and DQN is commonly called IQL. Similarly DDPG can be renamed to IDDPG to distinguish it as an independent learning algorithm. The centralised AAC algorithm is also sometimes called a naive critic. Instead we refer to this method as central-V as this is the first formalised name for this algorithm that we could find \cite{Foerster-2016-LearningCommunication}. Finally MAPPO \cite{yu2021} is referred to as MAPPO-shared for MAPPO with parameter sharing. However, parameter sharing if the norm amongst most cooperative MARL publications therefore, MAPPO-shared is simply renamed to MAPPO.

\begin{table}[hbt!]
  \caption{Algorithm annotations}
  \label{sample-table}
  \centering
  \makebox[1 \textwidth][c]{ 
  \begin{tabular}{lll}
    \toprule
    Name from paper & Standardised naming &  Our interpretation  \\
    \midrule
    IAC-V \citep{foerster-2018-counterfactual} & IAC  & IAC-V is the same as IAC. \\
    PSMADDPG (\cite{mao-2019-modelling}) & MADDPG  & The PS denotes parameter sharing.\\
    A3C \citep{Wang-2020-ActionSemantics} & IAC  & Asynchronous parallelization method for IAC.\\
    A2C \citep{Jaques-2019-SocialInfluence} & IAC  & Synchronous parallelization method for IAC.\\
    MADQN \citep{Tampuu2015} & IQL  & Old naming conventions.\\
    Naïve critic \citep{adams-2021-valueDecomp} & Central-V  & Naïve critic is the same as central-v.\\
    MAPPO-shared \citep{lee2021investigation} & MAPPO  &  Parameter sharing is the norm.\\
    MADR \citep{Park2020} & MADDPG  & MADDPG with recurrency. \\
    DDPG \citep{MAACMixed-2017-Lowe} & IDDPG  & Denote as independent learner. \\
    DQN \citep{Tampuu2015} & IQL  & Denote as independent learner. \\
    \bottomrule
  \end{tabular}}
\end{table}

\clearpage

\section{Additional Analysis}

This section provides additional insights from further analysis on our dataset of performance evaluation for cooperative MARL algorithms.

\subsection{Environment}
\subsubsection{Most used settings}

In this section, we are primarily interested in highlighting some of our further findings from the main papers. We first illustrate some of the most widely used settings for the most popular environments as illustrated in Table \ref{env:tab}. It should be noted that this analysis was conducted over 29 unique environments with 164 unique scenarios.  

\begin{table}[h]
  \caption{Most applicable parameters in each environment for the main papers}
  \label{env:tab}
  \centering
  \makebox[1 \textwidth][c]{ 
  \begin{tabular}{lclclclclclcl}
    \toprule
    Environment & Metric & R. Seed & Aggregate Function & Independent variable & Maps/Tasks & Mentions \\
    \midrule
    SMAC\cite{} & Win Rate (83.3\%) & 5 (41.7\%) & Median (48.4\%) & Timestep (97.3\%) & 39 & 37 \\
    MPE\cite{} & Reward (40\%) & 5 (34.8\%) & Mean (85\%) & Episode (48\%) & 25 & 33\\
    Matrix Games\cite{} & Return (100\%) & 5-10-100 & Mean (100\%) & Timestep (98.7\%) & - & 9\\
    MazeBase\cite{} & Win Rate (87.5\%) & 5 (80\%) & Mean (100\%) & Episode (44.1\%) & 2 & 7 \\
    \bottomrule
  \end{tabular}}
\end{table}

\textbf{StarCraft Multi-Agent Challenge (SMAC):} is a partially observable environment, with a diverse set of sophisticated micro-actions that enable the learning of complex interactions amongst collaborating agents, the fundamental concept of SMAC is a team of agents battling against another group of units. SMAC is the most widely used environment in our analysis, since it is employed as the experimental environment in 37 of the main papers presenting 46.9\% of the collected evaluation data. This finding is not surprising as we have recorded 39 unique SMAC scenarios with varying scales of difficulty. Moreover, many authors agree that SMAC offers a fair comparison of different algorithms since it provides an open-source Python-based implementation of numerous fundamental MARL algorithms.

\textbf{Multi-Agent Particle Environment (MPE):} is an environment that can be fully or partially observable, cooperative or competitive, and allow communication within some of its tasks. In this environment the agents primarily interact with the landmarks and other entities to achieve various goals. We discover that 33 of the 75 papers employ MPE for algorithm testing, accounting for 20.3\% of the collected evaluation data. MPE, like SMAC, is a diversified environment with 25 tasks; nevertheless, we observe a disparity in their utilization, with 27.3\% of the main papers utilizing Predator and Prey, followed by Spread which is used in 22.7\% of the collected main papers which use the MPE environment.

\subsubsection{Evolution of environment usage in MARL}
\begin{figure}[h]
\centering
\includegraphics[width=1\textwidth]{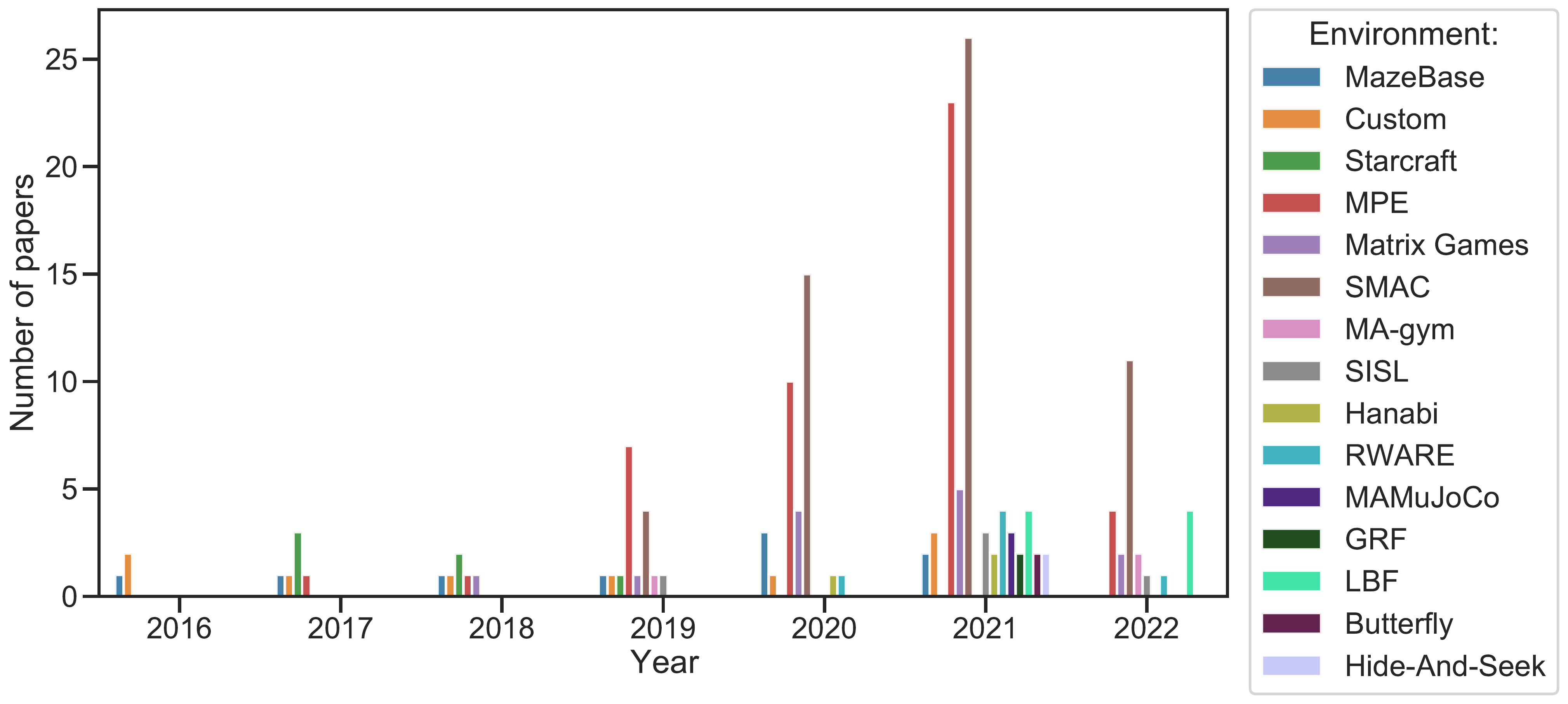}
\caption{Evolution of environment usage in all the papers}
\label{env}
\end{figure}

In the early years of MARL research there was a shortage of established multi-agent environments, as shown in figure \ref{env} \footnote{The plotted environments occur in at least two papers.}. Hence most publications tested their algorithms on environments created by the authors (paper-specific environments) as well as MazeBase. Although MazeBase was developed for single-agent environments, it is easily adaptable to the multi-agent case and was used to create the traffic junction combat tasks. This adaptability drove its early adoption. The Figure depicts that, since 2017, we can observe an increase in the use of MPE tasks like Predator-Prey and Spread, as well as StarCraft unit micromanagement. MPE was the most used environment in 2019 and, since 2020, we see SMAC dominating the others.

\subsection{Algorithms}

\subsubsection{Training schemes analysis}
\textbf{Independent Learning (IL or DTDE): } is a method that extends single-agent RL algorithms to the multi-agent space. Agents learn an independent policy based on their own local observations and, in the cooperative case, learn a policy based on a shared global reward. This type of learning has low convergence guarantees because the learning of other agents causes the environment to appear non-stationary to each individual agent since the agents' behavior changes the dynamics of the environment.

\begin{wrapfigure}{r}{0.5\textwidth} 
    \centering
    \includegraphics[width=0.5\textwidth]{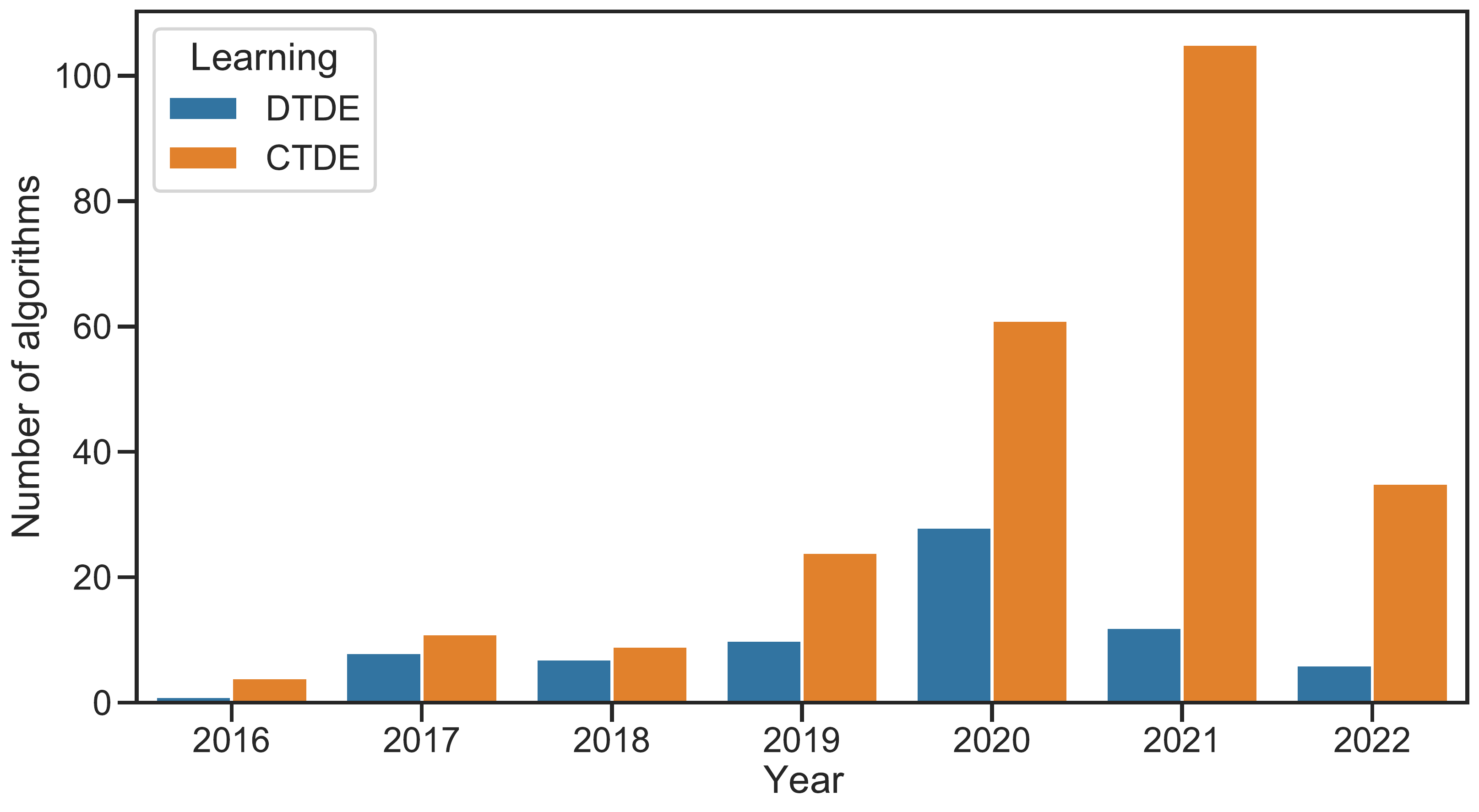}
    \caption{Number of algorithms based on learning schemes by years for all the papers}
\end{wrapfigure}

\textbf{Centralised Training Decentralised Execution (CTDE):}
much like IL, CTDE learns decentralised agent policies where agents act based on local observations. However, in the CTDE paradigm we can make use of additional information at training time that is normally not available to agents during execution. Typically this is done by using a \textit{centralised-critic} or some \textit{mixing network} which is allowed to condition on the global environment state information or, has access to open communication channels with all agents. The \textit{centralised-critic} or \textit{mixing network} is only used during training time which aids in finding better agent polices during training time without increasing computational overheads during execution time. 

\textbf{Is the decline in the use of IL over time a positive or negative sign?} CTDE has been demonstrated to be a powerful approach that outperforms decentralized training in many cases. Nevertheless, we cannot assume that it is the optimal solution in all cooperative MARL cases, since many studies, have shown that it is still hard for agents to act cooperatively during execution. This is because partial observability and stochasticity can easily break the learned cooperative strategy, resulting in miscoordination. Recently, we observe the increase of communication algorithms. These can make use of \textit{graph neural networks} as a communication channel to help agents obtain information during both training and execution.

\subsubsection{Benchmark algorithms}
In our analysis, we examine 150 algorithms where 73.3\% are used only once over the 75 main papers. In this section, we provide additional insights from the analysis of our data on the most relevant algorithms. We summarize the use of these algorithms in our dataset in Table \ref{Algo-table}.\\

\begin{table}[hbt!]
  \caption{Most used algorithms in the main papers}
  \label{Algo-table}
  \centering
  \begin{tabular}{llclclclc}
    \toprule
    Algorithms & Type of agent & CTDE & Policy  & Mentions  \\
    \midrule
    QMIX \citep{qmix-2018-Rashid} & Value-based  & Yes & Off & 35 \\
    MADDPG \citep{MAACMixed-2017-Lowe} & Actor-critic  & Yes & Off & 25\\
    VDN \citep{sunehag-2018-valueDecomp} & Value-based & Yes & Off & 23\\
    COMA \citep{foerster-2018-counterfactual} & Actor-critic  & Yes & On & 22\\
    IQL \citep{Tampuu2015} & Value-based  & No & Off & 20\\
    MAPPO \citep{yu2021} & Actor-critic  & Yes & On & 10\\
    QPLEX \citep{wang2021qplex} & Value-based  & Yes & Off & 10\\
    QTRAN \citep{qtran-2019-Son} & Value-based  & Yes & Off & 08\\
    IAC \citep{foerster-2018-counterfactual} & Actor-critic  & No & On & 08\\
    CommNet \citep{Sukhbaatar-2016-LearningCommunicationBackprop} & Policy optimization  & - & - & 06\\
    \bottomrule
  \end{tabular}
\end{table}

We note that one can select approximately five of these widely used algorithms, from Table \ref{Algo-table}, as baselines, against which one can evaluate the performance of a novel algorithm. As these algorithms are well-studied they may provide a meaningful current set for comparison. Athough we list these baselines, we do not consider this list to be exhaustive and researchers should strive to compare their algorithms to algorithms that are currently known to have state of the art (SOTA) performance. The five baselines we choose for discussion encompass both the CTDE and IL paradigm for cooperative MARL as well as policy gradient (PG) and Q-learning based methods. To meet these requirements we discuss QMIX \citep{qmix-2018-Rashid}, MADDPG \citep{MAACMixed-2017-Lowe}, COMA \citep{foerster-2018-counterfactual}, IQL \citep{Tampuu2015} and MAPPO \citep{yu2021}. QMIX is selected as it introduced the concept of monotonic value-decomposition which formed the basis for the development of many of the recent algorithmic developments. As shown by \citep{hu2021rethinking}, fine-tuned implementations of QMIX can still outperform newer methods that attempt to improve upon the original work. We discuss MADPPG since it was introduced in the most widely cited MARL algorithm paper with 2070 citation at the time of writing. We also note that MADDPG provides a baseline for algorithms that are used in mixed and competitive tasks. Although MADDPG was introduced as an algorithm to be used on environments with continuous action spaces, the algorithm may also be adapted to the discrete case. We discuss CommNet since it is a widely used algorithm, used in scenarios which require agent communication in order to find optimal solutions. Furthermore we discuss MAPPO due to recent work illustrating it's effectiveness in cooperative MARL tasks \citep{yu2021}. Lastly, we discuss COMA since it is a widely used actor-critic algorithm. Moreover, each of the algorithms mentioned have open-sourced code implementations available \citep{SMAC-2019-Samvelyan,papoudakis2021benchmarking, hu2021rethinking} which serve to decrease the amount of time researchers have to spend on implementing baselines to evaluate against.

\textbf{QMIX:} is a value-based algorithm introduced by \citep{qmix-2018-Rashid} following on from the success of VDN \citep{sunehag-2018-valueDecomp} in cooperative MARL tasks. Similarly to VDN, QMIX makes use of a factorized joint Q-value function to train all agents. What differentiates QMIX from VDN is that individual agents' utilities are joined using a mixing network instead of only summing them. Furthermore, the mixing network is constrained to having only positive weights, leading to a monotonic factorisation of individual agent utilities, and is allowed to condition on the global environment state during training time. QMIX follows the CTDE training paradigm and makes use of recurrent neural networks for individual agent policies. This enables agents to learn joint policies in partially observable settings. The initial performance of QMIX was illustrated by \citep{qmix-2018-Rashid} on the SMAC benchmark. 

In our analysis of QMIX, variants of QMIX and algorithms building on QMIX feature most prominently in the 2s3z (18), 3s vs 5z (14), 3s5z (14), MMM2 (13) and 6h vs 8z (11) SMAC scenarios. With numbers in parenthesis denoting the number of papers in which a QMIX variant is benchmarked on a particular scenario.

\textbf{CommNet:} \citep{Sukhbaatar-2016-LearningCommunicationBackprop} seeks to address the issue of effective agent communication in partially observable cooperative settings. What differentiated CommNet from previous communication works is that the communication protocol between agents is not fixed, but instead learnt as a neural model alongside agent training. This is possible due to agent communication being modeled using a continuous, differentiable vector which is output by each agent. We find that CommNET is used, most widely, on the TrafficJunction suite of environments which we find to be one of the most widely used communication benchmarks for MARL.

\textbf{Multi-Agent Deep Deterministic Policy Gradient (MADDPG): } introduced by \citep{MAACMixed-2017-Lowe}, is a multi-agent extension to the DDPG algorithm introduced by \citep{lillicrap2015continuous}. MADDPG is an off-policy actor-critic type of algorithm. By default, each agent has a unique policy network and Q-value critic network. Each agent's policy is only allowed to condition on an agent's partial observation of the full environment state while, during training time, each critic conditions on the actions selected by the policy networks of all other agents. MADDPG makes use of standard MLPs for both the agent policy and critic networks but variations of MADDPG exist which make use of recurrent neural networks (RNNs) for agent policies. Similarly, variations of MADDPG exist which make use of weight sharing across agent networks to aid in speedups of algorithm training. An advantage of MADDPG is that the algorithm is inherently applicable to both competitive, cooperative and mixed environments. This versatility is displayed in the seminal paper by \citep{MAACMixed-2017-Lowe}.\\ 
In our analysis, MADPPG is most widely used for benchmarking on the multi-agent particle environment suite (MPE) with the algorithm being most widely used on the Predator-Prey (12), Spread (10) and Speaker-Listener (5) scenarios. 

\textbf{Multi-Agent Proximal Policy Optimization (MAPPO):} is a multi-agent extension to the single-agent Proximal Policy Optimization (PPO) algorithm and mentioned explicitly by \citep{yu2021}. Similarly to PPO, MAPPO makes use of a value function, conditioned on the global environment state, to serve as a baseline leading to reduced variance in policy-gradient optimization. Furthermore, MAPPO may be implemented in the CTDE or IL paradigms depending on whether the value function is allowed to condition on some representation of the global environment state or only on an agent's local observation of the environment.\\
In our analysis, we find that MAPPO is used an equal amount of times (4) on the corridor, (3) MMM2, 5m vs 6m, 3s5z SMAC scenarios as well as on (2) the spread MPE scenario. 

\textbf{Counterfactual Multi-Agent Policy Gradients (COMA):} is an actor-critic algorithm the makes use of the CTDE paradigm by using a centralized critic, which is allowed to condition on the full environment state, with decentralized actors. This centralized critic is used during training time only and foregone at execution time. The core contribution of COMA is through addressing the agent credit assignment issue in MARL by utilizing a \textit{counterfactual} advantage function that is unique to each agent. 
In our analysis we find that COMA is used most frequently in the 2s3z (11), 3s5z (7), 1c3s5z (7) and the 3m (6) SMAC scenarios, as well (6) the Spread scenario from MPE. 

\subsection{Evaluation Settings}
\subsubsection{Metric}
\begin{wrapfigure}{r}{0.5\textwidth} 
    \centering
    \includegraphics[width=0.4\textwidth]{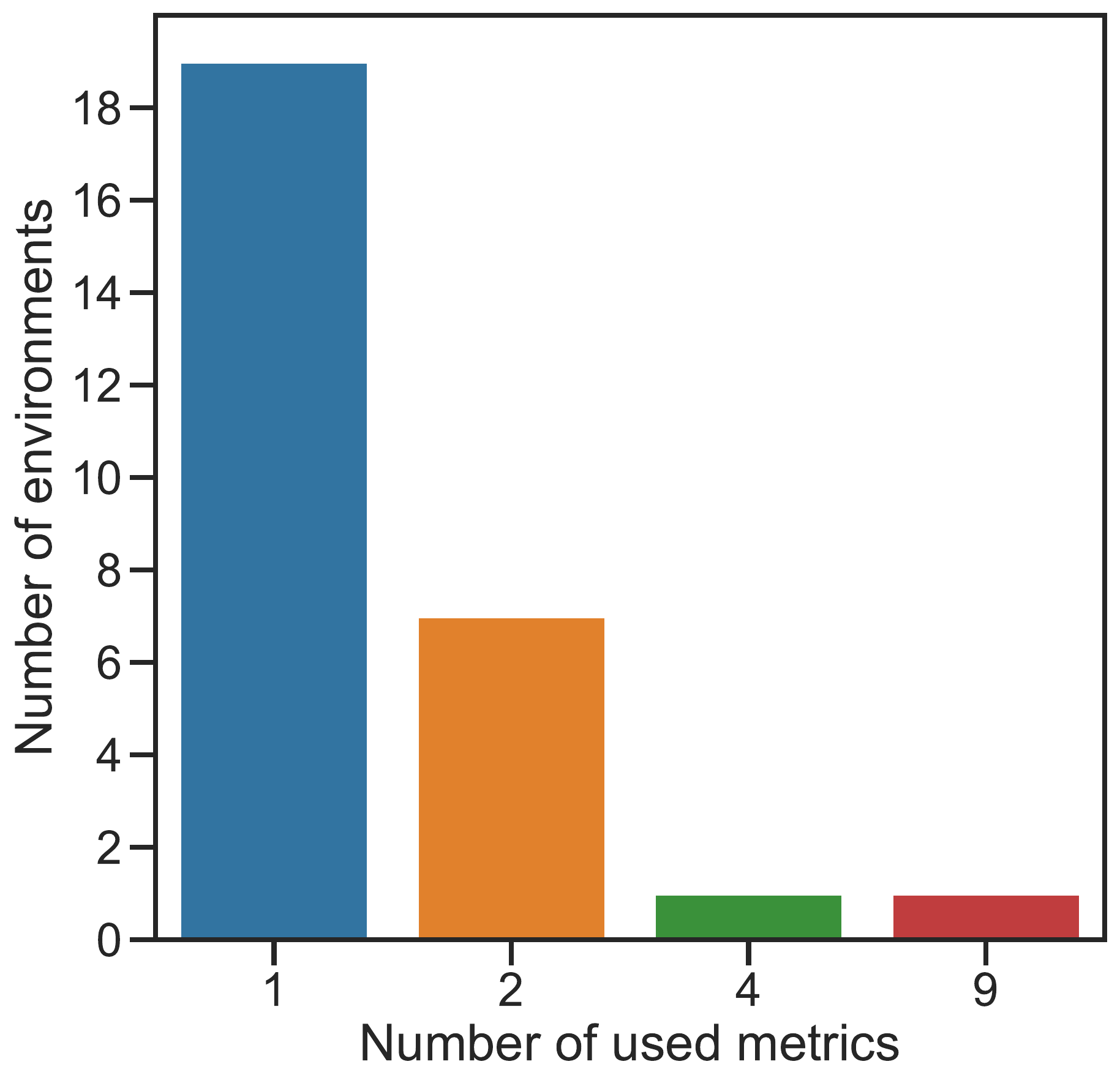}
    \caption{Number of metrics used per environment.}
\end{wrapfigure}
In general, metrics are used to monitor and quantify a model's performance. Through our analysis, we identify 25 unique metrics, and after unifying the data, based on our annotations as given in A4, we obtain 12 general metrics over the published papers.

The most common three metrics, referred to in our data are,  \textbf{Return}, \textbf{Reward} and \textbf{Win Rate} which are in 31.3\%, 14.6\% and 50\% of the main papers respectively. It is interesting to note that \textbf{Win Rate} is such a widely used metric, especially since it is environment specific. We believe this high percentage is due to the high use of the SMAC and Traffic junctions environments which commonly use Win Rate. 

We observe dependencies between the choice of the environment and metrics. Out of the collected SMAC data from the main papers 80.9\% use \emph{Win Rate} as a metric, meanwhile in MPE, out of the 720 rows of collected data related to the MPE environment, 35\% use \emph{Return} and 25.1\% use \emph{Reward}. Moreover, out of 29 environments over the main papers, we find 19 use only one metric type.

\subsubsection{Independent runs}
Independent training runs can take place across different \textbf{random seeds}. In some experiments multiple runs are completed for each random seed, for a fixed set of random seeds. Fixing the random seed is an attempt to control some of the experiment's sources of randomness. The number of runs is important in determining the reliability of the evaluation. More independent runs provide more data which allows for authors to report more accurate measures of spread alongside the point estimates of algorithm performance. \\
The authors used to employ 10 to 20 runs in the Unit Micromanagement version of the StarCraft environment, but since StarCraft II (SMAC) emerged, authors tend to use only around 5 independent runs. This decline may be due to the environment being more computationally expensive to run. However, we argue, similarly to \citep{agarwal2021deep} for the importance of having 10 independent training runs for reliable confidence intervals.

\subsubsection{Aggregate function }
An aggregate function, also known as a measure of central tendency, is a single value that intends to portray information about multiple results by determining the central position among a group of various results. For aggregations over algorithm performance, we differentiate between two aggregation steps: the first, which we refer to as the \textit{local aggregate function}, denotes how aggregation is done across evaluation episodes/evaluation runs in a fixed training run. The second, is the \textit{global aggregate function}. This denotes how we aggregate across independent training runs.\\
The performance of MARL algorithms is often reported using a point estimate of some task performance metric, such as the \textbf{mean} and \textbf{median} aggregated over the independent training runs. The mean is the most frequently used aggregation function, accounting for over 41.7\% of all data gathered from the main papers. It was the only utilized aggregate function in the early years of our recorded dataset. Since 2019, we have seen the introduction of the median as an aggregate function, with the launch of SMAC, and it has became one of the most widely used aggregate functions in SMAC, with some limited use in MPE. The widespread use of the median as an aggregate function can be attributed to the evaluation guideline proposed by \citep{SMAC-2019-Samvelyan}.

\subsubsection{Measure of spread}
The measure of spread plays an important role in delivering first hand information about the experiment findings. It expresses how far apart values are in a distribution and it provides a measure of the variability of values obtained across different random seeds or runs. It also serves as a basic way to quantify the uncertainty in a reported point estimate.\\
In our study, we discovered that 26 out of 75 studies did not mention the measure of spread at all. In some cases this resulted from when performance is only measured over one run, in other cases this is due to a lack of reported details within a paper. In statistics, there are various fundamental measures of spread, the following are the most frequently encountered in our MARL dataset:

\textbf{Standard deviation: }is a common measure of dispersion of a set of values from their mean. The standard deviation will be modest if the values are clustered together. Widely dispersed values will result in a larger standard deviation.

\textbf{Confidence interval (CI): } These provides an estimated possible range for an unknown value. We can choose from a variety of confidence limitations, where some of the most frequent are a 95\% or a 99.5\% confidence interval.

\textbf{Inter-quartile range: }This is a measure of dispersion which has the advantage of not being impacted by outliers and is important when the researchers want to know where the majority of the findings fall. It is used in 10 papers out of the main ones and it is commonly used in SMAC presenting 41.1\% of the SMAC collected data over the published cooperative papers. 

\subsubsection{Time Measurement}
\textbf{Independent Variable:} The training and evaluation time is a vital feature that must be stated by the author for a fair comparison of studies. We identified 9 options to define the independent experiment variable from the main papers, but the most commonly used measure is \textbf{time-steps}, which is employed in 39 papers, followed by \textbf{episodes}, which accounts for 18 studies. We discover an imbalance where the independent variable being used is strongly related to the environment, with 88.3\% of SMAC collected data using time-steps as an independent variable and 39.2\% of MPE collected data using episodes.

\textbf{Evaluation intervals (evaluation frequency): }is generally associated with the SMAC evaluation protocol. It refers to the fixed number of time-steps \textit{T}, after which training is suspended, to be able to evaluate an algorithm for a fixed number of runs/epsiodes \textit{E}. During these evaluation runs agents are usually only allowed to act greedily and in a decentralized manner. The test win rate is the percentage of episodes \textit{e}, in \textit{E}, for which the agents defeat all enemy units within the time limit. Although this is predominantly employed in SMAC experiments, occurring in 13 out of the 37 main papers that use SMAC, this evaluation approach is also used in the MPE and Level-Based Foraging (LBF) environments, with 6 and 2 papers adopting this methodology in these cases, respectively. The evaluation frequency must ideally be associated with a duration \textit{E} which we record as the evaluation duration.

\textbf{Number of independent evaluations per interval (evaluation duration): }as the name indicates, this is the amount of evaluations that are performed at each evaluation interval. This detail is required if an evaluation frequency is given, but it may also be provided by itself in the case evaluation is performed of the entire duration of an experiment.

\clearpage

\subsection{Evaluation procedure, best practices and guideline }

In this section, we summarize, in Table \ref{tab:five}, the number of papers that abide by the key practices that are recommended in the main body of this paper. We also show what percentage of the main papers and other papers include each specific practice in their evaluation and reporting protocol.

\begin{table}[h]
  \caption{Number and percentage of papers recorded that follow the details of the recommended evaluation guideline.}
  \label{tab:five}
  \centering
  \resizebox{0.85\textwidth}{!}{\begin{minipage}{\textwidth} 
  \begin{tabular}{lllllll}
    \toprule
    \textbf{Evaluation and Implementation details} & \multicolumn{3}{c} \textbf{The main papers}  & \multicolumn{3}{c}  \textbf{The other papers} \\ 
    & \textbf{Yes} & \textbf{No} & \hspace{0.5em} \textbf{\%} & \textbf{Yes} & \textbf{No} & \hspace{0.5em} \textbf{\%} \\
    \midrule
    \rowcolor{Gray}
    \textbf{Experiment details} & & & & & &\\
    Evaluate on multiple Environments & 38 & 37 & 50.7\% & 18 & 19 & 48.6\% \\
    \rowcolor{Gray}
    Evaluate on multiple Scenarios & 65 & 10 & 86.7\% & 36 & 01 & 97.3\%\\
    \textbf{Evaluation procedure details} & & & & & &\\
    \rowcolor{Gray}
    Report the training time & 65 & 10 & 86.7\% & 26 & 11 & 70.3\%\\
    Report the independent runs & 53 & 22 & 70.7\% & 26 & 11 & 70.3\%\\
    \rowcolor{Gray}
    Report the global aggregate function & 54 & 21 & 72.0\% & 30 & 07 & 81.1\%\\
    Report the measure of spread & 49 & 26 & 65.3\% & 21 & 16 & 56.7\%\\
    \rowcolor{Gray}
    Report the evaluation interval (evaluation frequency) & 20 & 55 & 26.7\% & 09 & 28 & 24.3\%\\
    Report the number of evaluation runs (evaluation duration) & 26 & 49 & 34.7\% & 16 & 11 & 43.2\%\\
    \rowcolor{Gray}
    Use statistical tests & 01 & 74 & 01.3\% & 00 & 37 & \hspace{1em} -  \\
    \textbf{Guideline \& Best practices} & & & & & &\\
    \rowcolor{Gray}
    Training for 2M timesteps & 20 & 55 & 26.7\% & 07 & 30 & 18.9\%\\
    Train on-policy for 20M and off-policy for 2M timesteps & 02 & 73 & 02.7\% & 00 & 37 & \hspace{1em} -\\
    \rowcolor{Gray}
    Use independent evaluation episodes per interval with \textit{E} = 32 & 04 & 71 & 05.3\% & 01 & 36 & 02.7\%\\
    Evaluation every 10000 timesteps & 04 & 71 & 05.3\% & 03 & 34 & 08.1\%\\
    \rowcolor{Gray}
    Use Mean Return metric & 14 & 61 & 18.7\% & 05 & 32 & 13.5\%\\
    Use Absolute metric & 02 & 73 & 02.7\% & 00 & 37 & \hspace{1em} -  \\
    \rowcolor{Gray}
    Use 95\% CI as a measure of spread & 16 & 59 & 21.3\% & 02 & 35 & 05.4\%\\
    Report plot results & 71 & 04 & 94.7\% & 33 & 04 & 89.2\%\\
    \rowcolor{Gray}
    Report tabular results & 40 & 35 & 53.3\% & 27 & 10 & 73.0\%\\
    Ablation study & 33 & 42 & 44.0\% & 18 & 19 & 48.6\%\\
    \rowcolor{Gray}
    Same baseline algorithms over all the experiment's environments & 54 & 21 & 72.0\% & 27 & 10 & 73.0\%\\
    Aggregate over the different maps and/or environments & 08 & 67 & 10,7\% & 02 & 35 & 05.4\%\\
    \rowcolor{Gray}
    Public repository & 36 & 39 & 48.0\% & 13 & 24 & 35.1\% \\
    \bottomrule
  \end{tabular} \end{minipage}}
\end{table}

\subsection{About SMAC}
\begin{wrapfigure}{r}{0.5\textwidth} 
    \centering
    \includegraphics[width=0.5\textwidth]{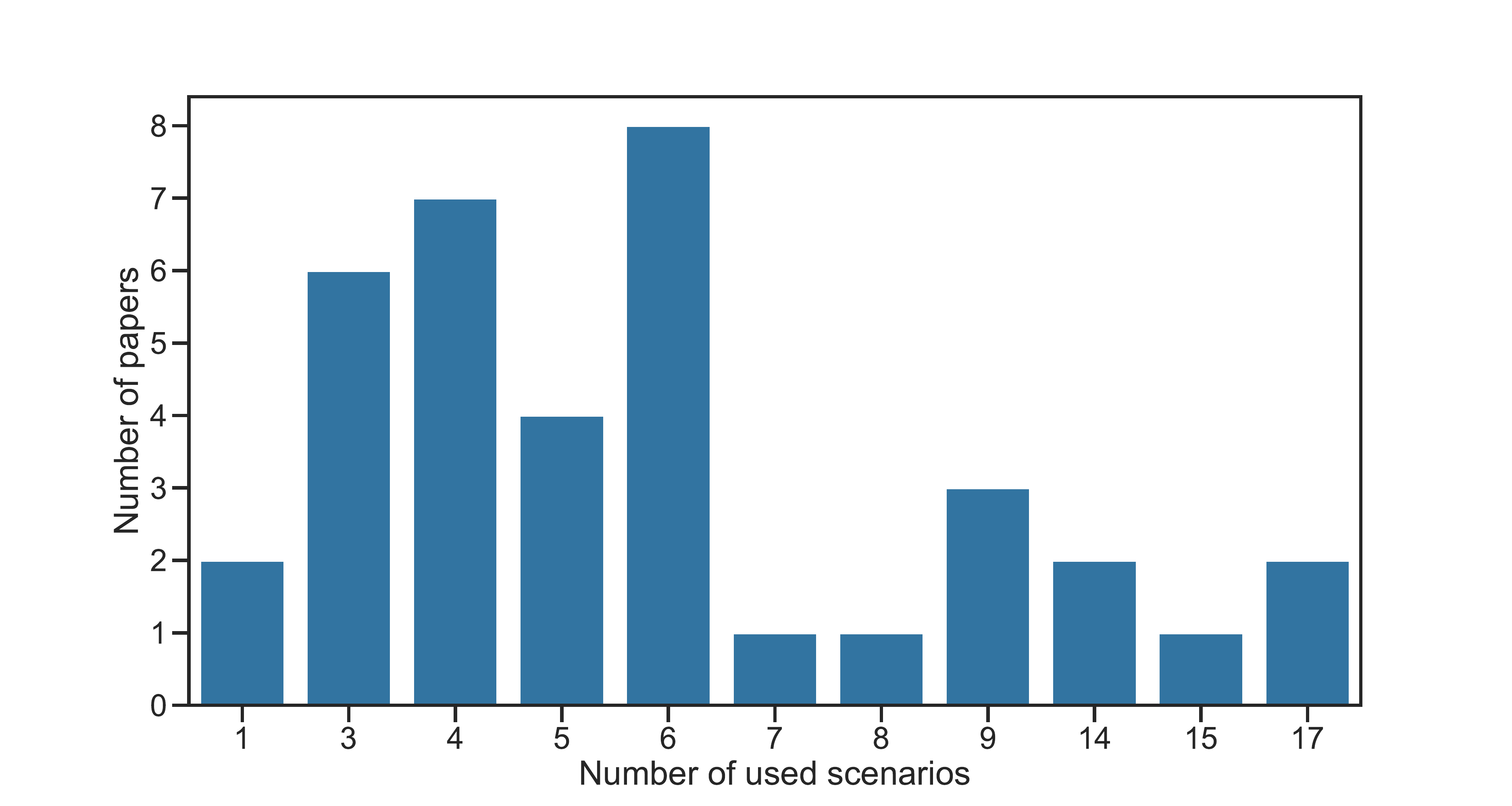}
    \caption{Number of used SMAC scenarios per paper}
    \label{fig:smac}
\end{wrapfigure}

In this section we will raise challenges uncovered in our analysis rather than provide answers. All challenges that are highlighted will be accompanied by all of relevant facts, as found in our dataset, and are from the SMAC benchmark.
We are not attempting to criticize the current scenarios or the environment itself, but want to emphasize the need of advocating for the use of SMAC to be standardized such that algorithm designers are limited to specified scenarios when testing their algorithms. The reason for this is to ensure fair comparison between works.\\
As we indicated in the environment section, SMAC is a popular benchmarking environment and we discovered that 13 publications out of 75 apply only SMAC to prove the trustworthiness of their experiment. We notice that, despite the fact that SMAC provides many testing scenarios (39 used ones in the published papers), most publications only employ a few of them in their reported trials, as seen in figure \ref{fig:smac}.

\begin{figure}[hbt!]
\centering
\includegraphics[width=.8\textwidth]{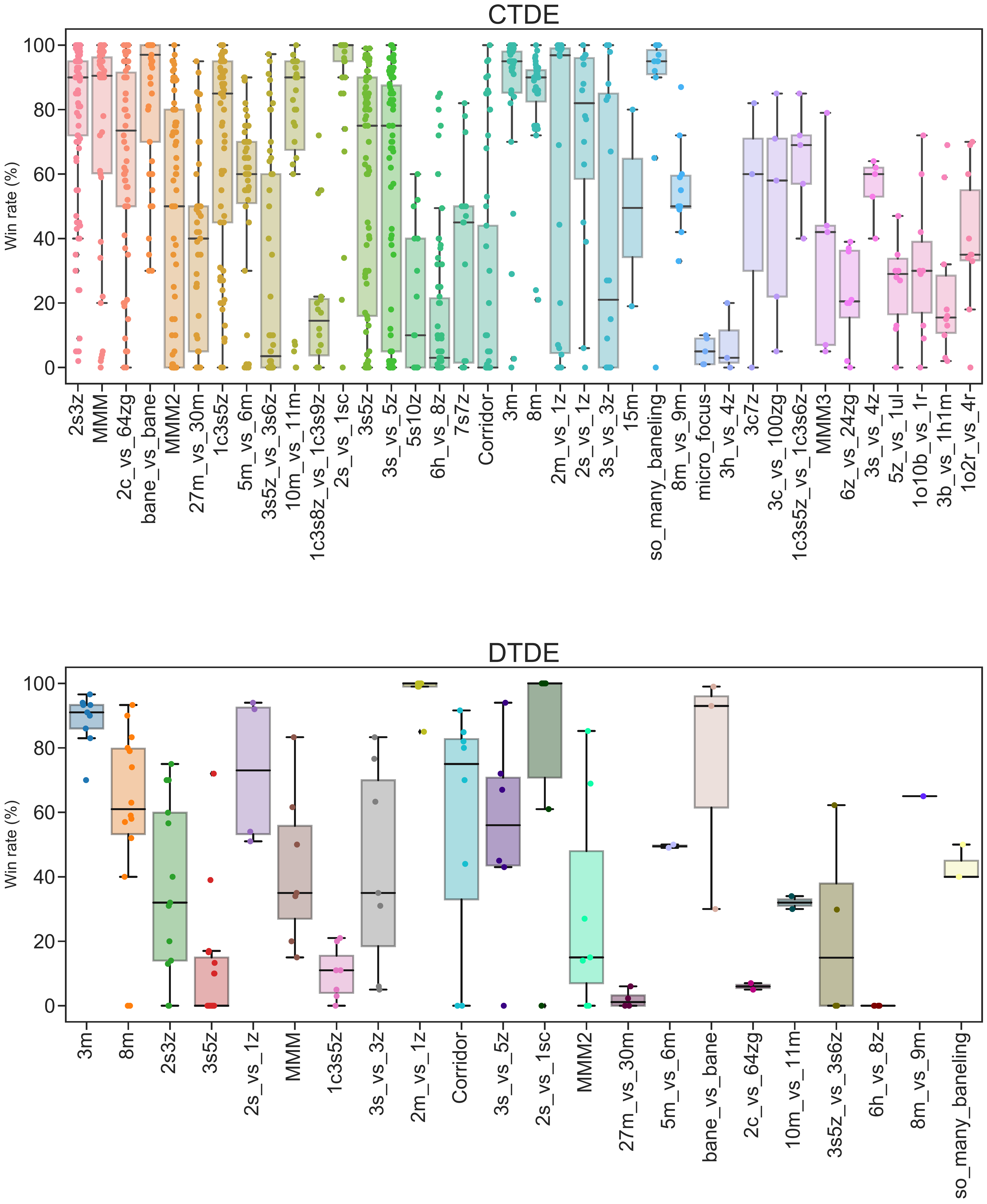}
\caption{SMAC win rate distribution based on training schemes from the main papers}
\label{fig: smac_ctde_dtde}

\end{figure}

\break
\textbf{What are the features needed to define the difficulty of a scenario?} After analyzing the win rate distribution under various settings, we discovered that several scenarios that were thought to be difficult turned out to be simple through using independent learning algorithms. A clear example of this is illustrated in Figure \ref{fig: smac_ctde_dtde} by the shift in the win rate distribution for CTDE and DTDE algorithms evaluated on the corridor scenario.

\break

\begin{figure}[hbt!]
\centering
\includegraphics[width=.6\textwidth]{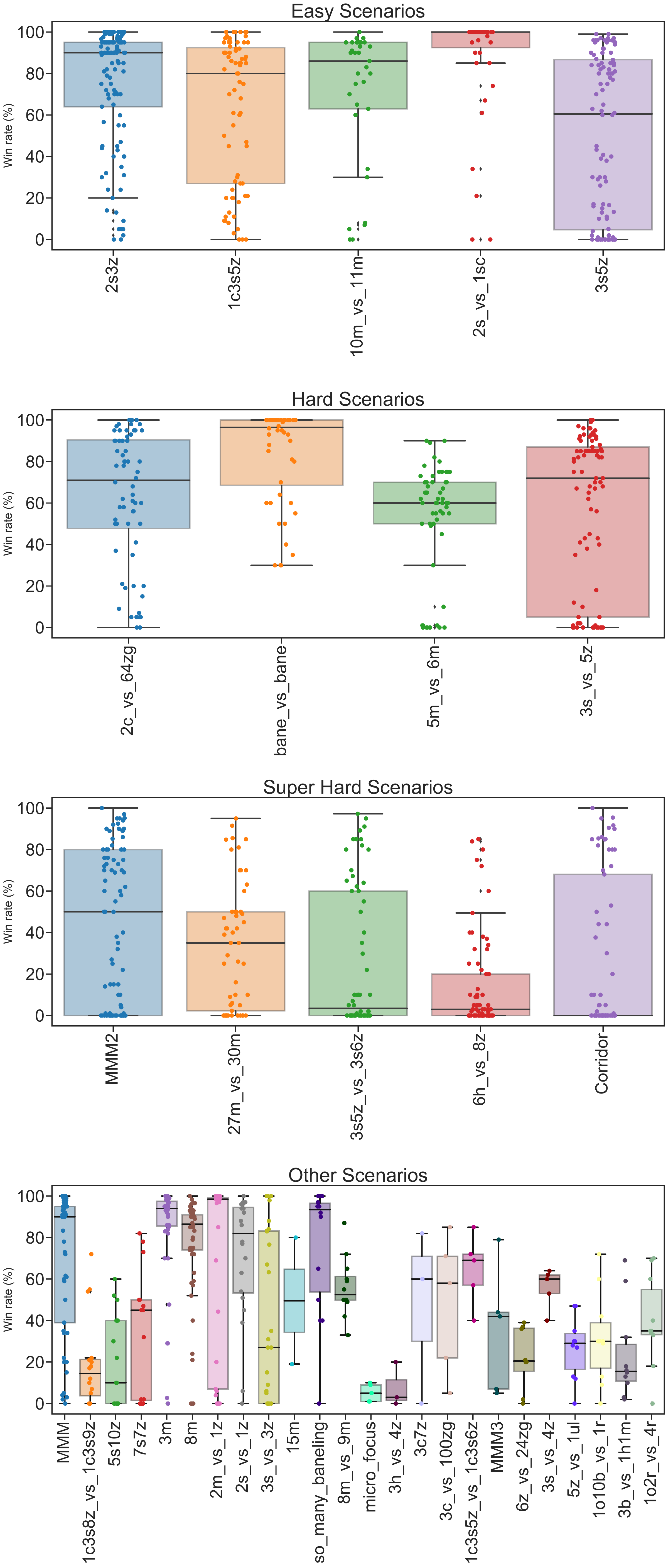}
\caption{SMAC win rate distribution based on difficulty from the main papers}
\end{figure}
\break 

\begin{figure}[h]
\centering
\includegraphics[width=.75\textwidth]{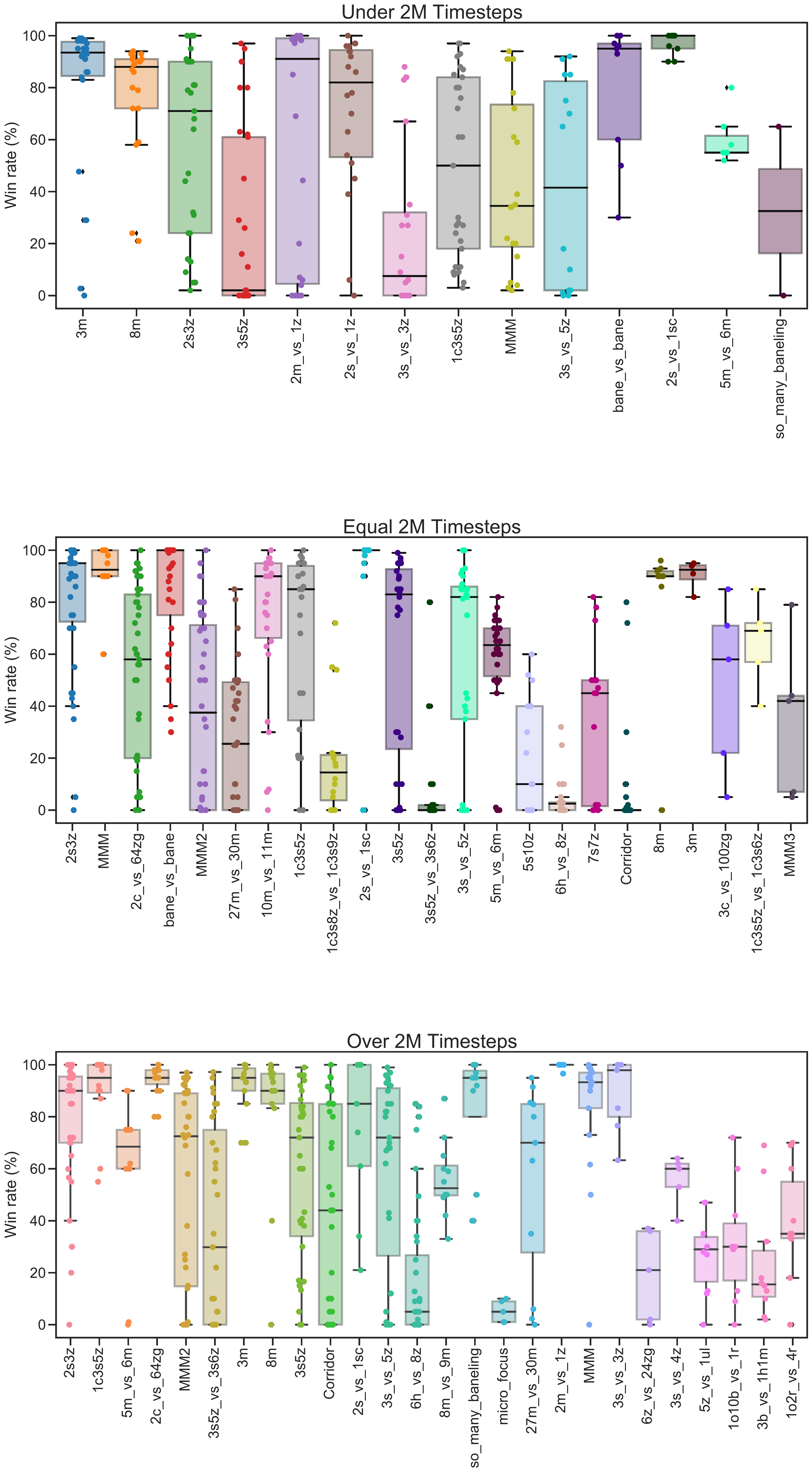}
\caption{SMAC win rate distribution based on training time from The main papers}
\label{SMAC:time}
\end{figure}

Furthermore, figure \ref{SMAC:time} emphasizes the importance of training until 2M timesteps. It demonstrates how the win rate, for even the easiest scenarios, has a wide spread when algorithms are trained for less than 2 million time steps. It can also be noted that, when algorithms are trained up to 2 million timesteps or more, that performance convergences to a higher win rates, not only for easy scenarios but also for hard and even super hard ones.

\break

\textbf{Which scenario to choose?} The choice of the scenarios for an algorithm designer is a critical task, considering the fact that each scenarios itself in SMAC has its own challenges, which can work in the algorithm's favor (e.g. IA2C in Corridor) or in its misfortune (e.g. IA2C in MMM2). Moreover, 50\% of the scenarios were used in one or two papers only, some of these scenarios were used for ablation studies or for a specific research direction like communication, nevertheless most of them do not have prior justification. 

\begin{figure}[h]
\begin{subfigure}{1\textwidth}
\centering
\includegraphics[width=0.45\textwidth]{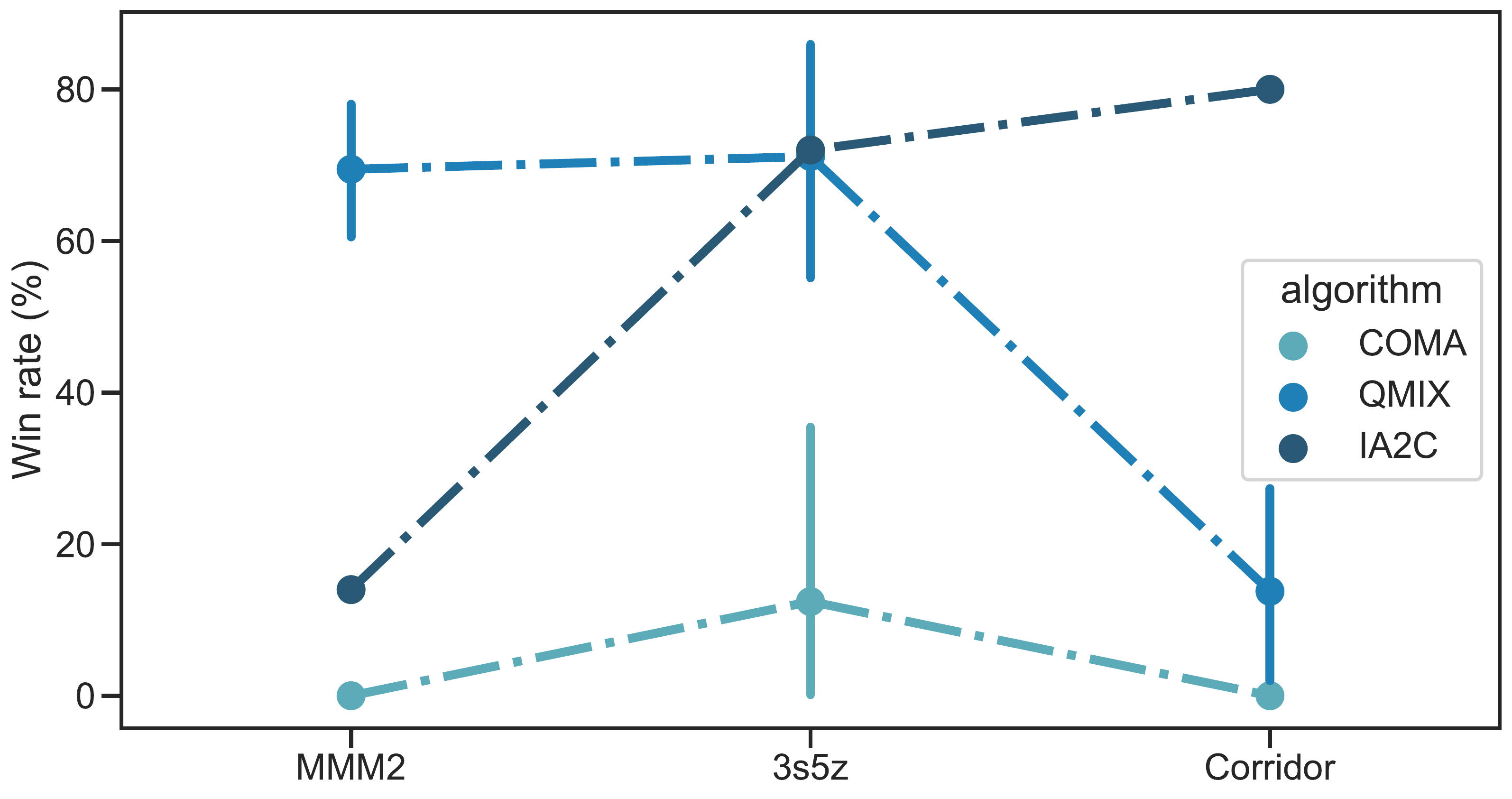}
\end{subfigure}
\begin{subfigure}{1\textwidth}
\centering
\includegraphics[width=0.45\textwidth]{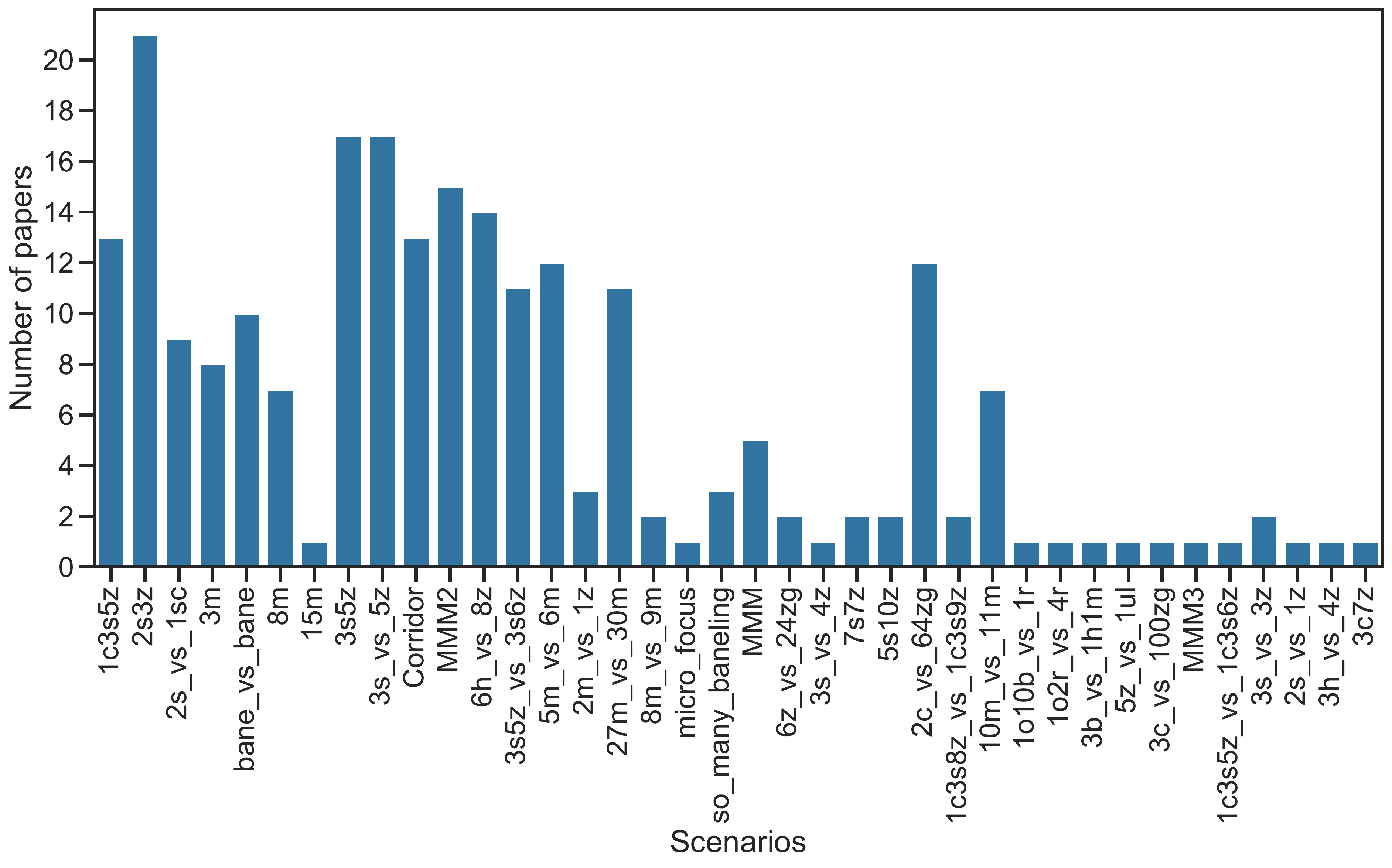}
\end{subfigure}
\caption{\textbf{Top:} The performance of COMA, QMIX and IA2C in 3 different SMAC scenarios. \textbf{Bottom:} Number of papers that use each scenario over the main papers}
\end{figure}

\begin{figure}[h]
\centering
\includegraphics[width=0.65\textwidth]{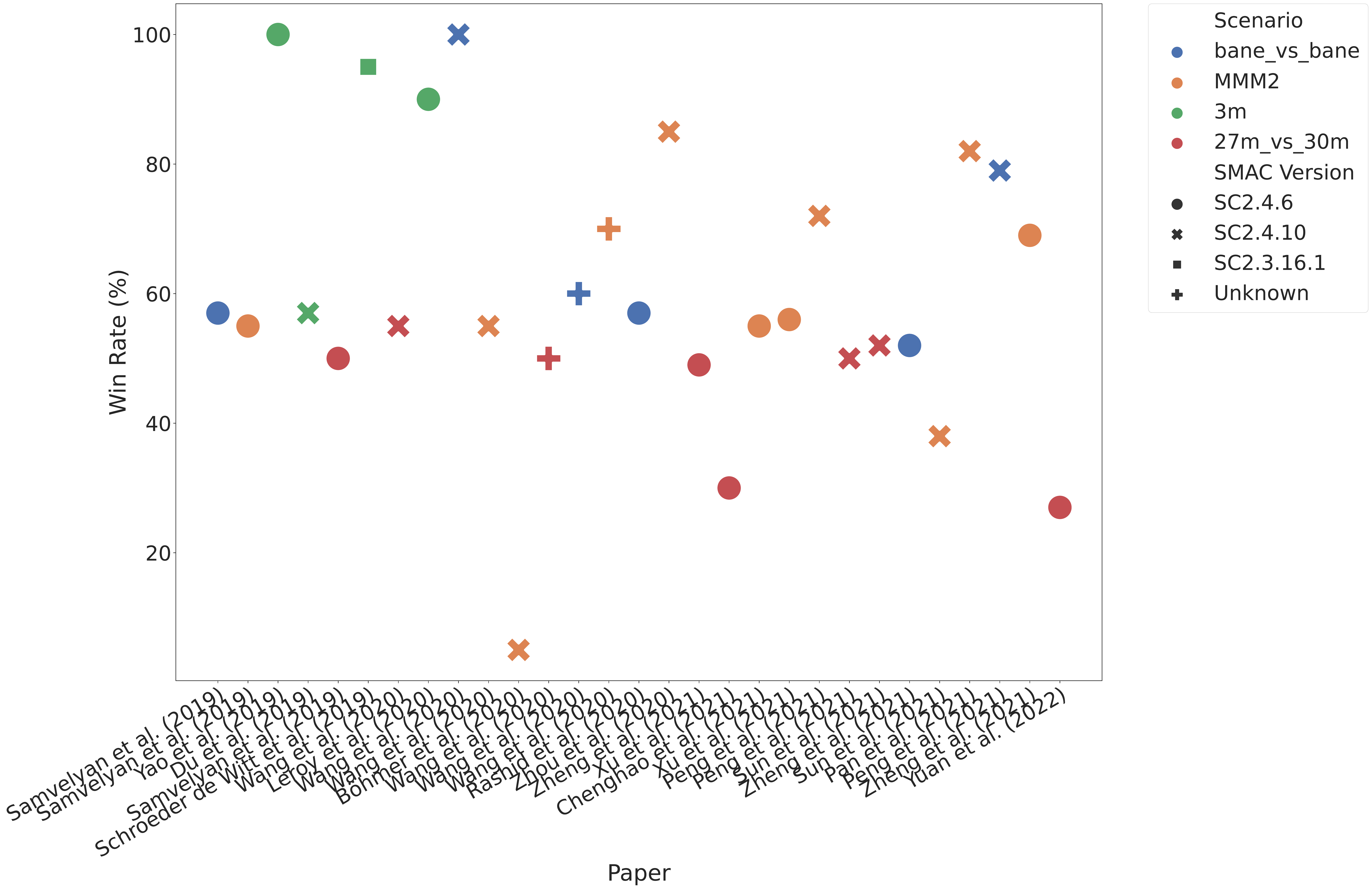}
\caption{Performance of QMIX on different SMAC scenarios trained for 2M timesteps}
\label{SMAC:subsampling}
\end{figure}

\textbf{Is the inconsistency in performance inescapable?} In Figure \ref{SMAC:subsampling}, we fixed the training steps to be 2 million for all recorded papers that use the bane vs bane, MMM2, 3m and 27m vs 30m SMAC scenarios. We achieve this by reading algorithm performance from plots produced in all relevant papers. It is known that the version of SMAC that is used can have an effect on algorithm performance, but here we see that merely fixing the training time steps across multiple papers leads to even greater performance discrepancies between papers than the SMAC version being used.

\section{Guideline}
\subsection{Motivation}
In the following section, we will demonstrate our evaluation guideline. We follow the steps as outlined but omit ablation studies in our work since we are not trying to introduce a novel algorithm. We also have not tuned any hyperparameters for any of the algorithms we consider. We wish to make the reader purposefully aware that the primary goal of this experiment is to provide an illustration of how to use our evaluation guideline and our experiment is not focused on the performance of the chosen algorithms. As such we are not striving to achieve state of the art performance on the flatland benchmark and will merely illustrate how results may be interpreted by a researcher. 

Please note that, following our guideline, we make all raw data  of our experiments available. Our code will be made publicly available soon. We provide a LaTeX template for the proposed reporting templates \ref{tab:reporting-proposal}. We envision that such a template will make it easier for other authors to define and report all details pertaining to their experiments. These include experimental details, evaluation protocols, environment  settings and all other details that authors wish to report. Of course, it's up to the author to choose the set of hyperparameters to be reported according to the algorithm class and its specific hyperparameters.

\subsection{Reporting templates}
Here, we present an example of a template that can be used to summarise the important information required to perform the evaluation of algorithms, see Table \ref{tab:reporting-proposal}.

Firstly, we have to list all the algorithms we are comparing. 
For hyperparameters we suggest listing all tunable parameters which are manually set by the researcher. Some parameters like the \textit{discount factor} are fairly consistent throughout published works and can easily be reused across papers. Other parameters can vary through papers due to computational constraints like \textit{batch size} which can be limited by the available memory of the GPU used in training and sometimes needs to be adjusted based on computational limitations and, \textit{replay buffer size} which is limited by the available memory of the training computers. Parameters like the \textit{target network update period} and \textit{$\epsilon$ schedule} are required to replicate the training scheme used by an algorithm as when mistuned they greatly alter results \cite{rashid-2021-weightedQMix}. 

Network architecture is also an important consideration for MARL algorithms. QMIX is one of the most popular value-decomposition methods in cooperative MARL and makes use of \textit{hypernetworks} to train the central value function. The parameterisation of these \textit{hypernetworks} must also be noted as their configurations have a strong effect on the effectiveness of the value function. The central value function for QMIX inspired value-decomposition methods is commonly called a \textit{mixing network} and is responsible for performing multi-agent credit assignment during training. The \textit{mixing networks} can vary across different methods but, without knowing how they are parameterised it is possible for networks to have a large variance in their complexity which makes direct comparisons difficult to interpret \cite{hu2021rethinking} 

Additionally not all methods consistently make use of recurrency in their architecture which is important for achieving high performance in partially observable settings.
Parameter sharing is also unique the cooperative MARL setting and is used in most publications, however, not all papers make use of this paradigm.
 
Code-level optimisations consist of any parameters that can be included in algorithm implementations for improving performance, but are not core components of the algorithm. \textit{Reward normalisation} is when the rewards over the episode are normalised which reduces variance and makes learning easier \cite{yu2021}. Not all settings make use of normalised rewards but they can be trivially implemented at the code level. \textit{Death masking} is important to note as different frameworks deal with dead agents in different manners which can make direct comparisons difficult. \textit{Clipped updates} are used in some papers to prevent exploding gradients and can be trivially implemented in most deep learning frameworks. \textit{Eligibility traces} can be used to adjust the variance and bias trade off for return calculations and are tuned using the $\lambda$ parameter. Although using $TD(\lambda)$ returns has been shown to improve performance for MARL algorithms it is not universally used and must be taken into account for evaluation. Optimiser choice has also been shown to have a large impact on the performance of MARL algorithms and cannot be interchanged arbitrarily (\cite{yu2021}).

Computational resources, although not important for algorithmic development are still relevant to research. Clarity of the resources required for a publication to be replicated provide an indication to researchers as how feasible replication is and, how similarly optimised their own implementations are. It also makes it clear where methods may perform better at the cost of compute.

Evaluation protocols need to be made clear in publications so that the results are easy to interpret. By providing all evaluation in the template details readers do not need to pick through a paper to determine how to interpret results. The evaluation framework and the version that is being used is also of importance. Evaluation frameworks are frequently updated and results might be incomparable in-between versions.

Finally it is important to provide the configurations of the environments being used to train and evaluate the algorithm. On one hand, in sample evaluation allows to evaluate an algorithm's performance on an environment configuration similar to the configurations it was trained on. On the other hand, out-of sample configurations help to test the ability of the algorithm to generalise to a different configuration of the environment that was not seen in the training. It is obvious that there are many standardised settings in MARL. There are also cases of publications using custom environments which are non-standard when compared to existing publications. These non-standard settings require a full detailing of specifications to make them easier to understand.

\begin{table}[H]
  \caption{Proposal for reporting experimental details}
  \label{tab:reporting-proposal}
  \centering
    \begin{center}
\centerline{\scalebox{0.80}{
  \begin{tabular}{llll}
    \toprule
    \midrule
    \rowcolor{lightgray}
    \textbf{Experimental setup} &  \hspace{2em} \textbf{Algo 1} &  \hspace{7em} \textbf{Algo 2} &  \hspace{3em} \textbf{Algo 3} \\
    \midrule
    \rowcolor{Gray}
    \textbf{Hyperparameters} & & &   \\
    \midrule
    Discount factor & & & \\
    Batch size  & & & \\
    Replay buffer size  & & & \\
    Minimum replay buffer size before updating  &  & &\\
    N steps bootstrapping  & & &\\
    Target network update period  & & &\\
    $\epsilon$ schedule (Decay steps, $\epsilon$ start, $\epsilon$ min)  & & &\\
    Value Network architecture  & & & \\
    Value Network initializer  & & & \\
    Value Network Layer size  & & &\\
    Value Network Layer normalisation  & & & \\
    Mixing network (architecture, size, activation) & & & \\
    Hypernetworks (size, activation)  & & &\\
    Parameter sharing  & & &\\ 
    Parallel workers  & & &\\
    Seed range  & & &\\
    \midrule
    \rowcolor{Gray}
    \textbf{Code-level optimisations}  & &  &\\
    \midrule
    Optimiser (type, parameters)  & & &\\
    Learning rate  & & &\\
    Reward normalisation  & & &\\
    Death masking  & & &\\
    Clipped updates  & & &\\
    Eligibility trace  & & &\\
    $TD(\lambda)$ value & & &\\
    \midrule
    \rowcolor{Gray}
    \textbf{Computational resources} & & & \\
    Average Wall-clock time per algorithm  & & & \\
    CPUs per experiment  & & & \\
    GPU per experiment  & & & \\
    RAM per experiment  & & & \\
    \midrule
    \rowcolor{lightgray}
    \textbf{Evaluation protocol}  & & &  \\
    \midrule
    Total training (timesteps)  & & & \\
    Evaluation interval (timesteps)  & & & \\
    Independent evaluation episodes  & & & \\
    Absolute metric (evaluation episodes, aggregation method)  & & & \\
    Local aggregation method  & & & \\
    Global aggregation method  & & & \\
    Metrics [Environment 1 name ]    & & & \\
    Metrics [Environment 2 name ]    & & & \\
    Metrics [Environment 3 name ]    & & & \\
    Exploration behaviour   & & & \\
    \midrule
    \rowcolor{lightgray}
    \textbf{MARL Framework}& \multicolumn{3}{c}{name (version)} \\
    \midrule
    \rowcolor{lightgray}
    \textbf{Environment settings} & & &  \\
    \midrule
    \rowcolor{Gray}
    \textbf{Environment 1 name (version)} &  \textbf{Training}  & \textbf{In sample evaluation configs}  & \textbf{Out of sample evaluation configs}\\
    \midrule
    Env related configs & & &\\
    \midrule
    \rowcolor{Gray}
    \textbf{Environment 2 name (version)} &  \textbf{Training}  & \textbf{In sample evaluation configs}  & \textbf{Out of sample evaluation configs}\\
    \midrule
    Env related configs & & &\\
    \rowcolor{Gray}
    \textbf{Environment 3 name (version)} &  \textbf{Training}  & \textbf{In sample evaluation configs}  & \textbf{Out of sample evaluation configs}\\
    \midrule
    Env related configs & & &\\
   \bottomrule
 \end{tabular}}}
 \end{center}
 \end{table}

\subsection{Experiment details}

Firstly we note the algorithms used for the experiments. For illustration purposes we use IQL which is an independent learning algorithm, VDN which is a linear value-decomposition method and finally QMIX which is a value-decomposition method that makes use of a central value function. It is important to note that not all parameters are applicable to all types of algorithms.

\subsubsection{Environment}

An environment can present various factors of variations forming two different context sets: the first being the set of all supported random seeds which makes use of Procedural Content Generation (PCG) and the second is the product of multiple factors of variations inside the environment. It has been noted that procedurally generated environments may reduce the precision of research \cite{kirk2021survey} while being able to control a factor of variations in an environment offers more flexibility to create environment configurations that match the evaluation of different algorithmic strengths. Regardless of the context being used, we strongly advocate that researchers should report all the environment settings used for training and for evaluation, see the environment settings section in Table \ref{tab:reporting-proposal}, as an example of reporting environment settings. Of course, all settings are environment specific.

For our experiments, we make use of the Flatland benchmark environment \cite{mohanty2020flatland} first introduced as a challenge in 2020 to investigate solutions to the vehicle rescheduling problem in railway systems. At a high-level, Flatland is a highly customisable, simplified 2D grid environment which aims to simulate the routing of trains from one city to another. 

\begin{figure}[H]
\centering
\includegraphics[width=0.9\textwidth]{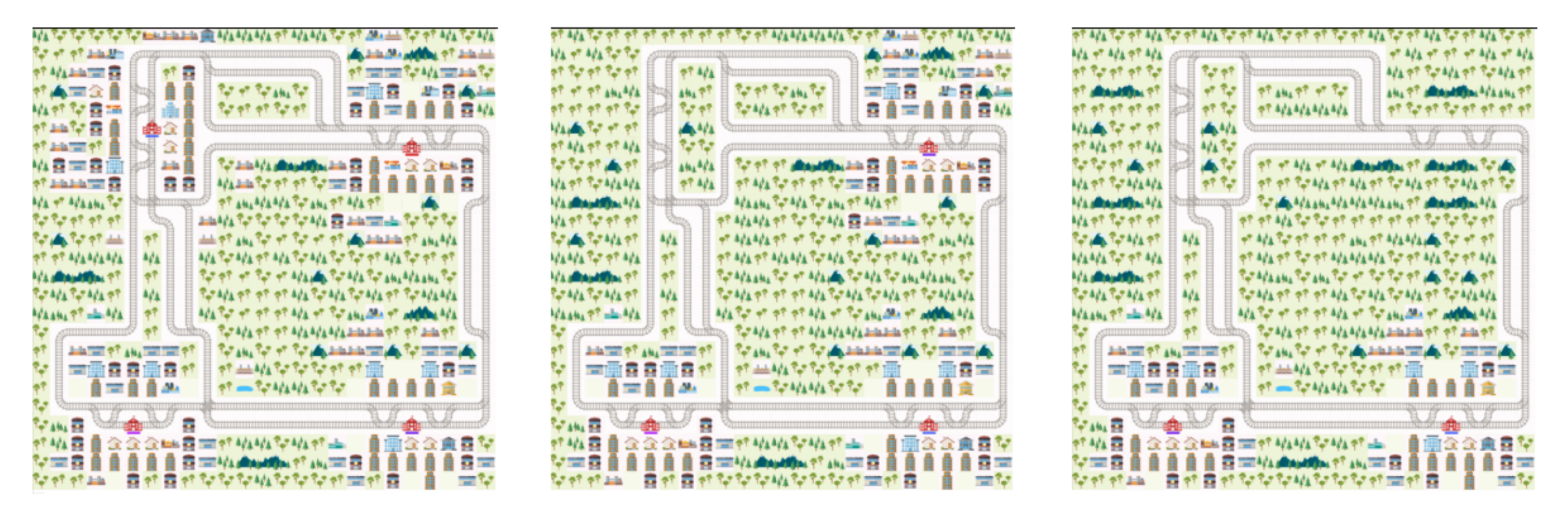}
\caption{Flatland maps varying between consecutive environment episodes.}
\label{fig: flatland_episode_gen}
\end{figure}

We particularly choose Flatland as a benchmarking environment due to the fact that the environment may be set to be non-static allowing it to change after each completed episode during both training and evaluation. This enables us to test the ability of algorithms to generalise outside of experience that was encountered during training. Flatland offers a high level of customisability with regards to this environment regeneration, but we opt for a relatively limited and simple approach. Once a map has been generated, we keep the rails of the map fixed but we allow the number of stations on the map to be randomly distributed at each new episode. This changes the location on the map where an agent starts at each episode as well as the destination that each agent must reach. An example of how the maps might change over 3 episodes is demonstrated in Figure \ref{fig: flatland_episode_gen}.

\textbf{Observation space.}
For the observations of each agent, we make use of what the Flatland authors refer to as tree observations. For these observations, an agent is allowed to construct a tree in four directions which follows permitted transitions. These trees are allowed to pass a fixed number of points on the grid where more than one action is allowed with these points being referred to as \textit{switches} by the authors. Each agent is then allowed to observe the grid up to a fixed number of switches (referred to as the maximum permitted tree depth) and then constructs local features based on the observed tree. These features then inform the agent's decision making.

For all algorithms, each agent only makes use of its own local tree observation to inform its action selection. Since Flatland does not return a global state for the entire grid at each training time step, on which QMIX can condition its mixing network during training time, we construct a simple global state representation which is the concatenation of all agents' local observations. 

\textbf{Action space.}
The action space in flatland is $discrete(5)$ and consists of the following actions: 
\begin{itemize}
    \item{Move forward},
    \item{Select a left turn},
    \item{Select a right turn},
    \item{Halt on the current cell},
    \item{Take no action}.
\end{itemize}

\textbf{Reward structure.} At each environment time step each agent, $i$ receives a reward calculated as: 
$$ r_i(t)= \alpha r_{l}^{i}(t) + \beta r_g(t) $$

Here $r_{l}^{i}$ denotes an agent's local reward which is $-1$ for all time steps until an agent reaches its destination after which it is $0$ until episode termination. $r_g$ denotes an additional team reward of $1$ which is received by all agents when all agents have reached their destination during an episode. $\alpha$ and $\beta$ are adjustable parameters which govern agent cooperation. 

After an episode is completed each agent receives a return $g_i$ which is computed as:

$$g_i = \sum_{t=1}^T r_i(t) $$

In order to keep track of team performance, we monitor and report the mean team episode return which may be calculated for $N$ agents as 

$$ g_t = \frac{1}{N} \sum_{i}^{N} g_i $$

Aside from only keeping track of the team return, we also record the team completion rate which is the proportion of agents that were able to reach their destination in a given episode.

\subsection{Evaluation protocol and experimental procedure}
Detailed flatland experiment settings are given in Table \ref{tab:reporting-flatland}. We perform 10 independent runs, each with a unique random seed for the initialization of the agent policy networks. For each independent run we evaluate algorithm performance for 32 episodes at every 10000 environment time steps. During these evaluation intervals  we freeze training such that agent policy network weights remain fixed and agents are only allowed to act greedily by selecting actions which an agent believes to have to highest Q-values. In order to report the overall team performance, we report the mean return and completion over of all agents in the environment at each episode. It should be noted that in Flatland agents receive their reward at the end of an episode and therefore episode returns and rewards are equivalent. In order to normalize the episode returns we keep track of the maximum and minimum return obtained over all evaluation episodes done during training for a given independent run and then normalize the mean episode return of each evaluation episode according to these global maximum and minimum values. In order to obtain the per task results, we compute the mean and 95\% CI over all independent runs at each evaluation interval for both the normalised mean returns and the completion rate. Additionally, for each independent training run, we keep track of both the maximum mean return and maximum completion rate computed at each evaluation interval and use these values to checkpoint the agent network parameters where performance for both these metrics are optimal. Once an independent run is complete we then evaluate the algorithm greedily for 320 episodes using the best model parameters found for both the mean episode return and completion rate and take the mean over these roll outs to compute the absolute metrics for both the completion rate and the mean episode return. In all cases we opt to use the mean instead of the inter-quartile mean since we assume there to be relatively few outliers due to the fact that all results are generated using the same fixed policy. For each independent run, we then normalise the absolute metrics across all algorithms that were being tested such that all absolute metrics fall within the range $[0, 1]$. In all cases it should be noted that, since the goal of normalisation is to constrain metrics to lie within the same $[0, 1]$ interval we omit normalising metrics that inherently lie on such a range, like the completion rate. Since we have only one task, we then construct a $(10 \times 1)$ vector per algorithm using the obtained normalised metrics in order to make use of the tools provided by \citep{agarwal2021deep} and obtain the following results.

\begin{table}[H]
  \caption{Reporting Flatland experimental details}
  \label{tab:reporting-flatland}
  \centering
    \begin{center}
\centerline{\scalebox{0.95}{
\begin{tabular}{llll}
    \toprule
    \midrule
    \rowcolor{lightgray}
    \textbf{Experimental setup} &  \textbf{IQL} & \textbf{QMIX} & \textbf{VDN}  \\
    \midrule
    \rowcolor{Gray}
    \textbf{Hyperparameters} & & &  \\
    \midrule
    Discount factor & 0.99 & 0.99 & 0.99\\
    Batch size & 32 & 32 & 32\\
    Replay buffer size &  5000 & 5000 & 5000\\
    Minimum replay buffer size before updating & 32 & 32 & 32\\
    N steps bootstrapping & 5 & 5 & 5\\
    Target network update period & 100 & 200  & 200\\
    $\epsilon$ schedule (Decay steps, $\epsilon$ start, $\epsilon$ min) & (100000,1.0,0.05) & (100000,1.0,0.05) & (100000,1.0,0.05)\\
    Value Network architecture & Recurrent & Recurrent & Recurrent \\
    Value Network initializer & Variance Scaling & Variance Scaling & Variance Scaling \\
    Value Network Layer size & [64,64] GRU & [64,64] GRU & [64,64] GRU\\
    Value Network Layer normalisation & True & True & True \\
    Mixing network (architecture, size, activation) & - & Feedforward,[32], ReLU & - \\
    Hypernetworks (size, activation) & - & [64], ReLU & - \\
    Parameter sharing & Yes & Yes & Yes \\ 
    Parallel workers & 8 & 8 & 8\\
    Seed range & \{0..9\} & \{0..9\} & \{0..9\} \\
    \midrule
    \rowcolor{Gray}
    \textbf{Code-level optimisations} & & & \\
    \midrule
    Optimiser (type, parameters) & Adam & Adam & Adam\\
    Learning rate & 1e-4 & 1e-4 & 1e-4\\
    \midrule
    \rowcolor{Gray}
    \textbf{Computational resources} & & & \\
    \midrule
    Average Wall-clock time per algorithm & 9h27m & 9h36m & 9h16m \\
    CPUs per experiment & \multicolumn{3}{c}{20} \\
    GPU per experiment & \multicolumn{3}{c}{1} \\
    RAM per experiment & \multicolumn{3}{c}{20 GB} \\
    \midrule
    \rowcolor{lightgray}
    \textbf{Evaluation protocol} & & &  \\
    \midrule
    Total training (timesteps) & \multicolumn{3}{c}{2000000} \\
    Evaluation interval (timesteps) & \multicolumn{3}{c}{10000} \\
    Independent evaluation episodes & \multicolumn{3}{c}{32} \\
    Absolute metric (evaluation episodes, aggregation method) & \multicolumn{3}{c}{320, Mean with normal 95\% CI} \\
    Local aggregation method & \multicolumn{3}{c}{Mean} \\
    Global aggregation method & \multicolumn{3}{c}{IQM with 95\% stratified bootstrap CI} \\ 
    Metrics [Flatland] &   \multicolumn{3}{c}{Return, Completion rate, Normalised score} \\
    Exploration behaviour  & \multicolumn{3}{c}{Disabled} \\
    \midrule
    \rowcolor{lightgray}
    \textbf{MARL Framework}& \multicolumn{3}{c}{MAVA (0.1.2)} \\
    \midrule
    \rowcolor{lightgray}
    \textbf{Environment settings} & & & \\
    \midrule
    \rowcolor{Gray}
    \textbf{Flatland (3.0.15)} &  \textbf{Training} \tablefootnote{Using same generator config from \url{https://gitlab.aicrowd.com/flatland/neurips2020-flatland-baselines/-/blob/flatland-paper-baselines/envs/flatland/generator_configs/small_v0.yaml}}   & \textbf{In sample evaluation} &  \\
    \midrule
    Number of agents & 5  & 5 & \\
    Grid size (width x height) & 25x25 & 25x25 &\\
    Maximum number of cities & 4  & 4  & \\
    Maximum rails between cities & 2 & 2 & \\
    Maximum rails in city & 3 & 3 &  \\
    Malfunctioning rate & 0 & 0 &  \\
    Observation (type, depth) & TreeObservation, 2 & TreeObservation, 2 &  \\
    Shortest Path Predictor max depth & 30 & 30 &  \\
    Grid mode & True & True & \\
    Regenerate schedule on reset & True & True &  \\
    Regenerate rail on reset & True & True &  \\
    Seed & 0 & 0 &  \\
  \bottomrule
\end{tabular}}}
\end{center}
\end{table}
\subsection{Results}

All plots that are generated here are made using the tools provided by \citep{agarwal2021deep}. 



\subsubsection{Sample efficiency curves}
The sample efficiency curves serve as a way to asses an algorithm's experience efficiency at improving on a particular metric during training time. For two algorithms that achieve the same final performance on some metric, the algorithm that does so with less environment steps could therefore be considered to be more sample efficient. We compute the sample efficiency curves by making use of the normalized mean return at each evaluation interval as well as the mean completion rate achieved at each evaluation interval. 

\begin{figure}[H]
    \centering
    \begin{subfigure}[t]{0.02\textwidth}
        \scriptsize
        \textbf{(a)}
    \end{subfigure}
    ~
    \begin{subfigure}[t]{0.45\textwidth}
        \includegraphics[width=\linewidth, valign=t]{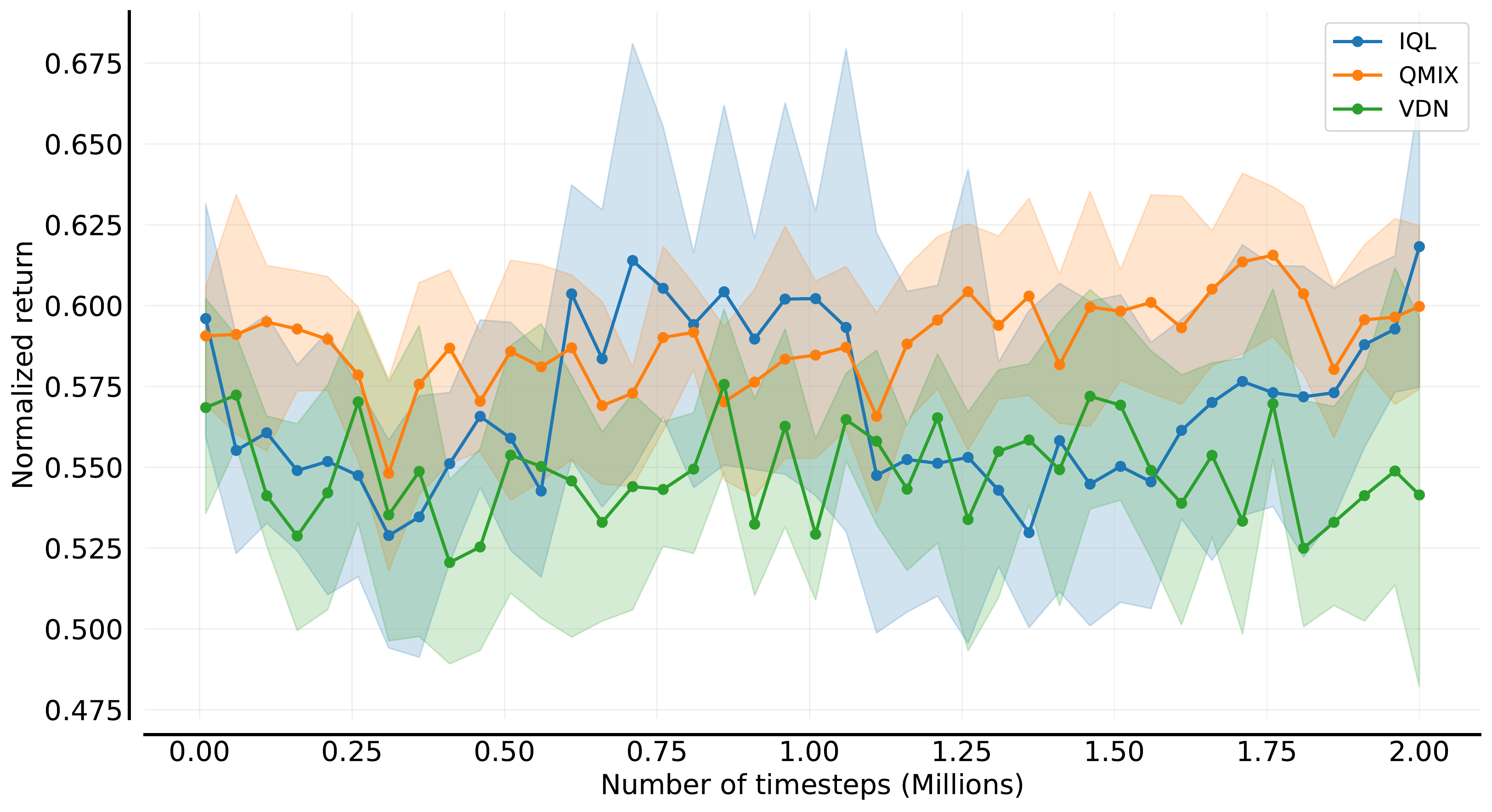}
    \end{subfigure}
    ~
    \begin{subfigure}[t]{0.02\textwidth}
        \scriptsize
        \textbf{(b)}
    \end{subfigure}
    ~
    \begin{subfigure}[t]{0.45\textwidth}
        \includegraphics[width=\linewidth, valign=t]{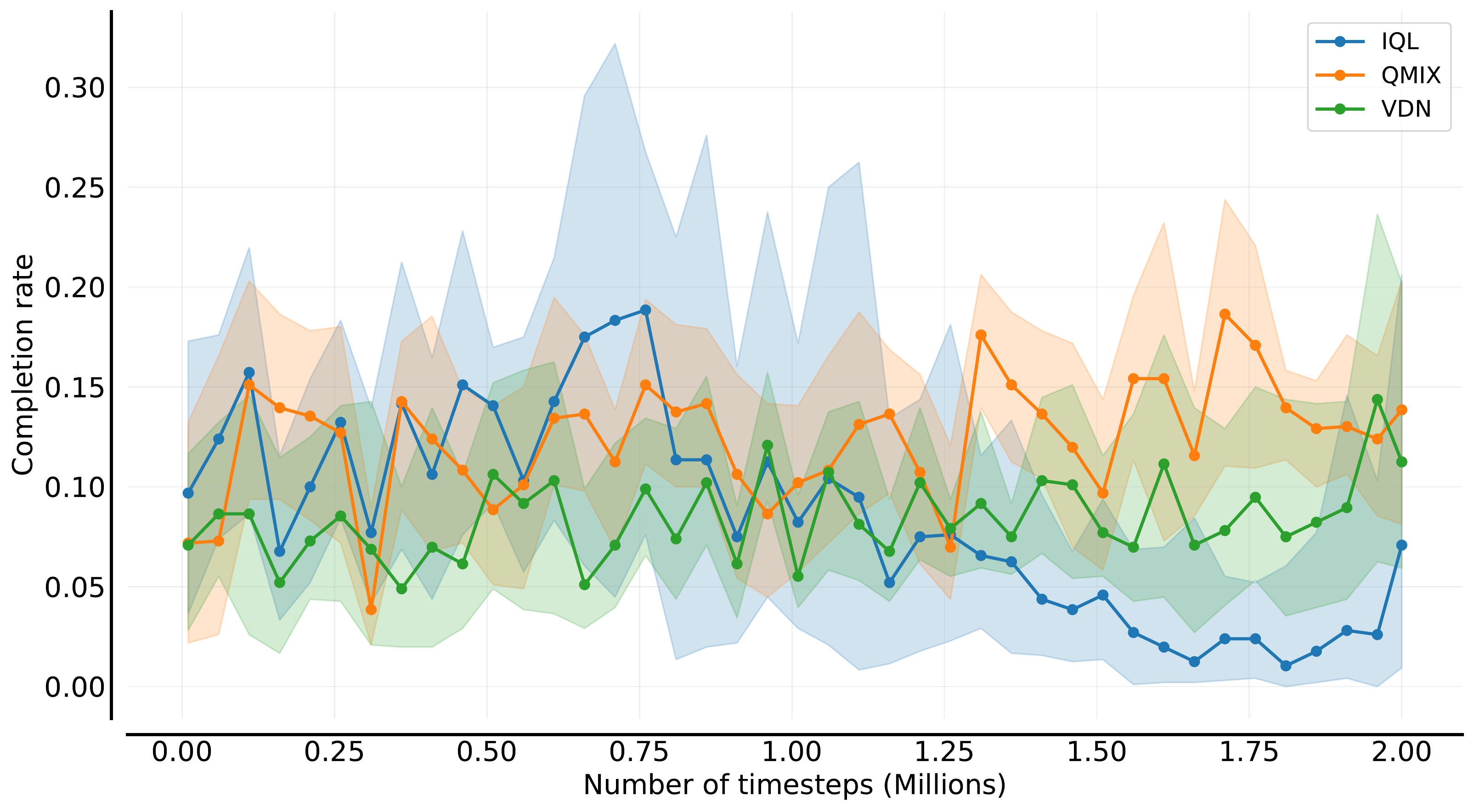}
    \end{subfigure}

    \caption{\textit{Sample efficiency curves for experiments}. \textbf{(a)} Normalized return. \textbf{(b)} Completion rate. }
    \label{fig:sample_efficiency}
\end{figure}

From Figure \ref{fig:sample_efficiency} it can be noted that no particular algorithm is more efficient than any other algorithm and that all algorithms achieve relatively similar final performance. From Figure \ref{fig:sample_efficiency} (a) it can be noted that IQL and QMIX do reach a slightly higher final mean return than VDN. 

\subsubsection{Aggregate score performance}
\begin{figure}[H]
    \centering
    
    \begin{subfigure}[t]{0.75\textwidth}
        \includegraphics[width=\linewidth, valign=t]{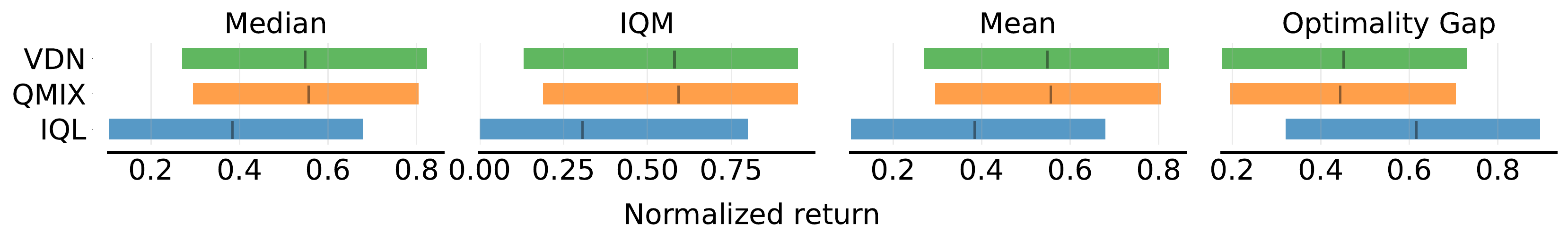}
    \end{subfigure}
    \hspace{1mm}
    
    \begin{subfigure}[t]{0.75\textwidth}
        \includegraphics[width=\linewidth, valign=t]{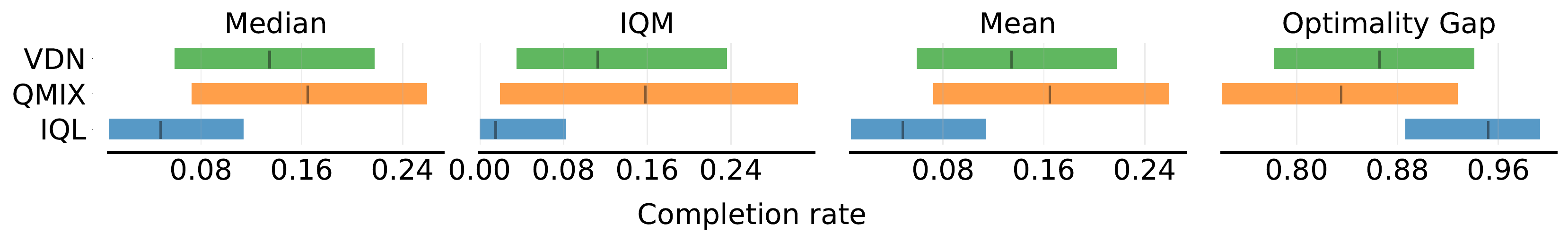}
    \end{subfigure}

    \caption{\textit{Per task performance on a $25 \times 25$ flatland grid}. \textbf{(Top)} Normalized return. \textbf{(Bottom)} Completion rate. }
    \label{fig: aggregate_performance}
\end{figure}

All aggregated scores are done using the aggregation functions as shown in Figure \ref{fig: aggregate_performance}. One aggregation function to note is the \textit{Optimality Gap} which may be thought of as the how far an algorithm is from optimal performance at a given task. For this reason, a lower score is considered to be desirable. The confidence intervals shown alongside the point estimates (black bars) are the 95\% stratified bootstrap confidence intervals. 

One can note from the top and bottom figure in Figure \ref{fig: aggregate_performance} that there is large variance in algorithm performance for both metrics used and it is hard to distinguish which algorithm has superior performance, particularly between VDN and QMIX. A clear outlier is IQL which consistently performs worse, across all metrics, than VDN and QMIX.

\subsubsection{Performance profiles}

\begin{figure}[H]
    \centering
    \begin{subfigure}[t]{0.02\textwidth}
        \scriptsize
        \textbf{(a)}
    \end{subfigure}
    ~
    \begin{subfigure}[t]{0.3\textwidth}
        \includegraphics[width=\linewidth, valign=t]{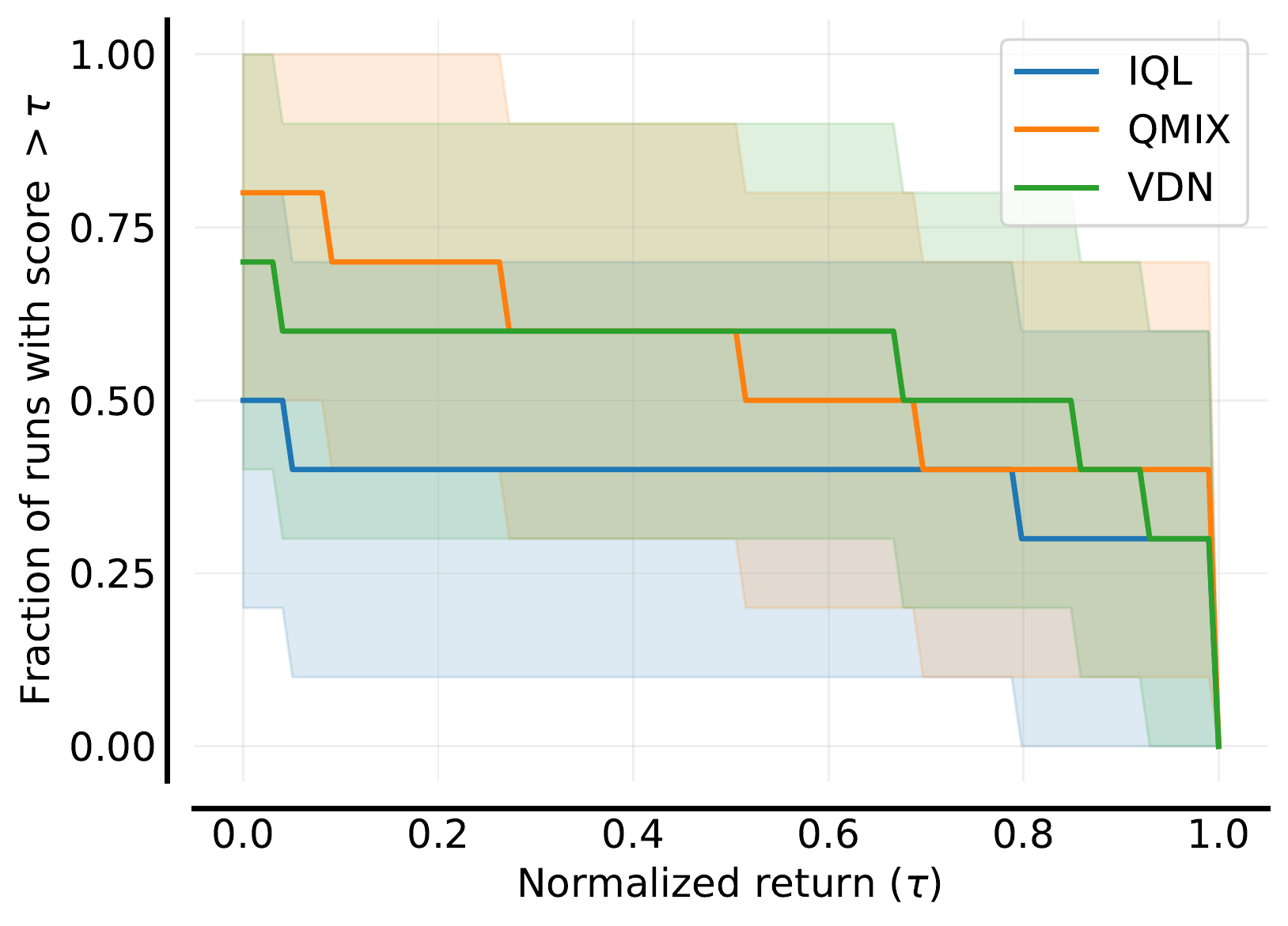}
    \end{subfigure}
    ~
    \begin{subfigure}[t]{0.02\textwidth}
        \scriptsize
        \textbf{(b)}
    \end{subfigure}
    ~
    \begin{subfigure}[t]{0.3\textwidth}
        \includegraphics[width=\linewidth, valign=t]{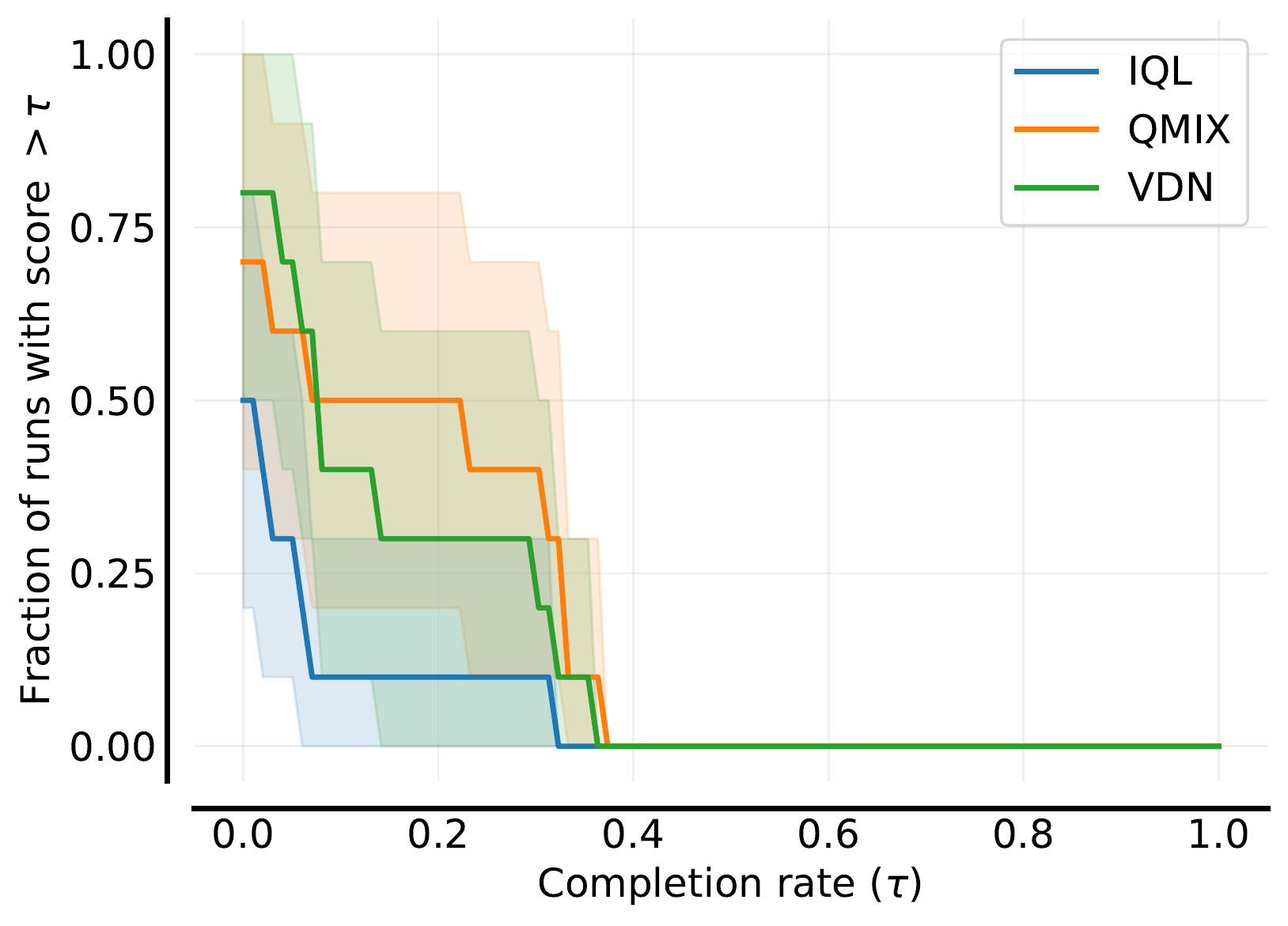}
    \end{subfigure}

    \caption{\textit{Performance profiles for experiments}. \textbf{(a)} Normalized return. \textbf{(b)} Completion rate. }
    \label{fig: performance_profiles}
\end{figure}

From Figure \ref{fig: performance_profiles} in can be noted that the performance profiles paint a similar picture to the sample efficiency curves and the aggregated algorithm scores. IQL is consistently outperformed by VDN and QMIX. The performance profiles also clearly illustrate the no algorithm achieves a particularly high completion rate, highlighting the poor overall performance of all algorithms on the environment task. 

\subsubsection{Probability of improvement}

Probability of improvement plots should be interpreted as the probability that an algorithm X has superior performance than algorithm Y with a low score indicating that algorithm Y likely to be better than algorithm X and vice versa for a high score.

\begin{figure}[h]
    \centering
    \begin{subfigure}[t]{0.02\textwidth}
        \scriptsize
        \textbf{(a)}
    \end{subfigure}
    ~
    \begin{subfigure}[t]{0.3\textwidth}
        \includegraphics[width=\linewidth, valign=t]{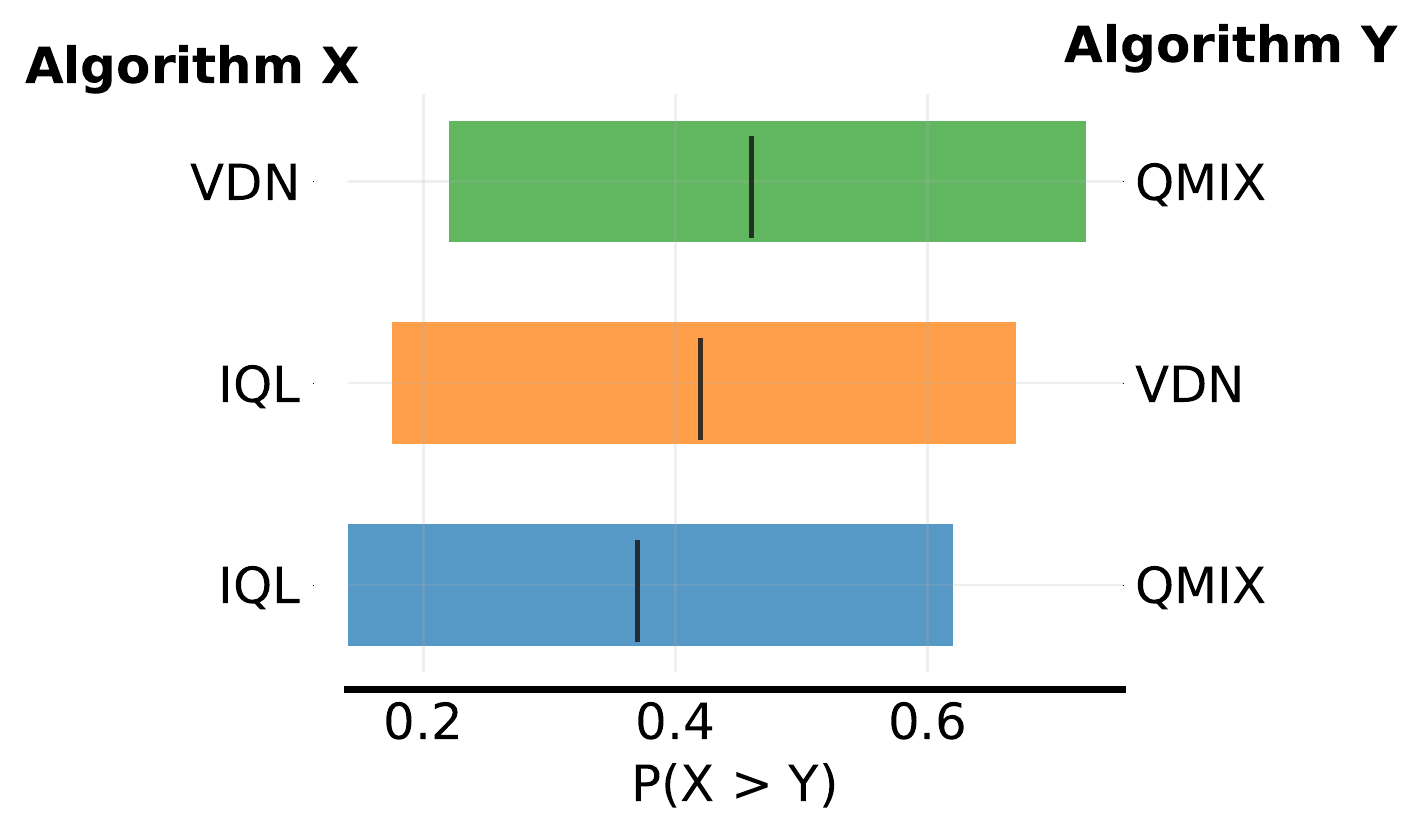}
    \end{subfigure}
    ~
    \begin{subfigure}[t]{0.02\textwidth}
        \scriptsize
        \textbf{(b)}
    \end{subfigure}
    ~
    \begin{subfigure}[t]{0.3\textwidth}
        \includegraphics[width=\linewidth, valign=t]{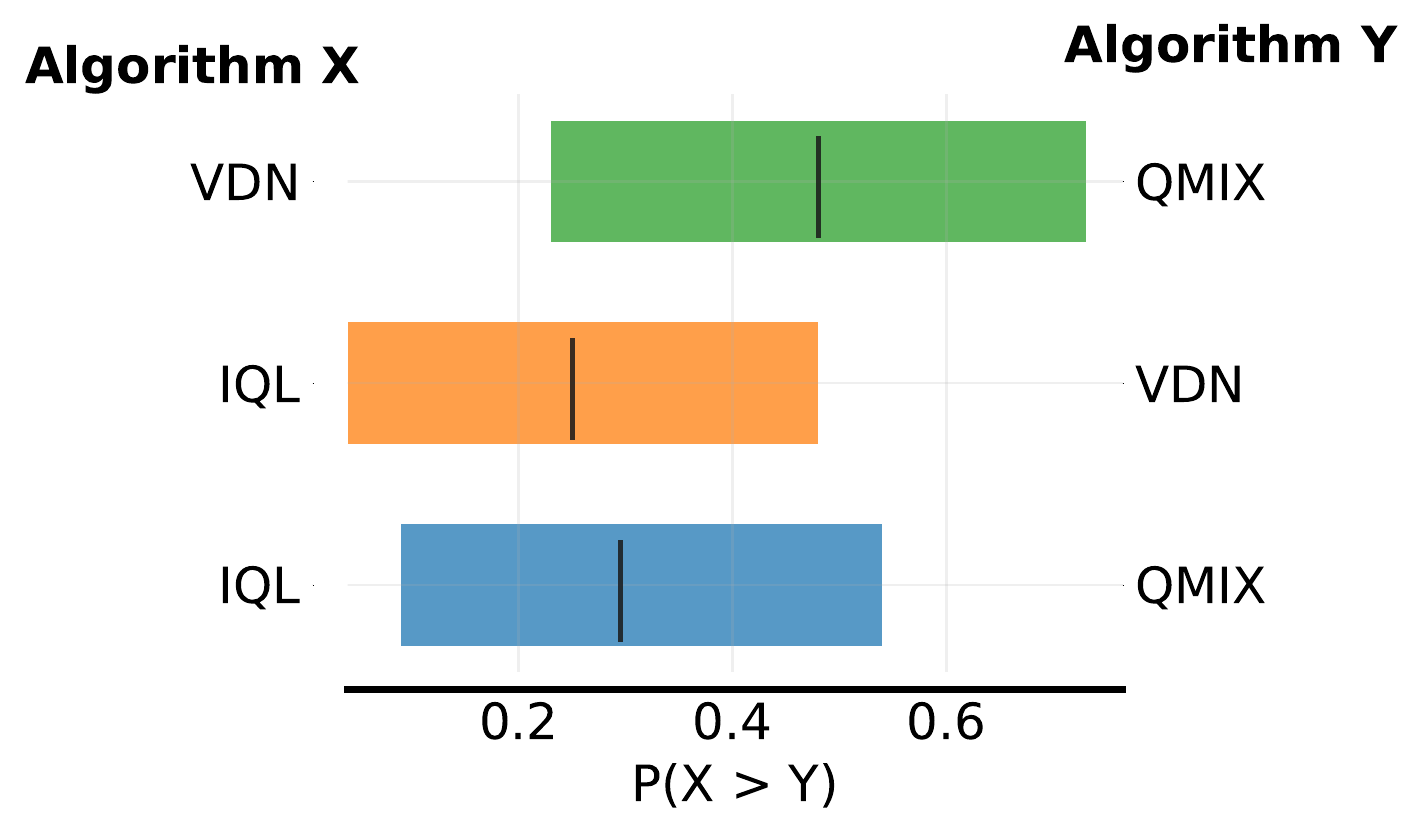}
    \end{subfigure}

    \caption{\textit{Probability of improvement plots}. \textbf{(a)} Normalized return. \textbf{(b)} Completion rate. }
    \label{fig: prob_improvement}
\end{figure}

From Figure \ref{fig: prob_improvement} one can note again that IQL is outperformed by VDN and QMIX using both metrics considered. It can also be noted again, that the performance of VDN and QMIX are relatively similar. 

\subsubsection{Tabular Results}

We report the IQM aggregated over the task for all algorithms with the 95\% stratified bootstrap CI as well as the mean absolute performance of all algorithms on the task with the 95\% CI. These scores collectively, as well as the sample efficiency curves sketch a full picture of the performance for a given algorithm. One can, at a glance, see the performance that the the best policy for a particular algorithm is able to achieve from the tabular results, but one can also get a clear sense of the robustness of a particular algorithm by considering the sample efficiency curves. This makes for transparent result reporting. 

The tabular results once again confirm all previous results in that IQL has inferior performance on the task when compared to its value factorisation counterparts and that VDN and QMIX obtain very similar performance. Due to the larger confidence intervals however, no clear conclusions can be drawn since the performance of all algorithms overlap when taking the CIs into account. One can also notice from Tables \ref{tbl: bootstrap_IQM} \& \ref{tbl: absolute_metrics} that the absolute metric and IQM scores are very similar. The reason from this is because we only consider a single task in our environment. The true power of the tools that were used will be better illustrated when multiple tasks are considered.

\begin{table}[h]
  \caption{IQM of absolute metrics for experiments with 95\% Stratified Bootstrap CIs}
  \label{tbl: bootstrap_IQM}
  \centering
  \makebox[1 \textwidth][c]{ 
  \begin{tabular}{ccc}
    \toprule
    Algorithm & Normalized Returns & Completion Rate  \\
    \midrule
    IQL & 0.307 (0.0, 0.799) & 0.015 (0.0, 0.083) \\
    QMIX & 0.593 (0.189, 0.949) & 0.158 (0.019, 0.304) \\
    VDN & 0.581 (0.131, 0.949 & 0.113 (0.035, 0.236) \\
    \bottomrule
  \end{tabular}}
\end{table}

\begin{table}[h]
  \caption{Mean per task absolute metrics with 95\% CIs}
  \label{tbl: absolute_metrics}
  \centering
  \makebox[1 \textwidth][c]{ 
  \begin{tabular}{ccc}
    \toprule
    Algorithm & Normalized Returns & Completion Rate  \\
    \midrule
    IQL & 0.384 (0.08, 0.688) & 0.048 (0.00, 0.109) \\
    QMIX & 0.556 (0.284, 0.828) & 0.164 (0.066, 0.263) \\
    VDN & 0.548 (0.254, 0.842) & 0.134 (0.05, 0.218) \\
    \bottomrule
  \end{tabular}}
\end{table}

\subsubsection{Overall findings}
Due to the large variance in algorithm performance, we cannot draw any strong conclusions regarding algorithm performance from these experiments, but we have been able to illustrate to use of our guideline and how it gives a full overview of both the absolute and overall performance of a set of algorithms on a particular task. We will continually update our demonstration by adding more flatland tasks, tuning algorithms and, ultimately, adding more environments to this experiment.

\end{document}